\documentclass[9pt,journal,technote]{IEEEtran}

\usepackage[usenames]{color}
\usepackage[ansinew]{inputenc}
\usepackage{inputenc}
\usepackage{graphicx}
\usepackage{algorithm}
\usepackage{algorithmic}
\usepackage{amsfonts}
\usepackage{amssymb}
\usepackage{amsbsy}
\usepackage{multirow}
\usepackage{array}
\usepackage{color}
\usepackage{soul}
\usepackage[normalem]{ulem} 
\usepackage{times}
\usepackage[all]{xy}
\usepackage{lscape}
\usepackage{flushend,pmat}
\usepackage{url}
\usepackage{float}
\usepackage{rotating}

\newcommand{\e}{\mathbf{e}}

\newcommand{\x}{\mathbf{x}}

\renewcommand{\v}{\nu}
\newcommand{\w}{\mathbf{w}}

\newcommand{\A}{\mathbf{A}}
\newcommand{\B}{\mathbf{B}}

\newcommand{\D}{\mathbf{D}}
\newcommand{\K}{\mathbf{K}}
\newcommand{\II}{\mathbf{I}}

\newcommand{\W}{\mathbf{W}}

\newcommand{\E}{\mathbf{E}}

\renewcommand{\baselinestretch}{1}  
\newcommand{\rota}[1]{\begin{sideways}#1\end{sideways}}

\begin{document}

\title{Optimized Kernel Entropy Components}

\author{Emma Izquierdo-Verdiguier, Valero Laparra,\\ 
Robert Jenssen,~\IEEEmembership{Senior~Member,~IEEE}, 
Luis~G\'omez-Chova,~\IEEEmembership{Senior~Member,~IEEE}, 
and Gustau~Camps-Valls,~\IEEEmembership{Senior~Member,~IEEE} 
\IEEEcompsocitemizethanks{\IEEEcompsocthanksitem EIV, VL, LGC and GCV are with the 
Image Processing Laboratory (IPL), Universitat de Val\`encia, Spain.\protect\\
\{emma.izquierdo,valero.laparra,luis.gomez-chova,gustau.camps\}@uv.es, http://isp.uv.es. \protect\\
RJ is with the University of Troms\o, Norway. robert.jenssen@uit.no, http://ansatte.uit.no/robert.jenssen/}
\thanks{This work has been supported by the Spanish Ministry of Economy and Competitiveness (MINECO) under project TIN2012-38102-C03-01 (LIFE-VISION), the Generalitat Valenciana under project GV/2013/079, and the ERC consolidator grant SEDAL-647423. The work has also been supported by the Research Council of Norway over FRIPRO/IKTPLUSS grant no. 234498.}
\thanks{Copyright (c) 2011 IEEE. Personal use of this material is permitted. Permission from IEEE must be obtained for all other users, including reprinting/republishing this material for advertising or promotional purposes, creating new collective works for resale or redistribution to servers or lists, or reuse of any copyrighted components of this work in other works. DOI: 10.1109/TNNLS.2016.2530403.}}

\markboth{IEEE Transactions on Neural Networks and Learning Systems, 2016}%
{Izquierdo et al.~\MakeLowercase{\textit{et al.}}: Optimized Kernel Entropy Components}

\IEEEcompsoctitleabstractindextext{%
\begin{abstract}
This work addresses two main issues of the standard Kernel Entropy Component Analysis (KECA) algorithm: the optimization of the kernel decomposition and the optimization of the Gaussian kernel parameter. KECA roughly reduces to a sorting of the importance of kernel eigenvectors by entropy instead of by variance as in Kernel Principal Components Analysis. In this work, we propose an extension of the KECA method, named Optimized KECA (O\-KE\-CA), that directly extracts the optimal features retaining most of the data entropy by means of compacting the information in very few features (often in just one or two). The proposed method produces features which have higher expressive power. In particular, it is based on the Independent Component Analysis (ICA) framework, and introduces an extra rotation to the eigen-decomposition, which is optimized via gradient ascent search. This maximum entropy preservation suggests that OKECA features are more efficient than KECA features for density estimation. In addition, a critical issue in both methods is the selection of the kernel parameter since it critically affects the resulting performance. Here we analyze the most common kernel length-scale selection criteria. Results of both methods are illustrated in different synthetic and real problems. Results show that 1) OKECA returns projections with more expressive power than KECA, 2) the most successful rule for estimating the kernel parameter is based on maximum likelihood, and 3) OKECA is more robust to the selection of the length-scale parameter in kernel density estimation.
\end{abstract}
}

\maketitle

\IEEEdisplaynotcompsoctitleabstractindextext

\IEEEpeerreviewmaketitle

\section{Introduction}
Kernel entropy component analysis (KECA)~\cite{Jenssen2009,Jenssen13spm} was recently proposed as a general information-theoretic method for feature extraction and dimensionality reduction in pattern analysis and machine intelligence. It has proven useful in different applications, e.g. remote sensing data analysis~\cite{Gomez12,LuoWu2012,LuoEtal2013}, face recognition~\cite{ShekarEtal2011}, chemical processes modelling~\cite{JiangEtal2013}, high-dimensional celestial spectra reduction~\cite{HuEtal2013} and audio processing~\cite{XieGuan2012}.  
Several extensions have been proposed for feature selection~\cite{ZhangHancock2012}, class-dependent feature extraction~\cite{ChengEtal2011} and semisupervised learning as well~\cite{MyhreJenssen2012}. The KECA algorithm is fundamentally different from, but still intimately related to, the vastly successful spectral kernel multivariate signal processing methods such as kernel principal components analysis (KPCA)~\cite{Scholkopf98}, kernel canonical correlation analysis (KCCA)~\cite{Lai00} and kernel partial least squares (KPLS)~\cite{rosipal2001kplshilbert}, just to name a few~\cite{Arenas13spm}. 

One distinguishing feature of KECA is that the method originates from kernel density estimation (KDE) \cite{Silverman86,Girolami02,Duin76}, as do e.g.\ principal curves estimation \cite{OzertemErdogmus2011} and the family of information theoretic learning methods \cite{Principe2010}. In KDE, the key is the kernel function, locally approximating the underlying probability density function (PDF). This in turn enables estimation of entropy, a quantity that describes the shape of the PDF \cite{CoverThomas}. The KDE kernel must be a non-negative function that integrates to one (i.e.\ a density) but need not be positive semi-definite (PSD). The KDE kernel is versatile in that sense. However, many KDE kernels are PSD, well-known examples include the Gaussian kernel, the Student kernel and the Laplacian kernel \cite{KimScott2012} functions.

If the KDE kernel used in KECA is PSD, then there are close relations to the aforementioned \emph{kernel} signal processing methods, in the sense that the kernel computes an inner-product in a reproducing kernel Hilbert space (RKHS). In this situation, KECA,  KPCA, KCCA and KPLS are based on RKHS learning algorithms to maximize e.g.\ the feature space variance, correlation or alignment with the output variables. PSD KECA hence bridges KDE, information theoretic learning and RKHS learning.

Although both KDE and RKHS kernel methods have experienced great success, all kernel-based methods, including the one in this work, are sensitive to the kernel function used. For instance, many kernel methods depend heavily on a bandwidth, or length-scale, parameter. In addition, all the aforementioned spectral methods may need a considerable number of components (eigenvalues and eigenvectors) in order to properly describe the data. This may be undesirable e.g.\ in compression and data visualization contexts.

In this work, we take advantage of the KDE foundation of KECA (see also ~\cite{Girolami02} for further details), and introduce an optimization procedure aiming at compressing the entropy information into optimal directions in feature space. The proposed approach is the first kernel-based unsupervised feature extraction method in an information theoretical sense. The approach introduces a rotation procedure that resembles the one in Fast Independent Component Analysis (ICA)~\cite{Hyvarinen01}. Extracted OKECA components presents two major advantages:
\begin{enumerate}
\item OKECA shows great robustness to the kernel bandwidth parameter. This is important, as there are no universally accepted kernel size selection procedure for unsupervised KDE-based kernel methods.
\item We use OKECA in order to improve the KDE. This is achieved with far fewer components compared to KECA.
\end{enumerate}

The rest of the paper is organized as follows. Section~\ref{KECA2} presents the OKECA formulation and proposes a density estimation that exploits kernel feature characteristics. Section~\ref{experiments} is devoted to the analysis of the results. We use OKECA as a feature extraction method and analyze the retained entropy, show the estimated PDF, and perform data classification. We conclude the paper in Section~\ref{conclusions}.

\section{Optimized Kernel Entropy Components (OKECA)}
\label{KECA2}

KECA relies on the eigen-decomposition of the uncentered kernel matrix (see below) and sorts the eigenvectors according to the so-called entropy values of the projections. This is tightly related to information-theoretic concepts and KDE. The entropy-relevant dimensionality reduction transforms the dataset in a way that reveals cluster structure and hence information about the underlying classes in the data~\cite{JenssenCh9,Jenssen13spm}. 

\subsection{Kernel Entropy Components (KECA)}

To be more precise, the measure of information used in~\cite{Jenssen2009} is the Renyi's second order entropy, given by
\begin{equation}
H = - \log \int p^2(\x) d\x,
\end{equation}
where $p(\x)$ is the PDF generating the data.  Given a dataset ${\mathcal D}=\{\x_1,\ldots,\x_n\}$ of dimensionality $d$, the entropy may be estimated through KDE~\cite{Silverman86} (see Sec.\ \ref{KDE}) as $- \log \v$, where $\v$ is the so-called \emph{information potential}~\cite{Principe2010}:
\begin{equation}
\v = \frac{1}{n^2} \boldsymbol{1}_n^\top \K \boldsymbol{1}_n 
\label{eq:KECA_entropy_orig}
\end{equation}
where $\K_{ij}=k(\x_i,\x_j)$ is any valid KDE kernel comprising the $(n \times n)$ kernel matrix and $\boldsymbol{1}_n$ is a $n$-dimensional vector of ones.  Using the 
kernel decomposition introduced in~\cite{Jenssen2009}:
\begin{equation}
	\K = \A \A^\top = (\E \D^{\frac{1}{2}})(\D^{\frac{1}{2}} \E^\top),
\label{eq:KECA_decomposition}
\end{equation}
we may write 
\begin{equation}
	\v(N_c) = \sum_{j=1}^{N_c} \bigg(\sum_{i=1}^n \A_{ij} \bigg)^2 = \sum_{j=1}^{N_c} \left( \lambda_j^{\frac{1}{2}} {\mathbf 1}_{n}^{\top} \mathbf{e}_j \right)^2.
\label{eq:KECA_entropy}
\end{equation}
In this expression, $\E$ contains the eigenvectors in columns, $\E=[\e_1,\e_2,\ldots,\e_n]$, and $\D$ is a diagonal matrix containing the eigenvalues of $\K$, i.e. $\D_{ii}=\lambda_i$, and $N_c \leq n$ is the number of retained components. The terms $( \lambda_j^{\frac{1}{2}} {\mathbf 1}_{n}^{\top} \mathbf{e}_j )^2$ are denoted entropy values.

As mentioned earlier, if the KDE kernel is PSD, then there is a close connection between KECA and un-centered KPCA since the kernel function in that case reproduces the dot product between two samples mapped to a RKHS ${\mathcal H}$ via $\boldsymbol{\phi}(\cdot)$, i.e. $\K_{ij}=k(\x_i,\x_j)=\boldsymbol{\phi}(\x_i)^\top\boldsymbol{\phi}(\x_j)$. Note that centering of the kernel matrix $\K$ makes no sense in the KDE and entropy context of Eq.\ (\ref{eq:KECA_entropy_orig}), as this would correspond to $\v = 0$, i.e.\ infinite entropy.  Hence, $\D^{\frac{1}{2}} \E^\top$ is the uncentered projection of the feature space data ${\mathcal D}_{\mathcal H} = \{\boldsymbol{\phi}(\x_1),\dots,\boldsymbol{\phi}(\x_n)\}$ onto all the principal axes in the feature space as~\cite{Jenssen2009,Jenssen13spm}. These projections may be sorted according to their contribution to the input space entropy as measured by the information potential (the entropy values in Eq.\ (\ref{eq:KECA_entropy})), constituting the KECA procedure.

However, the projections and their entropy content are fully dependent on the quality of the KDE performed via the kernel function. Moreover, using the eigen-decomposition procedure may not be optimal to find the best projections from an entropy perspective.

\subsection{Proposed Optimized KECA (OKECA)}

The novel approach proposed in this work searches for a basis that maximizes the information potential in as few components as possible. Towards that end, we propose a new optimal (in information-theoretic sense) feature extraction and unsupervised dimensionality reduction method. The procedure corresponds to optimally capturing in these components the high information potential part of the data (low entropy), which typically corresponds to the structure of the data in terms of class or cluster information.

Unlike the KECA method which only applies a different sorting of KPCA features, the proposed method is motivated by the classical Independent Component Analysis (ICA) formulation~\cite{Hyva00} in which, after the whitening step (applying $\D^{\frac{1}{2}} \E^{\top}$), there is an extra rotation (applying $\W^{\top}$) that maximizes the independence between components. Note that $\W$ is an orthonormal linear transformation, i.e. $\W \W^\top = \II$. Similar ideas have been applied in kernel-based components analysis (see for instance~\cite{Pan11}). Following the ICA rationale, we now aim at a new kernel matrix decomposition: 
\begin{equation}
	\K = \B \B^\top = (\E \D^{\frac{1}{2}} \W)(\W^\top \D^{\frac{1}{2}} \E^\top).
\label{eq:KECA_new_decomposition}
\end{equation}
Note that the kernel matrix does not change, but the modification allows us to directly find the basis that maximize the information potential with respect to the number of retained components. Therefore, for each column vector $\w_k$ in $\W=[\w_1,\ldots,\w_n]$, we maximize:
\begin{equation}
	{\mathcal L} = \left( {\mathbf 1}_{n}^{\top} \E \D^{\frac{1}{2}} \w_k \right)^2,
\label{eq:W_optimization}
\end{equation}
where each $\w_k$ is restricted to be normal $\|\w_k\|_2 = 1$ and to be orthonormal to the previous $\w_l$, $\forall l<k$. This deflationary procedure ensures that the obtained solution retains more (or equal) information potential than the one obtained by the standard KECA in fewer components. 

In order to solve the OKECA optimization problem in Eq.\ (\ref{eq:W_optimization}), a gradient-ascent approach can be followed:
\begin{equation}
\w_k(t+1) = \w_k(t) + \tau \frac{\partial \mathcal{L}}{\partial \w_k(t)},
\label{eq:gradient_ascent}
\end{equation}
where $\tau$ is the step size and the gradient is:
\begin{equation}
\frac{\partial \mathcal{L}}{\partial \w_k} = 2  ( {\mathbf 1}_{n}^{\top} \E \D^{\frac{1}{2}} \w_k )   ( {\mathbf 1}_{n}^{\top} \E \D^{\frac{1}{2}} )^{\top}.
\label{eq:gradient}
\end{equation}
In this paper, we adopted a simple scheme for early stopping that ensures that the gradient achieves a region where additional iterations did not modify the cost function significantly.
A pseudocode summary of the OKECA feature extraction procedure is given in Algorithm~\ref{pckTrain}. A Matlab implementation of the algorithm is available at {\tt http://isp.uv.es/code/okeca.htm} for the interested reader. While other more sophisticated optimization algorithms could be deployed here, in our experiments we observed that this simple gradient-ascent strategy performed consistently even in the presence of noise.

\begin{figure*}[t!]
  \begin{center}
  \setlength{\tabcolsep}{0pt}
  \begin{tabular}{ccccc}
    &  $\sigma_{ML}$ &  $\sigma_{Silv}$ &  $\sigma_{d1}$ &  $\sigma_{d2}$ \\
  \includegraphics[width = 2.8cm]{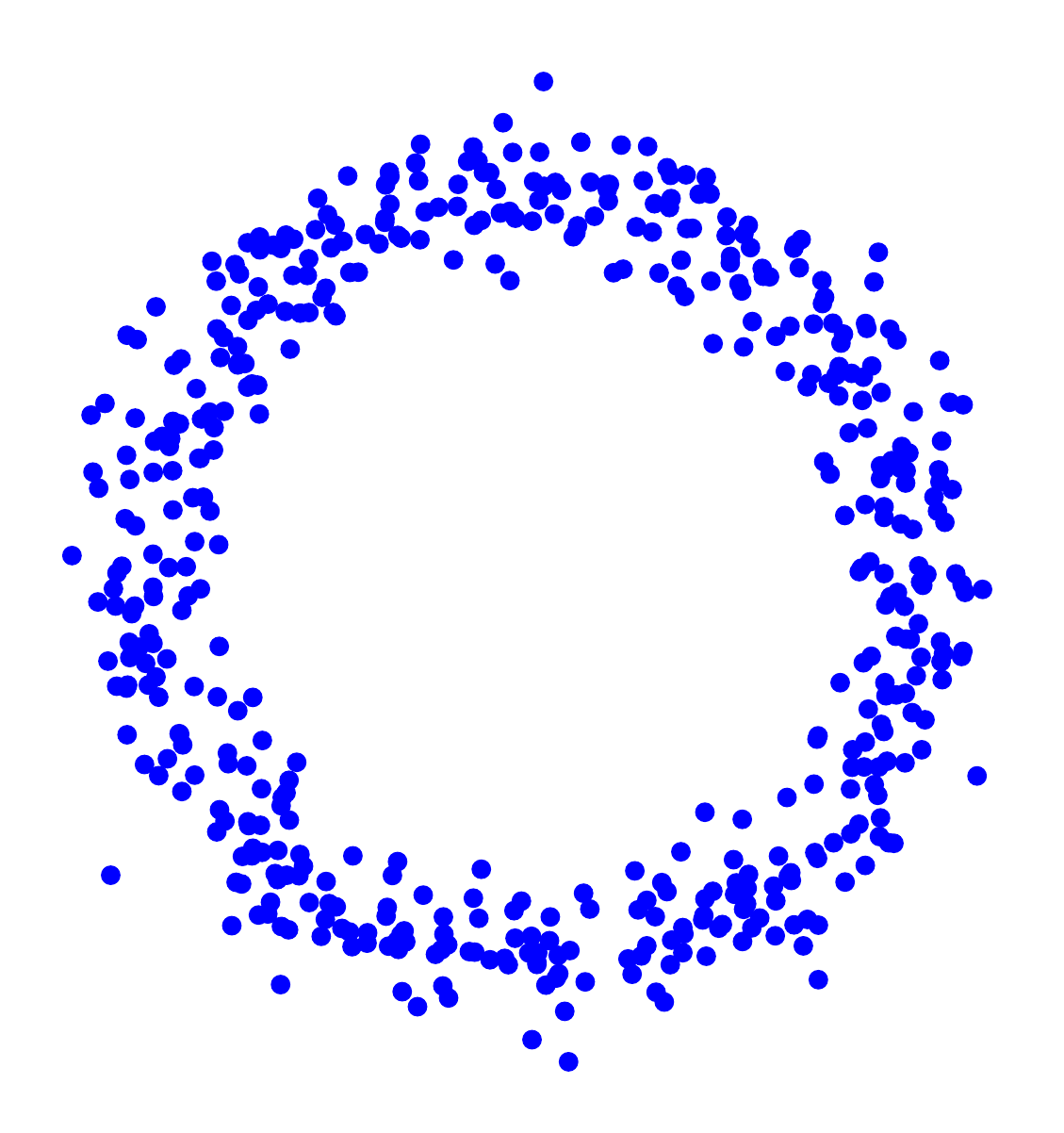} & \includegraphics[width=3.5cm]{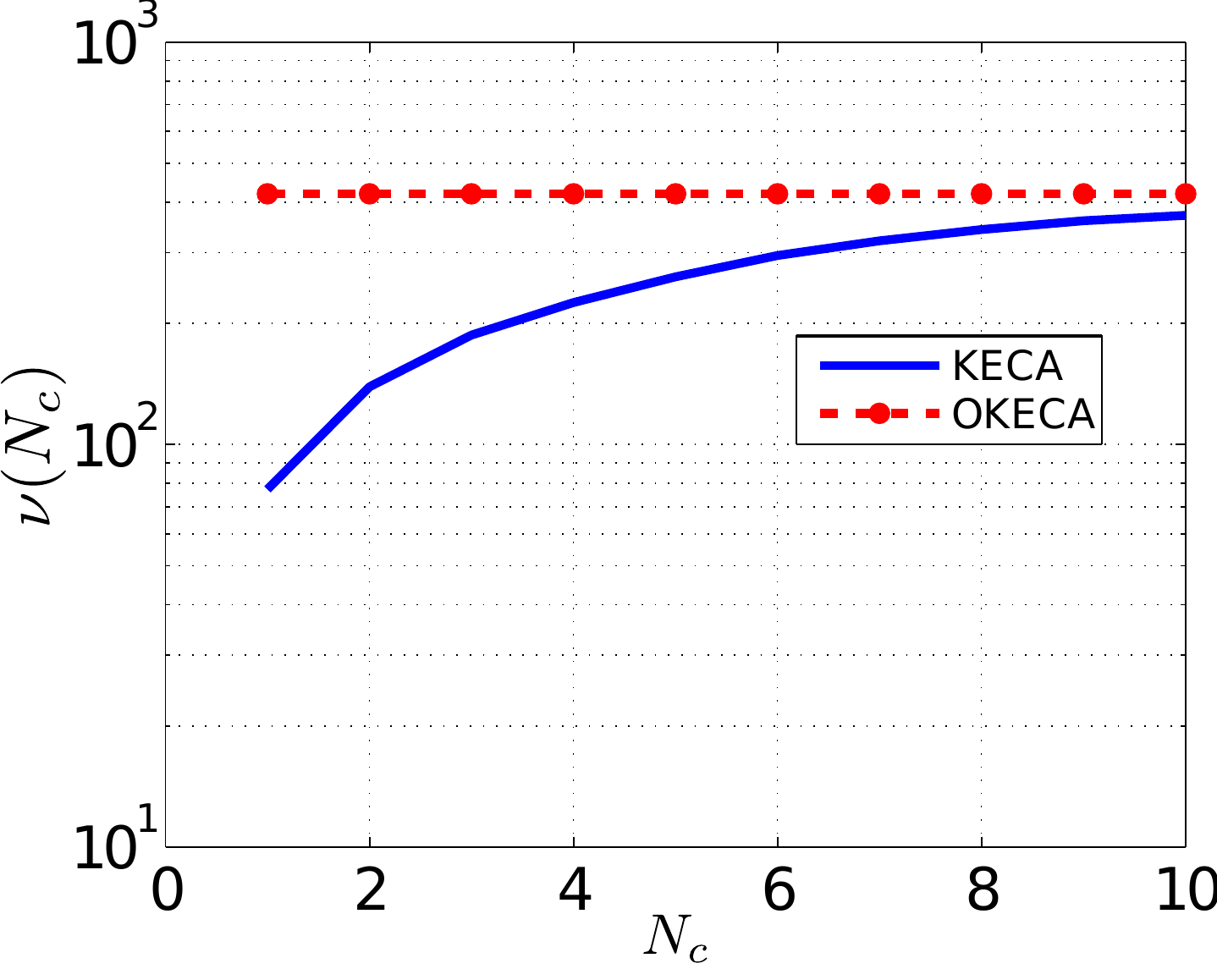} & \includegraphics[width=3.5cm]{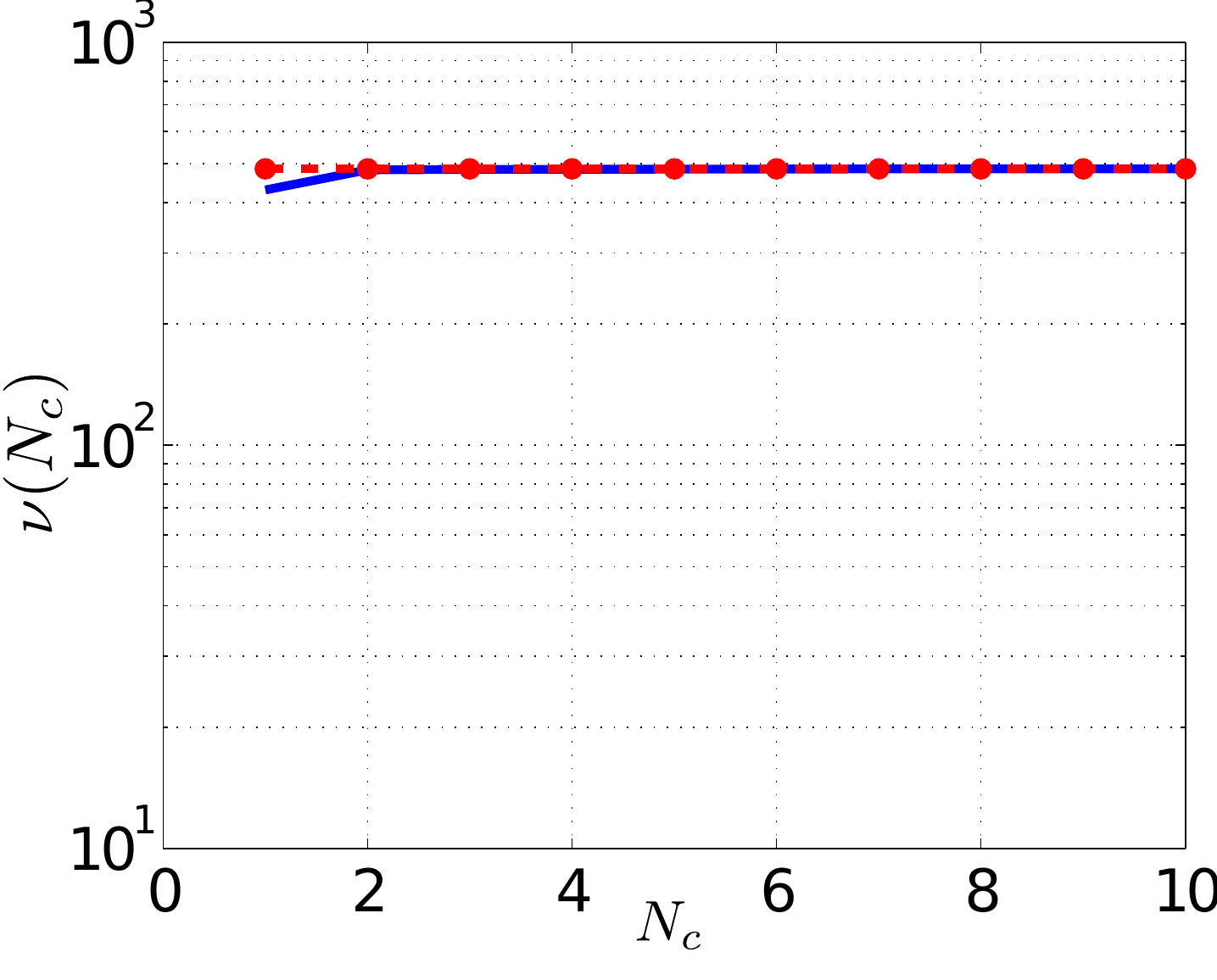} & \includegraphics[width=3.5cm]{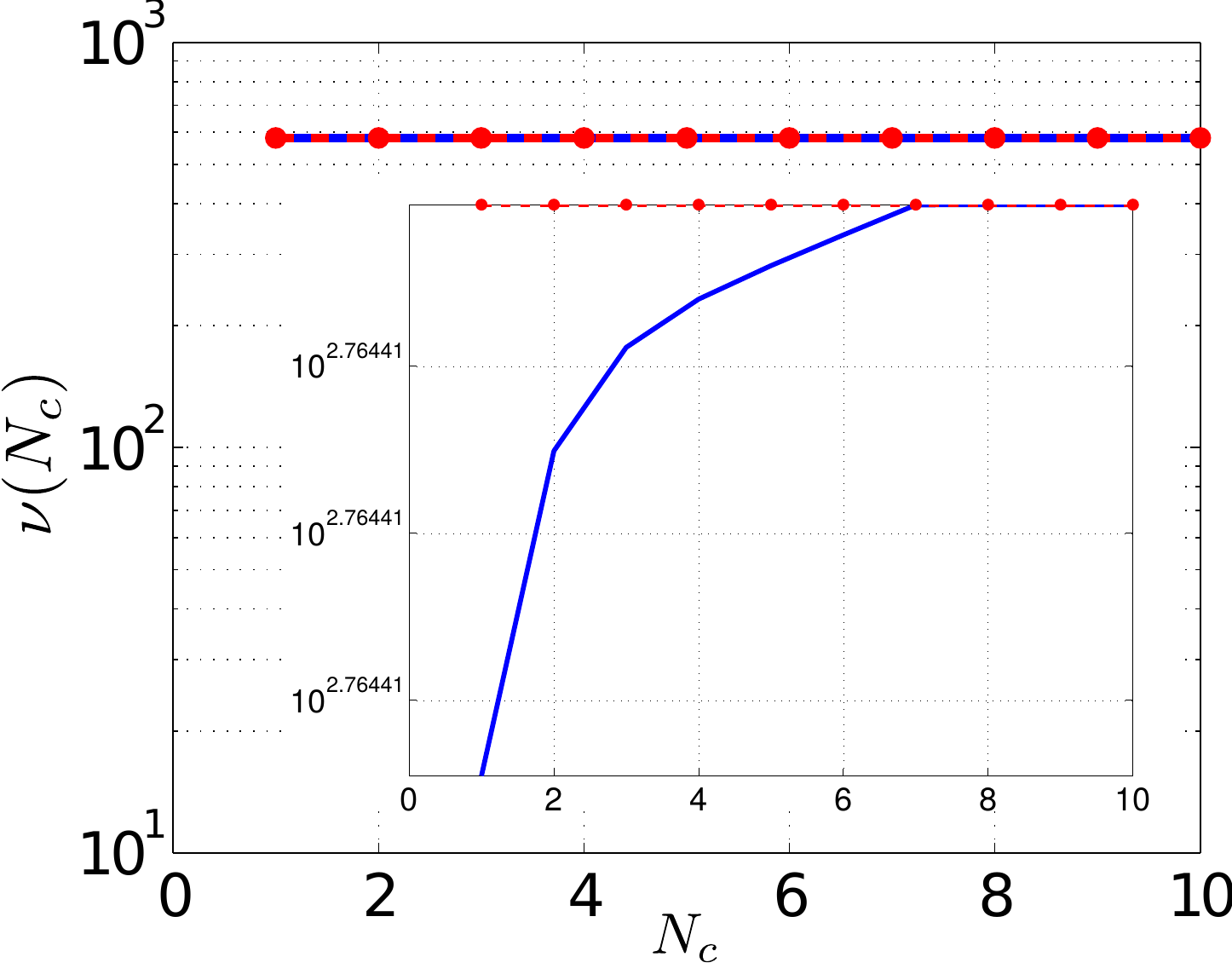} & \includegraphics[width=3.5cm]{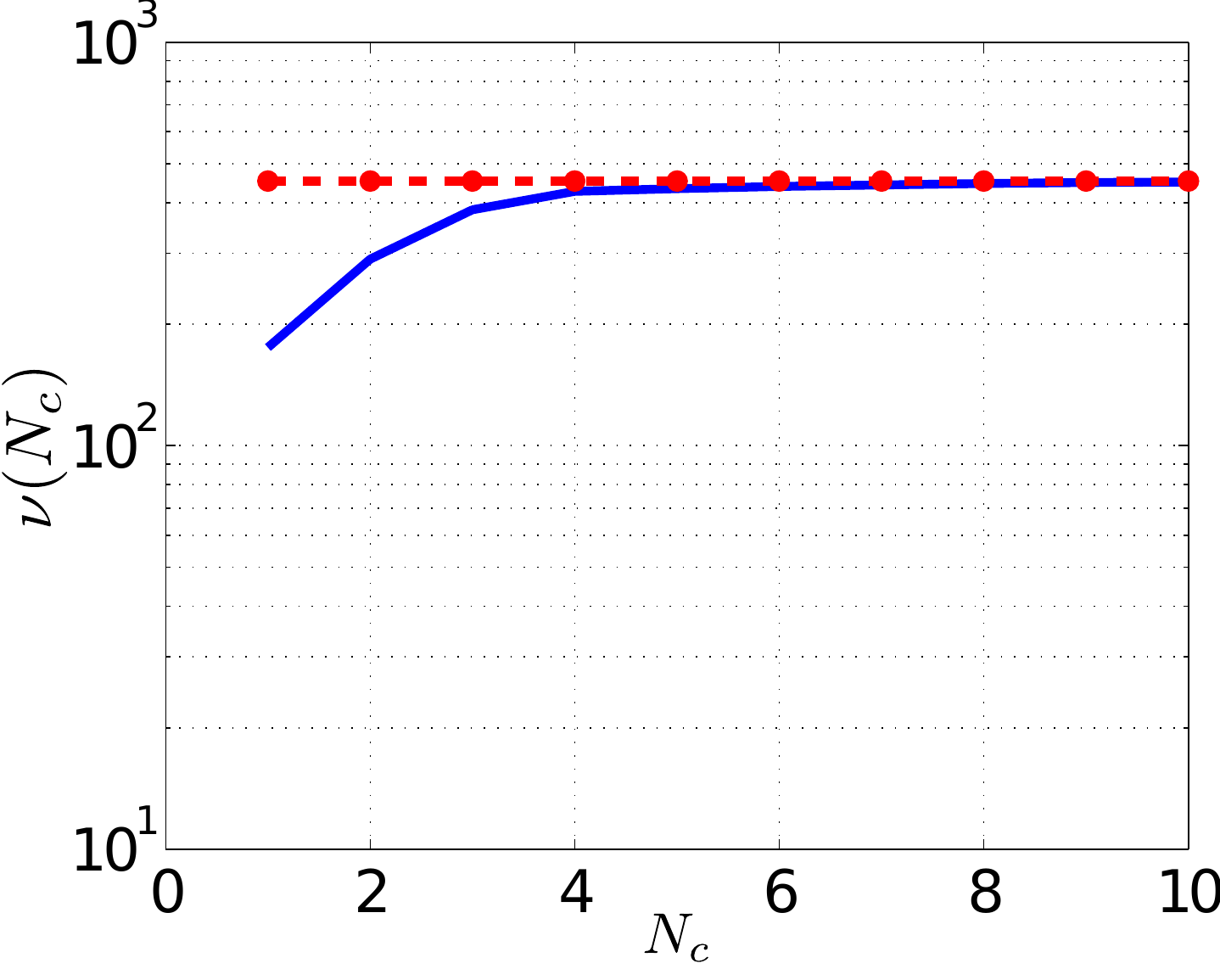}  \\
  \includegraphics[width = 2.8cm]{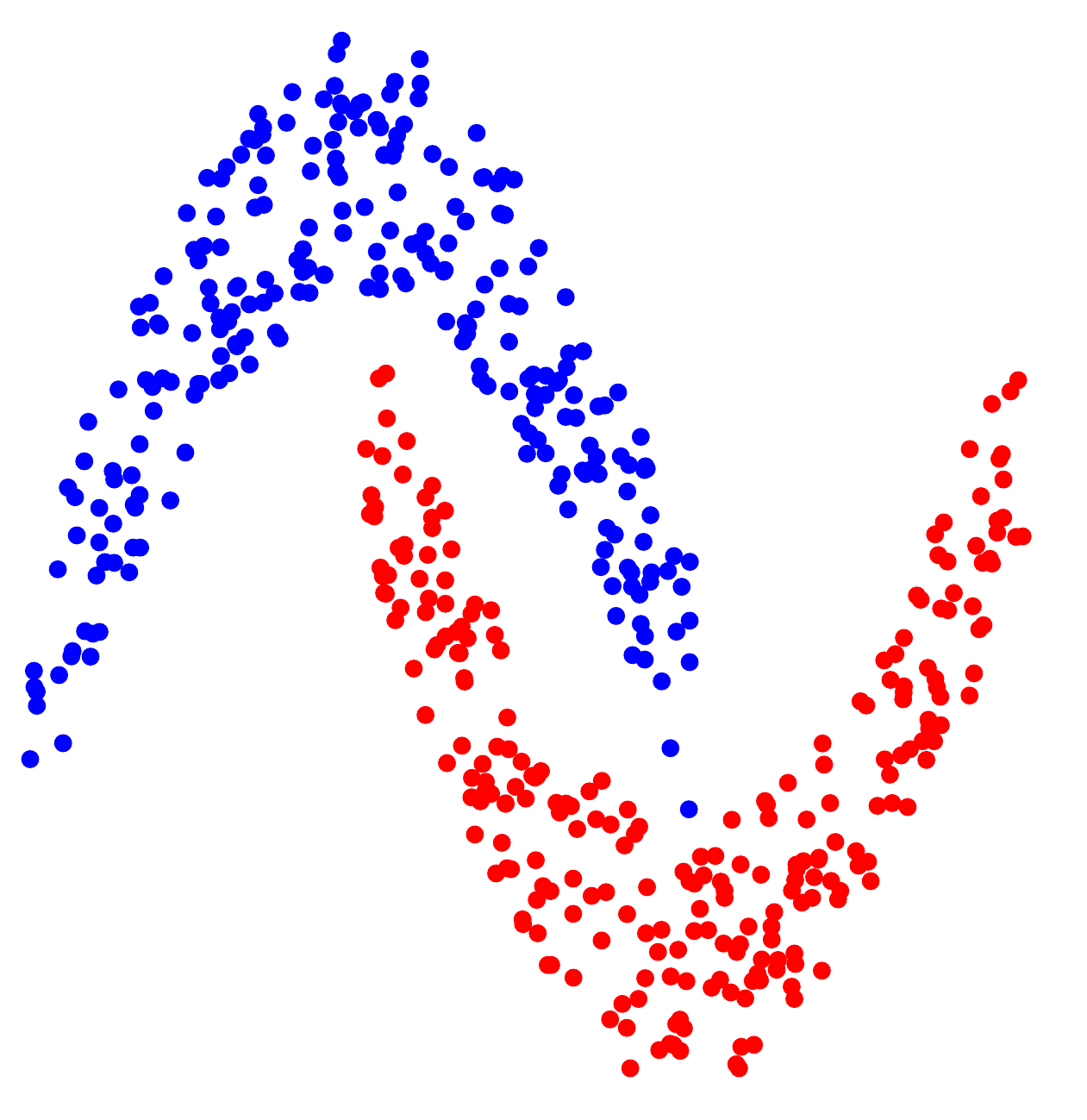} & \includegraphics[width=3.5cm]{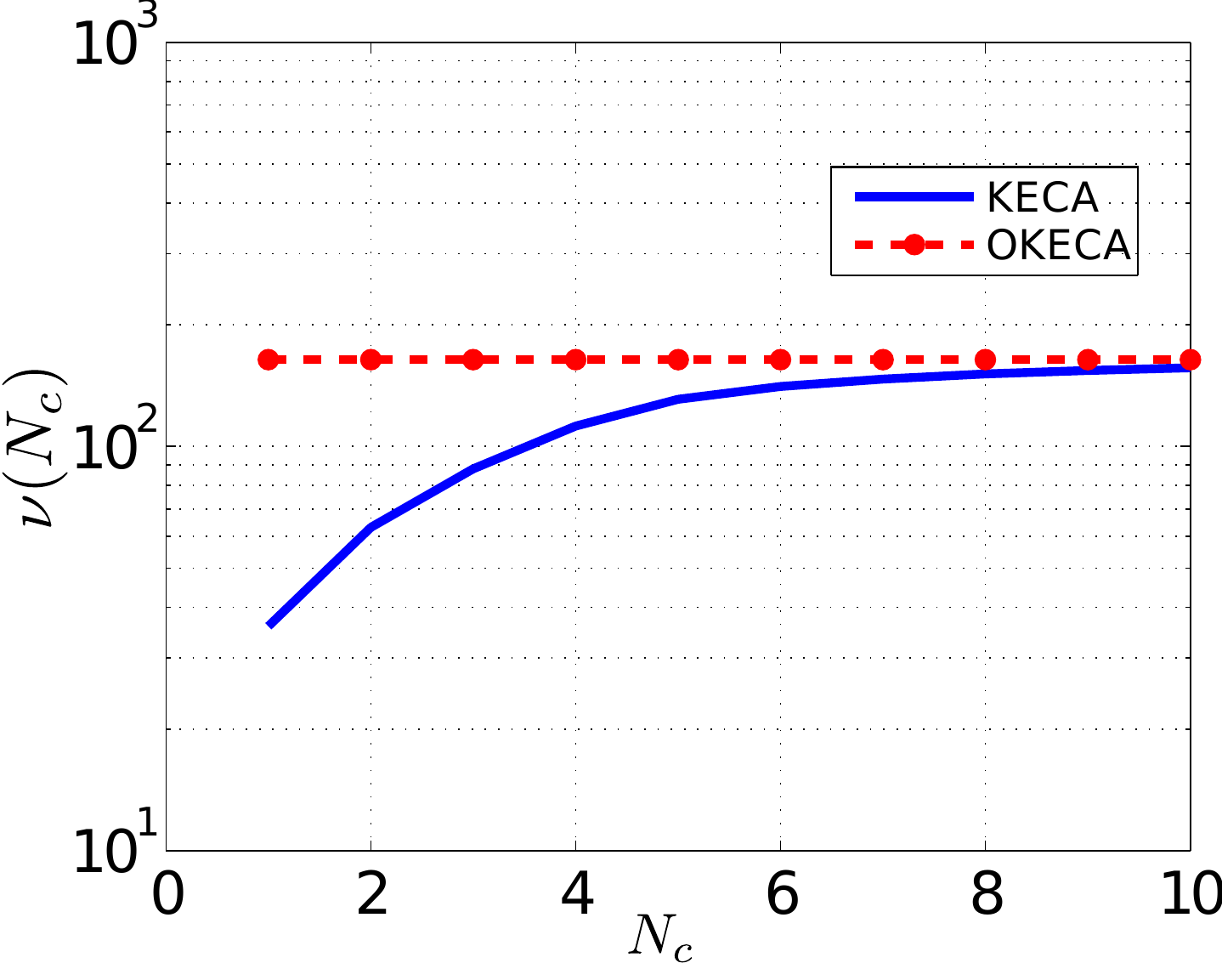} & \includegraphics[width=3.5cm]{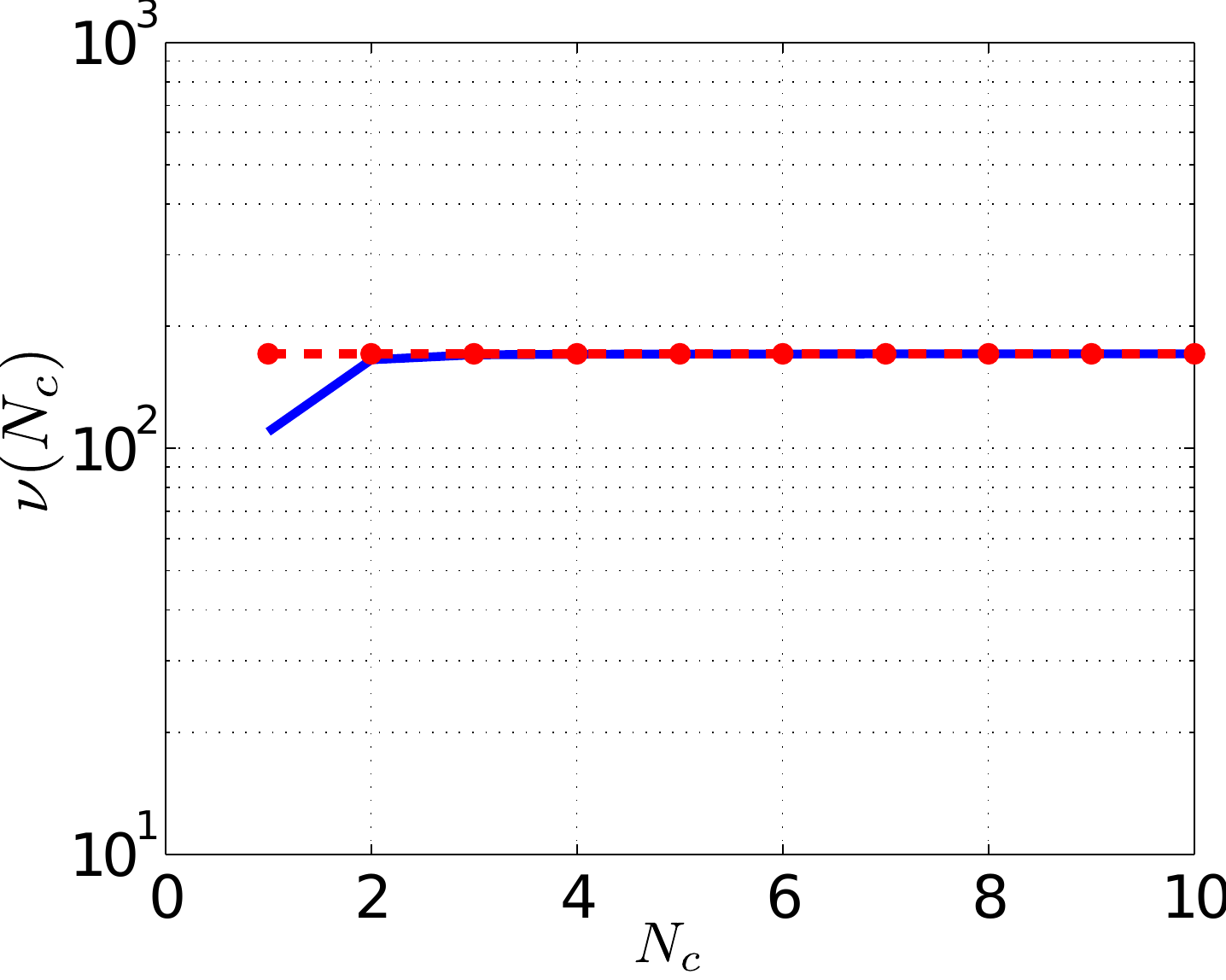} & \includegraphics[width=3.5cm]{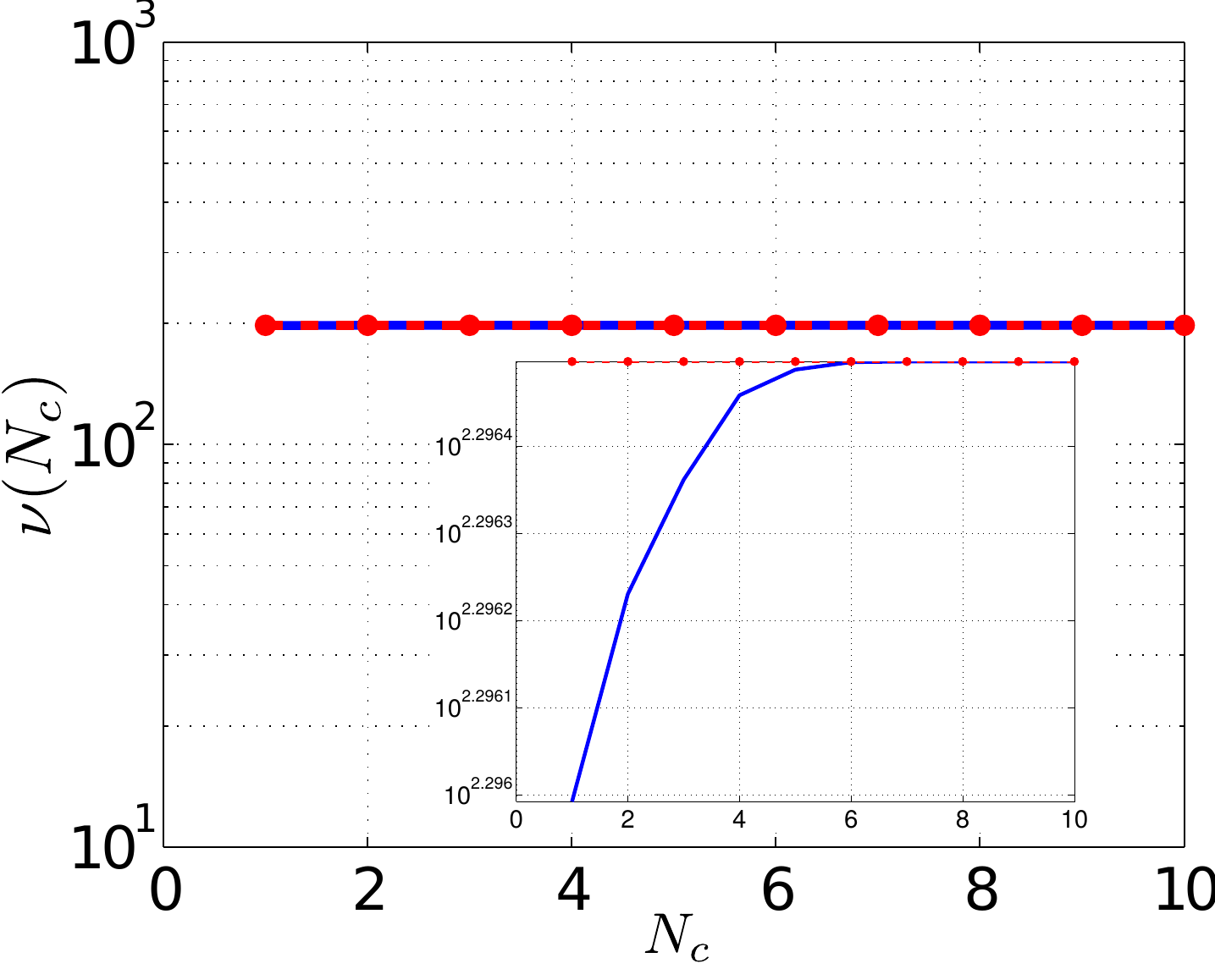} & \includegraphics[width=3.5cm]{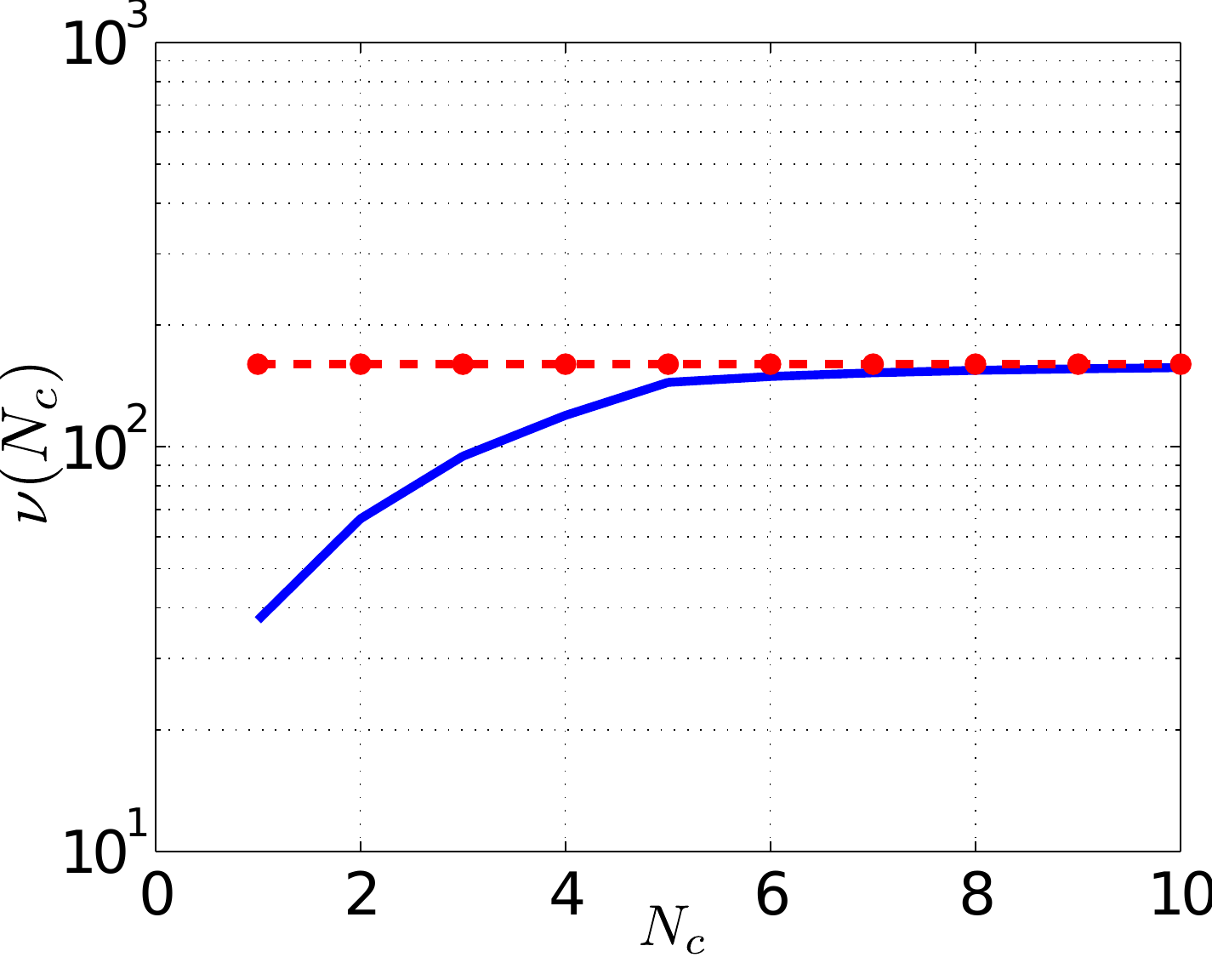}  \\
  \includegraphics[width = 2.8cm]{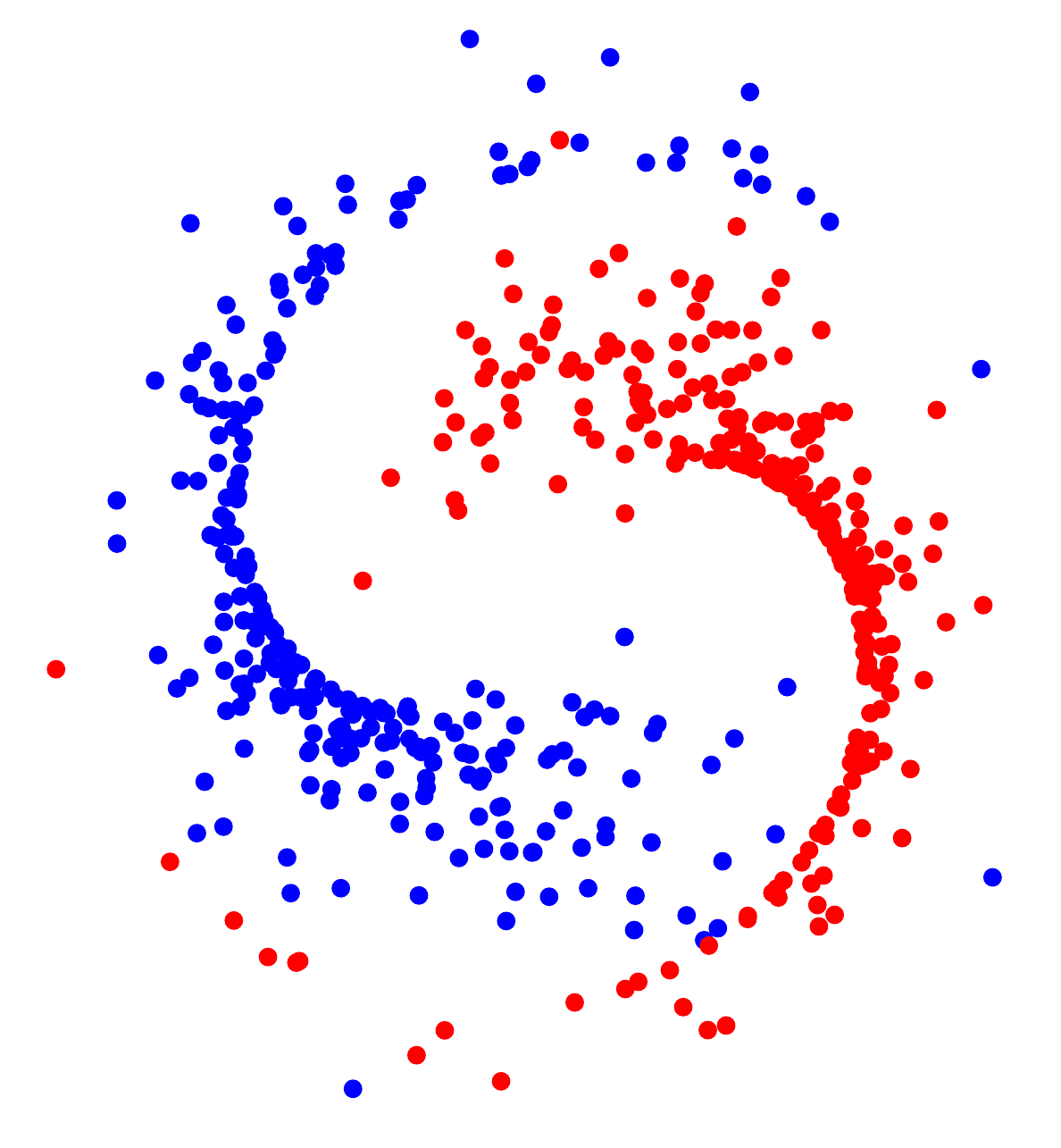} & \includegraphics[width=3.5cm]{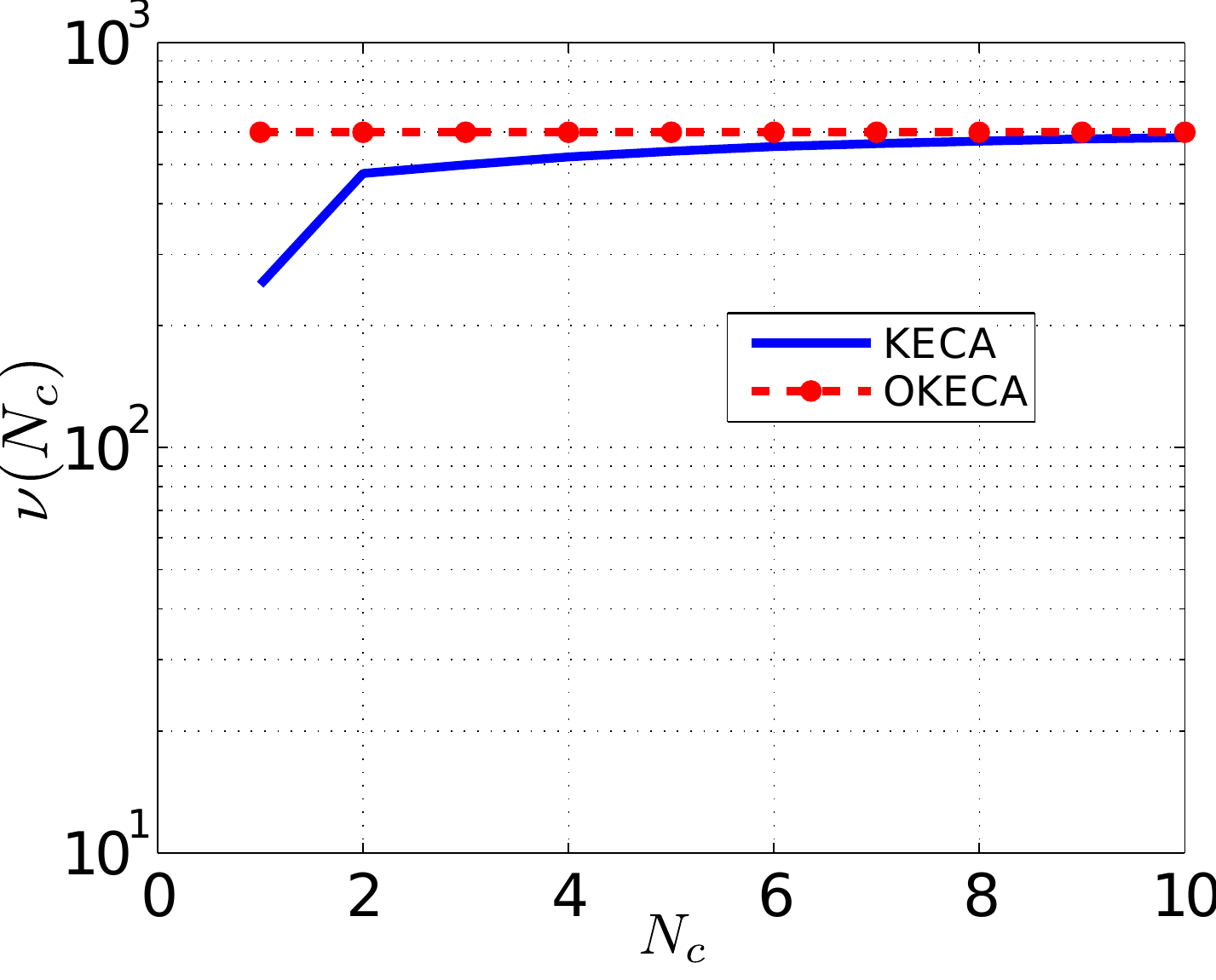} & \includegraphics[width=3.5cm]{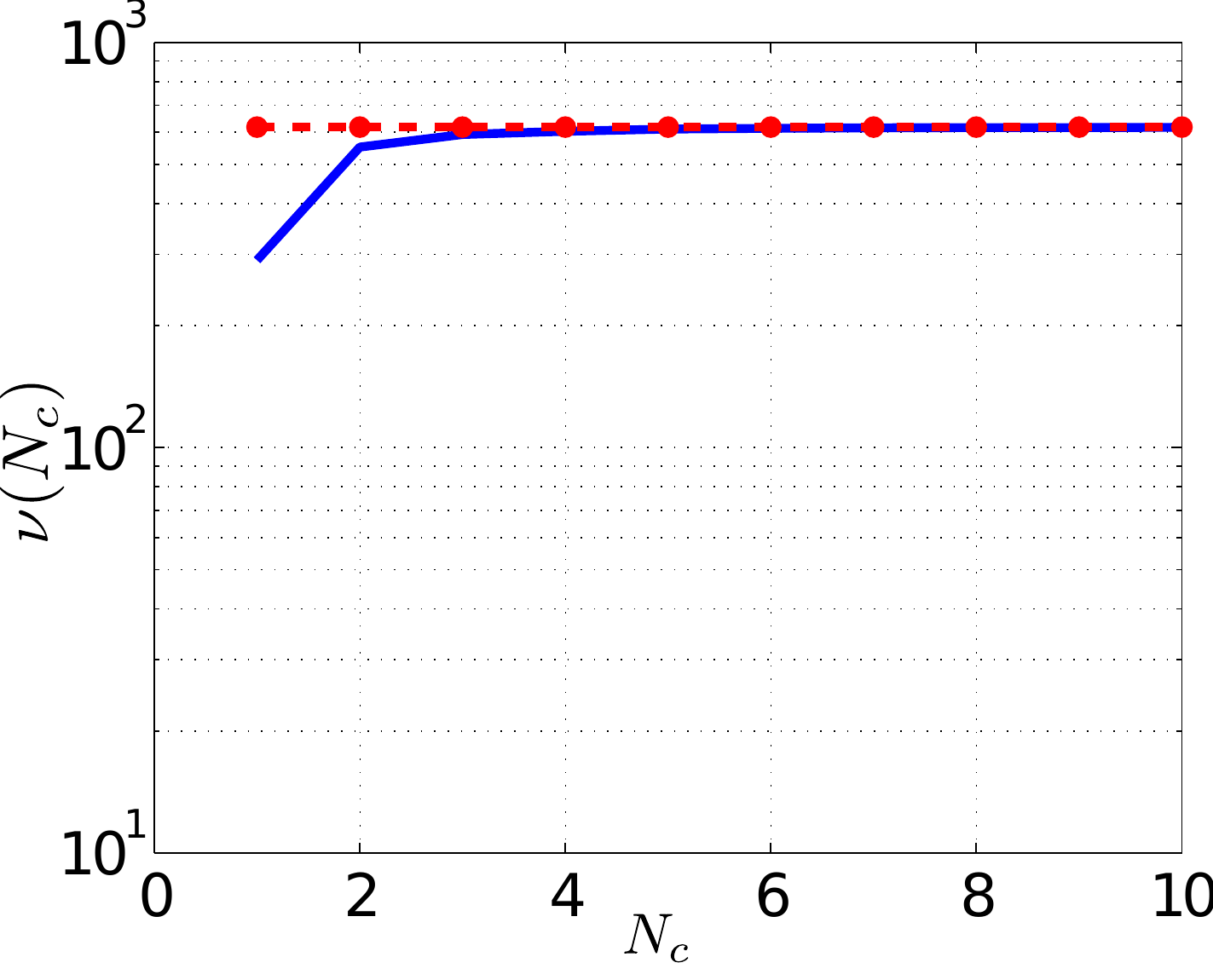} & \includegraphics[width=3.5cm]{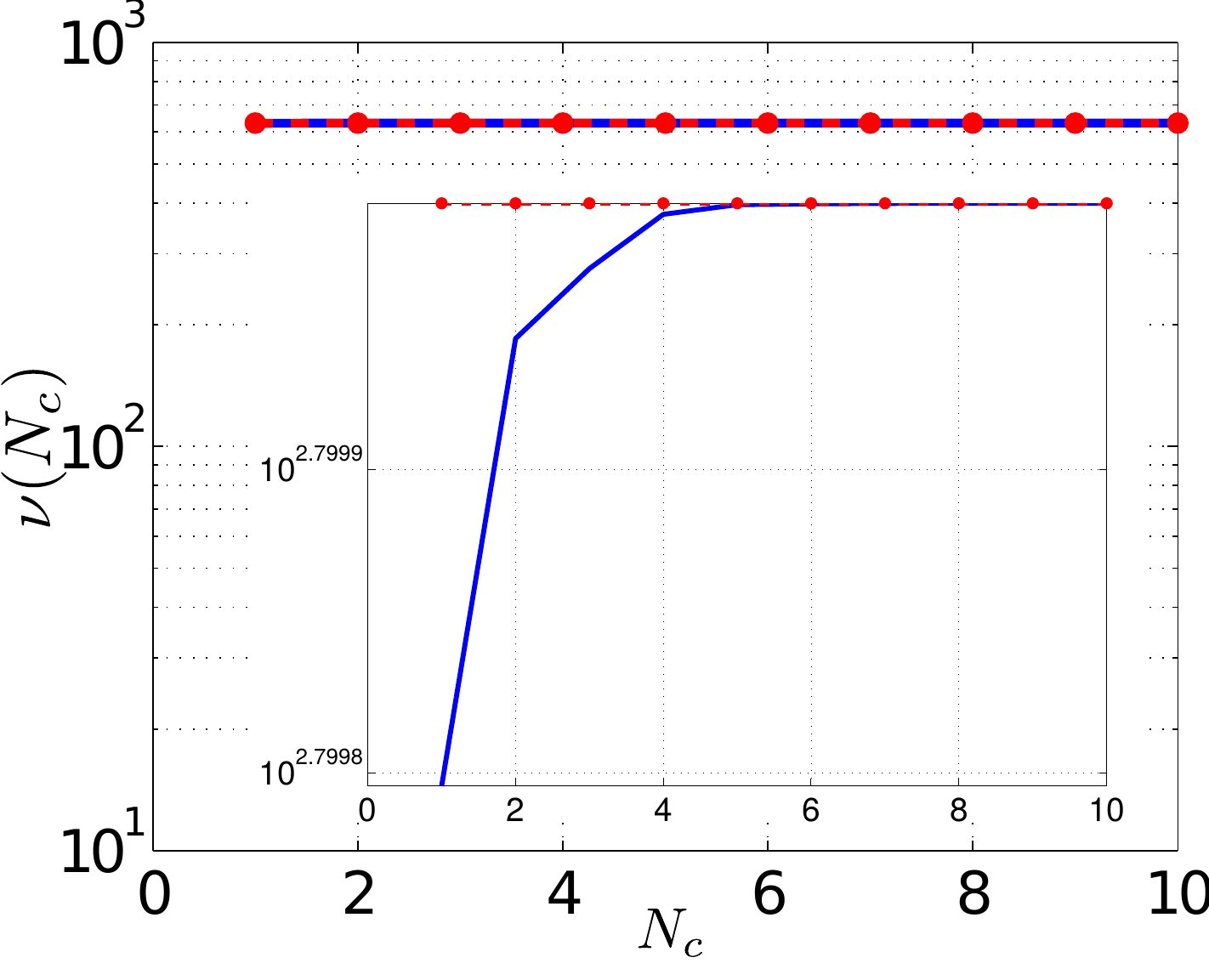} & \includegraphics[width=3.5cm]{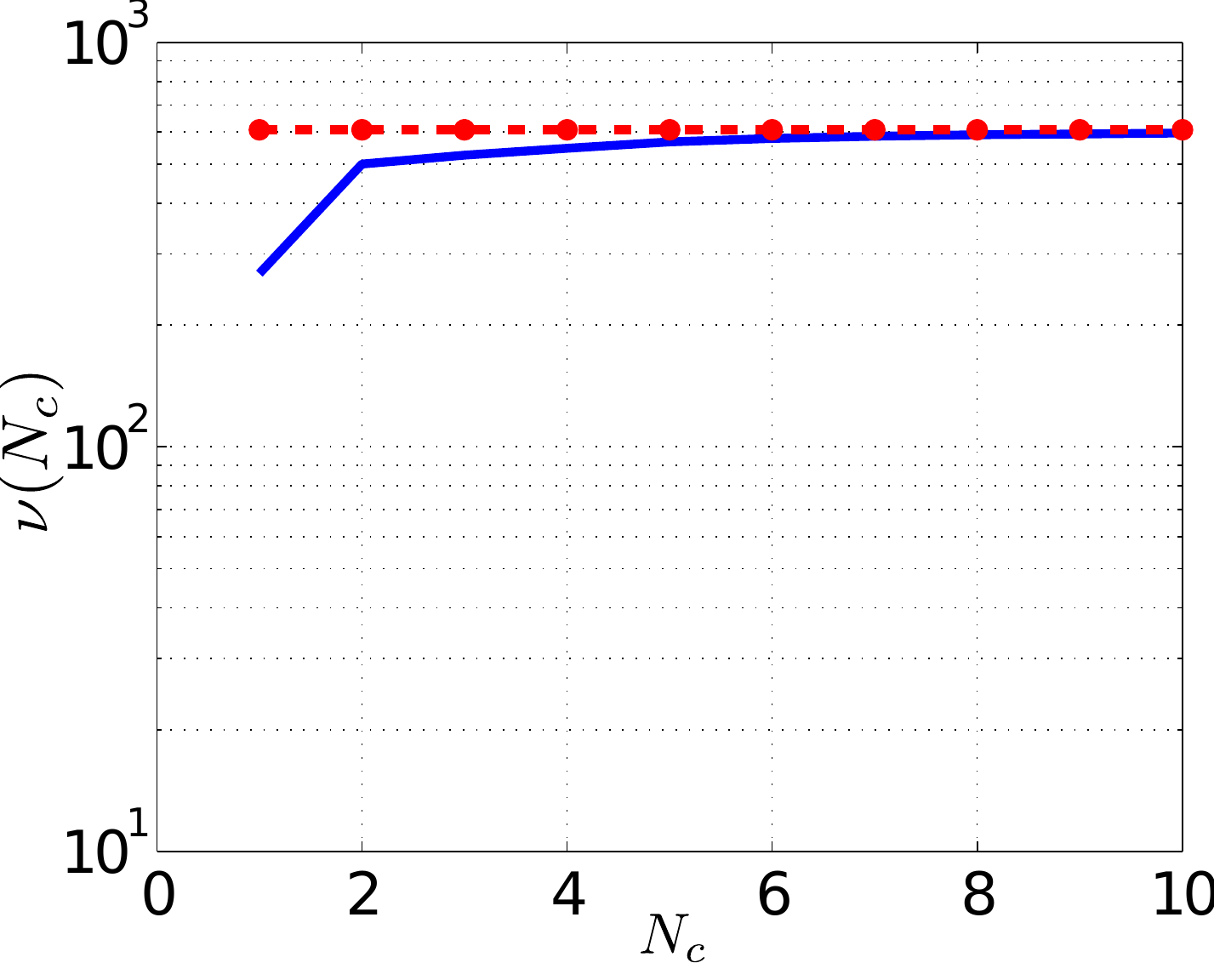}  \\
  \end{tabular}
  \end{center}
  \caption{\label{fig:Eigenvalues} Cumulative estimated information potential ($\v(N_c)$) versus the number of dimension components for different toy datasets, using KECA and OKECA and different $\sigma$ estimation approaches. }
\end{figure*}

\begin{algorithm}[t!]
\begin{algorithmic}[1] 
\INPUT $\K$ 
\OUTPUT $\B$, $\W$
\STATE $ [\E, \D] \leftarrow$ eig of $\K$
\STATE Initialize $\tau$, $\W$
\FOR {$t$ iterations}
\STATE ${\bf dJ} \leftarrow 2 ({\mathbf 1}_{n}^{\top} \E \D^{\frac{1}{2}} \w_k) ({\mathbf 1}_{n}^{\top} \E \D^{\frac{1}{2}})^{\top}$
\STATE $\w_k(t+1)  \leftarrow \w_k(t) + \tau~{\bf dJ}$
\STATE $\E\leftarrow (\E \D^{\frac{1}{2}})~\w_k(t+1) $
\ENDFOR
\STATE $\B \leftarrow \E \D^{\frac{1}{2}}\W$
\STATE $\nu \leftarrow \sum_j (\lambda_j^{\frac{1}{2}} {\mathbf 1}^{\top} \mathbf{e}_j)^2$
\STATE $\B \leftarrow$ sort$_\nu$ $\B$, $\W \leftarrow$ sort$_\nu$ $\W$
\end{algorithmic}
\caption{OKECA feature extraction\label{pckTrain}}
\vspace{0.11cm}
\end{algorithm}

\subsection{Computational cost} 
The proposed method is particularly promising for dimensionality reduction since it compacts the information in very few features with higher expressive power. When using this method as the first step in a data processing pipeline its properties will help to reduce the burden and time computing in the next processing steps. Application of the OKECA method requires just of a kernel generation and a matrix multiplication, as in KPCA and KECA. That means that once it is trained its application to new samples is not demanding. However the training step is more computationally demanding than KECA. While KECA method is based on an eigen-decomposition of a kernel matrix, OKECA additionally requires the gradient ascent  procedure in order to refine the obtained features. In particular, the KECA method requires $O(n^3)$, while OKECA spends $O(n^3 + 4tn^2)$, where $n$ is the number of samples and $t$ is the number of iterations of the gradient ascent process. Note that the number of iterations to converge $t$ depends on the particular problem at hand, but importantly, this just has to be computed once. After the kernel is optimized entropic decomposition is learned, the application is straightforward and as demanding as the KPCA and KECA counterparts.

\subsection{Kernel decomposition in density estimation}
\label{KDE}

This section illustrates the benefits of using the proposed decomposition for KDE~\cite{Parzen62}. KDE is a classical method for estimating a PDF in a non-parametric way. Essentially, KDE defines the PDF as a sum of kernel functions, $k(\cdot,\x_i)$, defined over the training samples $\x_i$ of the dataset ${\mathcal D}$ as follows:
\begin{equation}
	\hat{p}(\x_\ast) = \frac{1}{n} \sum_{i=1}^n k(\x_\ast,\x_i).
\label{pdf1}
\end{equation}
As mentioned before, KDE kernel functions need not in general be PSD but have to be nonnegative and integrate to one to ensure that $\hat{p}$ is a valid probability density function. A classical example of such a kernel function is the Gaussian distribution, $k(\x_\ast,\x_i) = (2\pi \sigma^d)^{-1/2} \exp(-\|\x_\ast-\x_i\|^2/(2\sigma^2))$, but as mentioned, other choices exist. Then, the corresponding kernel matrix can be used for KDE
\begin{equation}
	\hat{p}(\x_\ast)= \frac{1}{n} \sum_{i=1}^n k(\x_\ast,\x_i) = \frac{1}{n} {\mathbf 1}_n^{\top} {\bf k}_\ast,
\label{pdf2}
\end{equation}
where ${\bf k}_\ast$ is the vector of kernel evaluations between the point of interest $\x_\ast$ and all samples in the training dataset ${\mathcal D}$.  As explained in \cite{Girolami02}, if the decomposition of the un-centered kernel matrix follows the form $\K = \E \D \E^{\top}$, where $\E$ is orthonormal and $\D$ is a diagonal matrix, then the kernel-based density estimation may be expressed as
\begin{equation}
	\hat{p}(\x_\ast)={\mathbf 1}_n^{\top} \E_r \E_r^{\top}{\bf k}_\ast,
\label{pdf3}
\end{equation}
where $\E_r$ is the reduced version of $\E$ by keeping columns for $r<n$. Note that when using $\E$ instead of $\E_r$, Eq. (\ref{pdf3}) reduces to Eq. (\ref{pdf2}). This shows that retained KECA components may be used for KDE \cite{Girolami02},  by selecting the dimensions that maximize the information potential in Eq.\ (\ref{eq:KECA_entropy}).

A novel aspect of this paper is to use the OKECA components for KDE in a similar manner. Note that the KECA decomposition 
in Eq.\ (\ref{eq:KECA_decomposition})
is not exactly the same as the proposed OKECA in Eq. (\ref{eq:KECA_new_decomposition}). Nevertheless, it is easy to find a basis that fulfills the same decomposition form, i.e. $\widetilde{\E}\widetilde{\D}=\B$, where $\widetilde{\E}$ is the $\B$ matrix with normalized column vectors and $\widetilde{\D}$ is diagonal matrix containing the norms of each column in $\B$. Therefore, the eq.~(\ref{pdf3}) is $\hat{p}(\x_\ast)={\mathbf 1}_n^{\top} \widetilde{\E}_r \widetilde{\E}_r^{\top}{\bf k}_\ast$. 
In Sec.\ \ref{PDF_est_examples}, we will empirically show that, even though the OKECA basis is not orthonormal in principle, it is possible to obtain an accurate PDF estimation.

\subsection{Model estimation}
\label{Model_estimation}

In this work, we consider the Gaussian RBF kernel since it is the most common in both RKHS kernel methods and KDE~\cite{Parzen62}.  This kernel induces a probabilistic Gaussian mixture model, and it only introduces one scalar free parameter, $\sigma$. Note that more complicated models could be taken into account in both frameworks. However, a recurrent and unsolved problem in both approaches is the estimation of the length-scale parameter $\sigma$.

A plethora of heuristics and rules for estimating the length-scale have been proposed in the machine learning and statistics literatures. 
Roughly speaking, one finds two main approximations. The first approach considers maximizing a particular objective function through a cross-validation procedure. The objective function may be optimized using unsupervised (e.g. maximum likelihood~\cite{Duin76}, denoted by $\sigma_{ML}$ in the experiments) or supervised (e.g. a classification accuracy score, denoted by $\sigma_{class}$ in the experiments) approaches. The second approach resorts to empirical rules of performance or theoretical bounds. Good examples of this second approach, which are considered in this paper, are: 1) the Silverman's rule~\cite{Silverman86}, which is the classical rule of thumb in KDE, $\sigma_{Silv}$ in the experiments; 2) the mean distance between training points, which is a common approach in kernel methods for classification, $\sigma_{d1}$ in the experiments; and 3) the $15\%$ of the median distance between points, which is the classical employed in KECA, $\sigma_{d2}$ in the experiments.

\section{Experiments}\label{experiments}

We compare the performance of the standard KECA and the proposed OKECA for both density estimation and data classification. We analyze the methods in terms of the retained information potential as a function of the extracted features, the impact of the model selection criteria, and the classification accuracies in synthetic and real datasets.

\subsection{OKECA for optimally entropic representations}\label{Entro}

The first experiment considers three well-known 2D toy examples for analyzing the methods: a {\em ring}-shaped distribution consisting of one class only, and the binary {\em two-moons} and {\em pinwheel} datasets. In this section, we illustrate the ability of the proposed method to obtain projections that maximize the information potential, hence minimizing the squared R{\'e}nyi entropy. In the results, we used $80$, $20$, and $45$ training samples for each problem, respectively. 

Figure~\ref{fig:Eigenvalues} shows the original data distributions and the estimated cumulative information potential ($\v$ and $\mathcal{L}$ defined in (\ref{eq:KECA_entropy}) and (\ref{eq:W_optimization}), respectively) attained by KECA and OKECA as a function of the 10 top components and all the considered kernel length-scale selection criteria. 
For all datasets and for all $\sigma$ values, OKECA reaches almost the maximum entropy value with just one feature; whereas KECA cumulative entropy values need five or more components to saturate. 
This behaviour is almost independent of the chosen criterion to set the $\sigma$ parameter. 
The higher information content may translate into more informative features potentially useful for density estimation and classification as we illustrate in the next sections.

\subsection{OKECA for PDF estimation}
\label{PDF_est_examples}

\begin{figure}[h!]
  \centering
  \setlength{\tabcolsep}{1pt}
  \begin{tabular}{c|cccccc}
\hline
PDF & $N_c$ & 1 & 2 & 3 & 4 & 5 \\  
\hline
   $\sigma_{ML}$ & \rota{KECA} &\includegraphics[width=1.3cm]{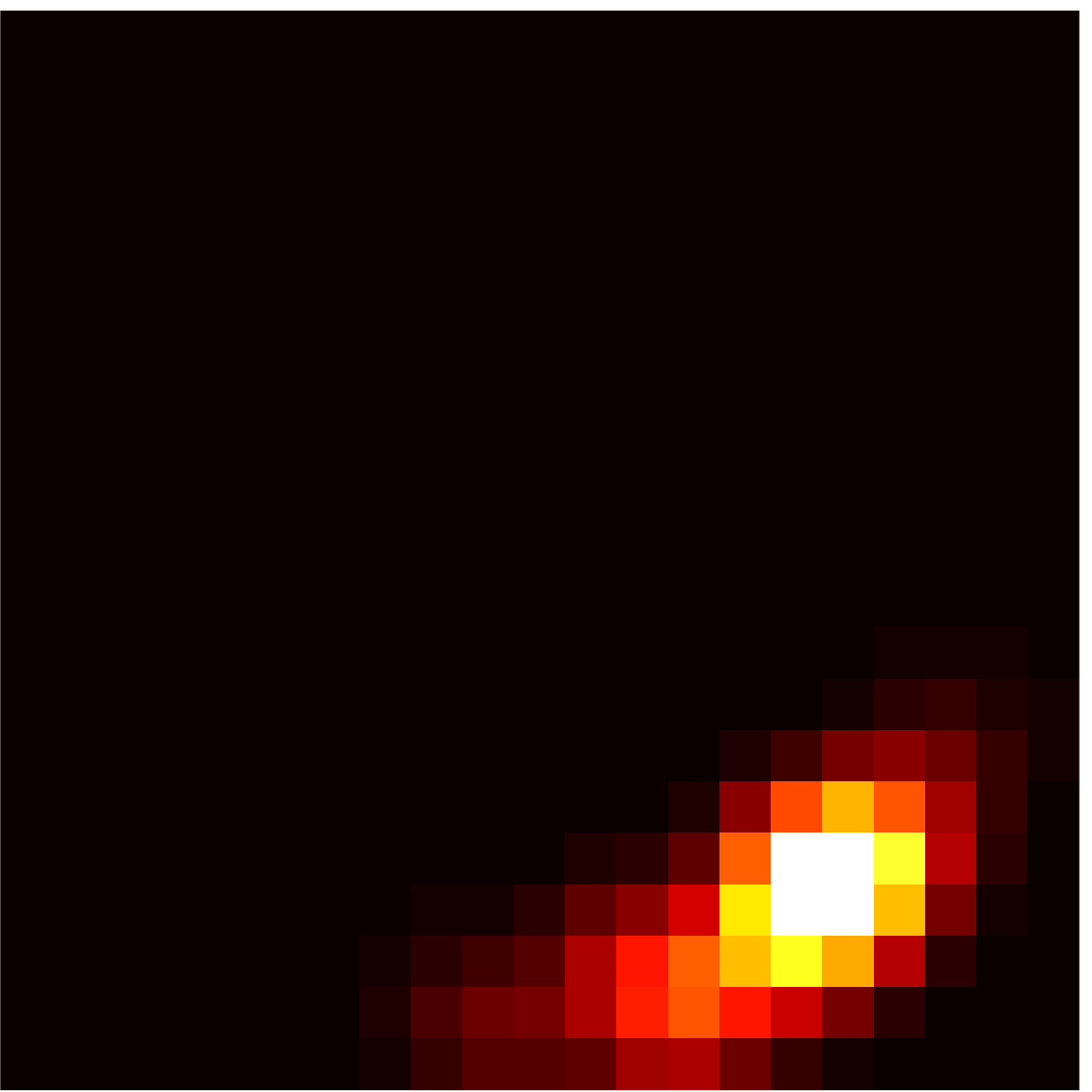} & \includegraphics[width=1.3cm]{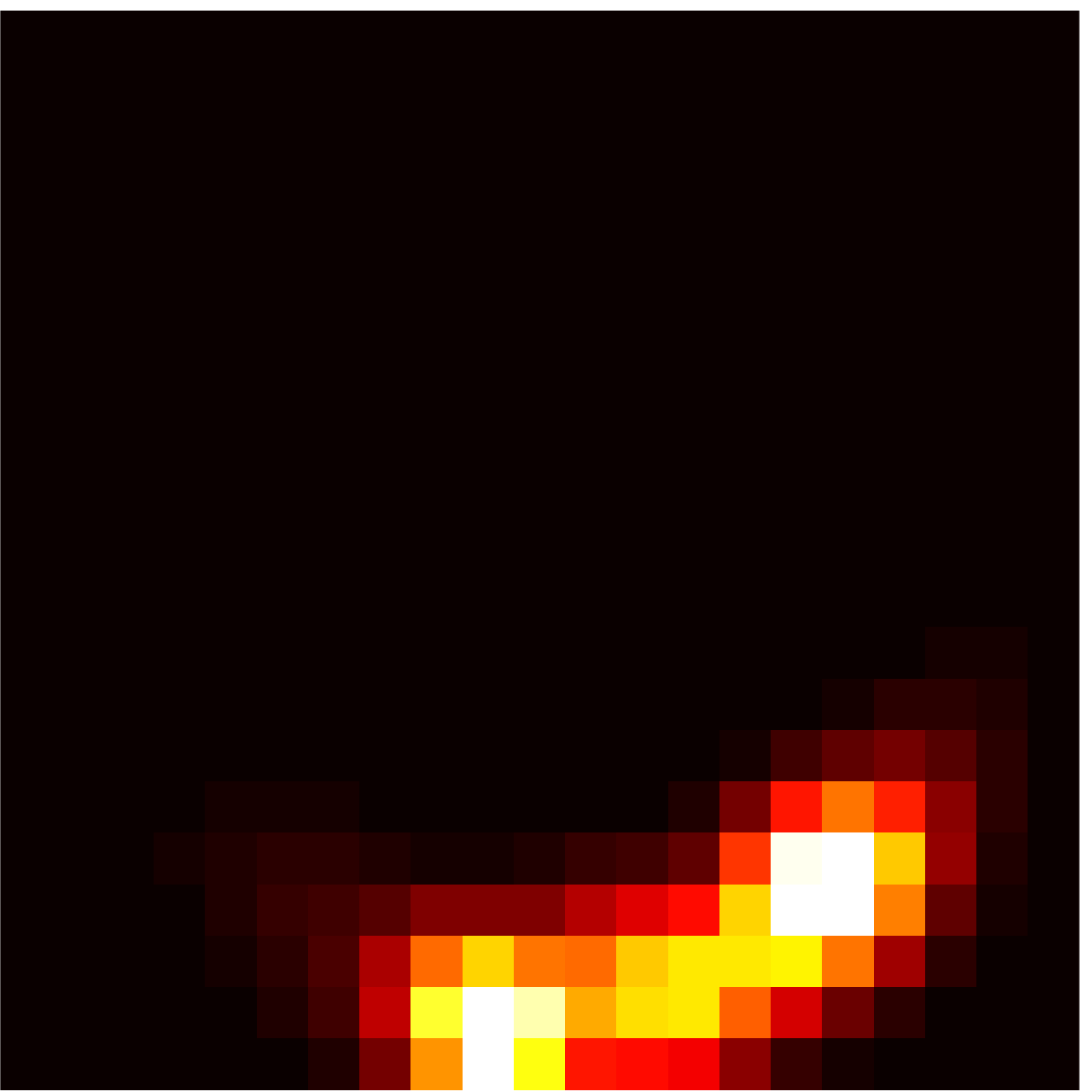} & \includegraphics[width=1.3cm]{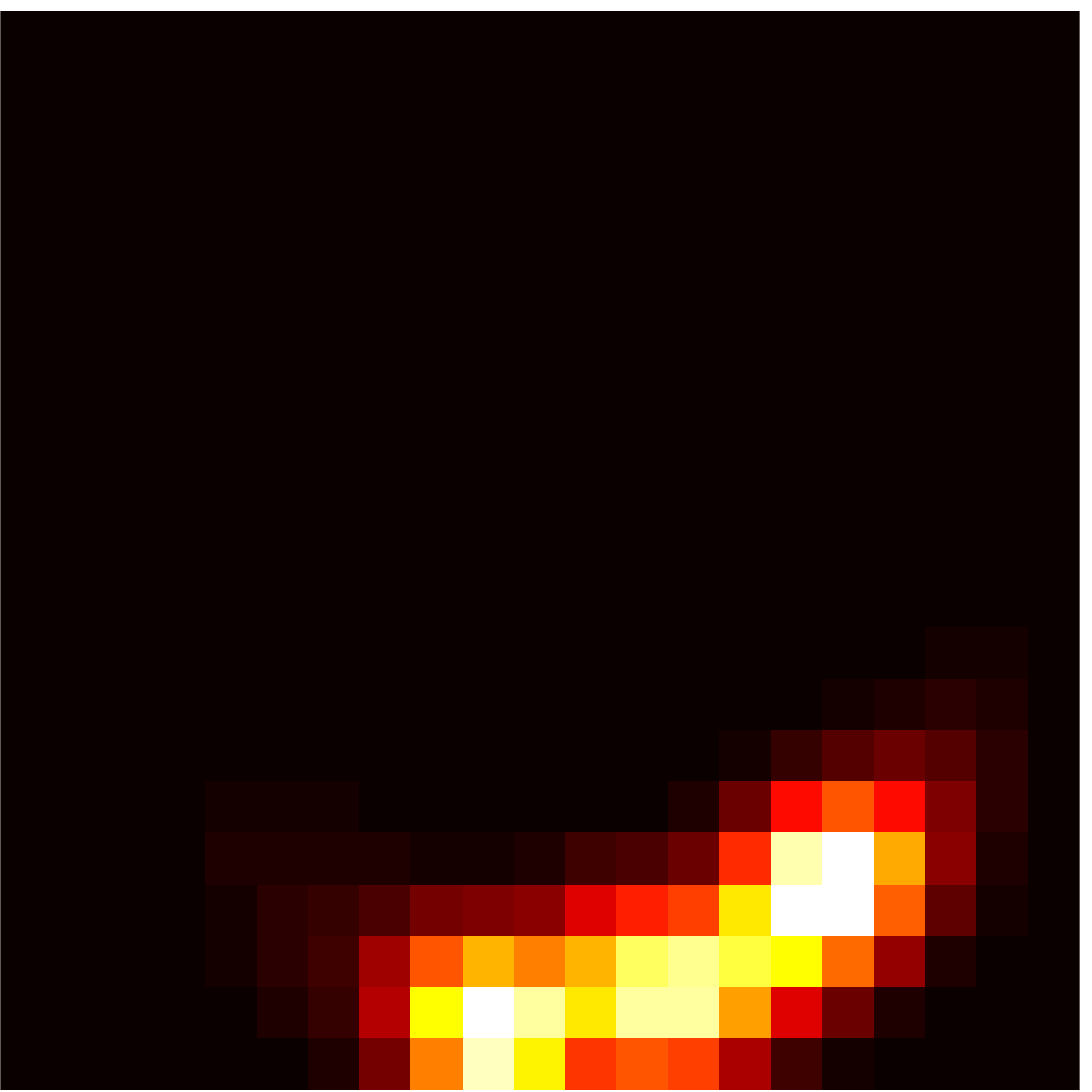} & \includegraphics[width=1.3cm]{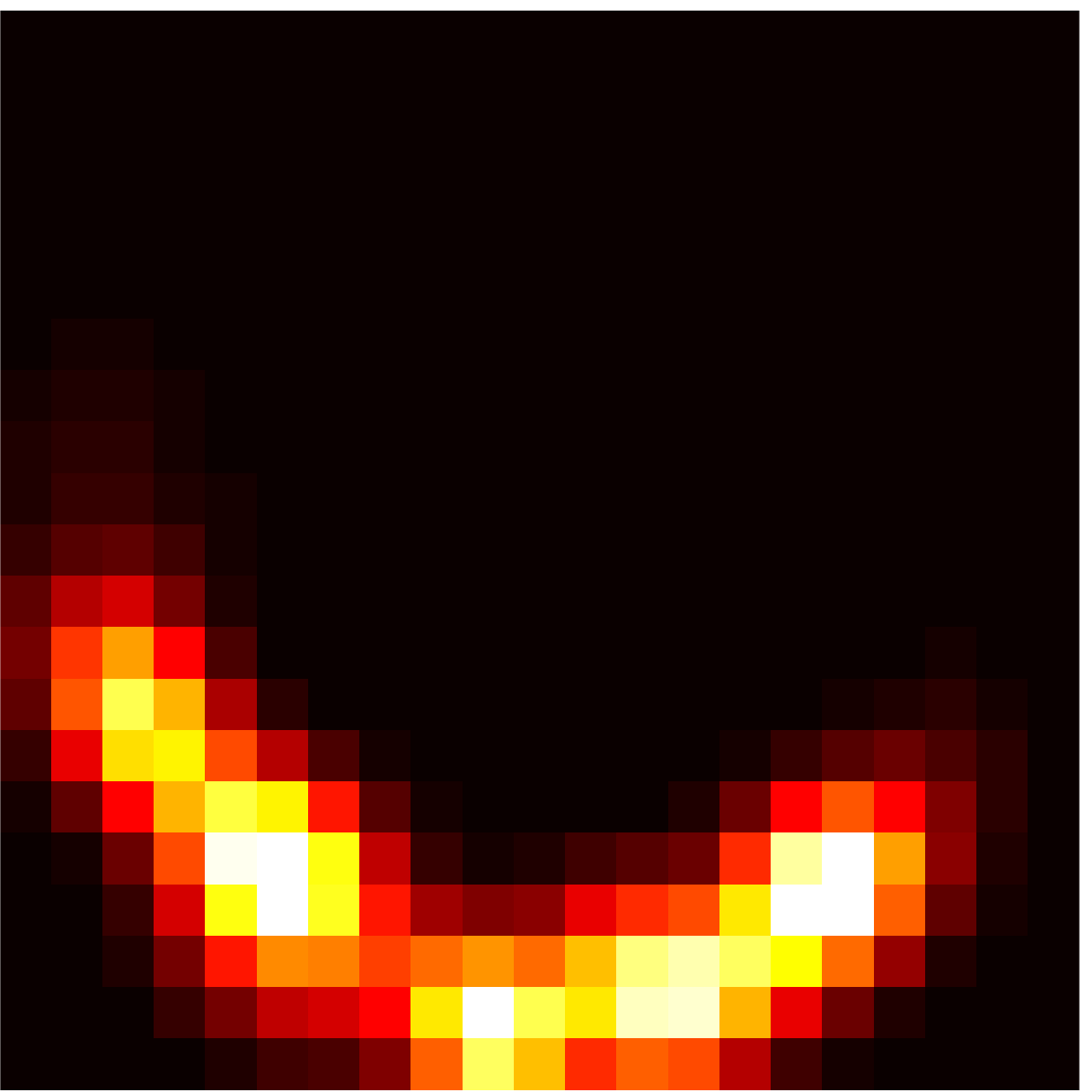} & \includegraphics[width=1.3cm]{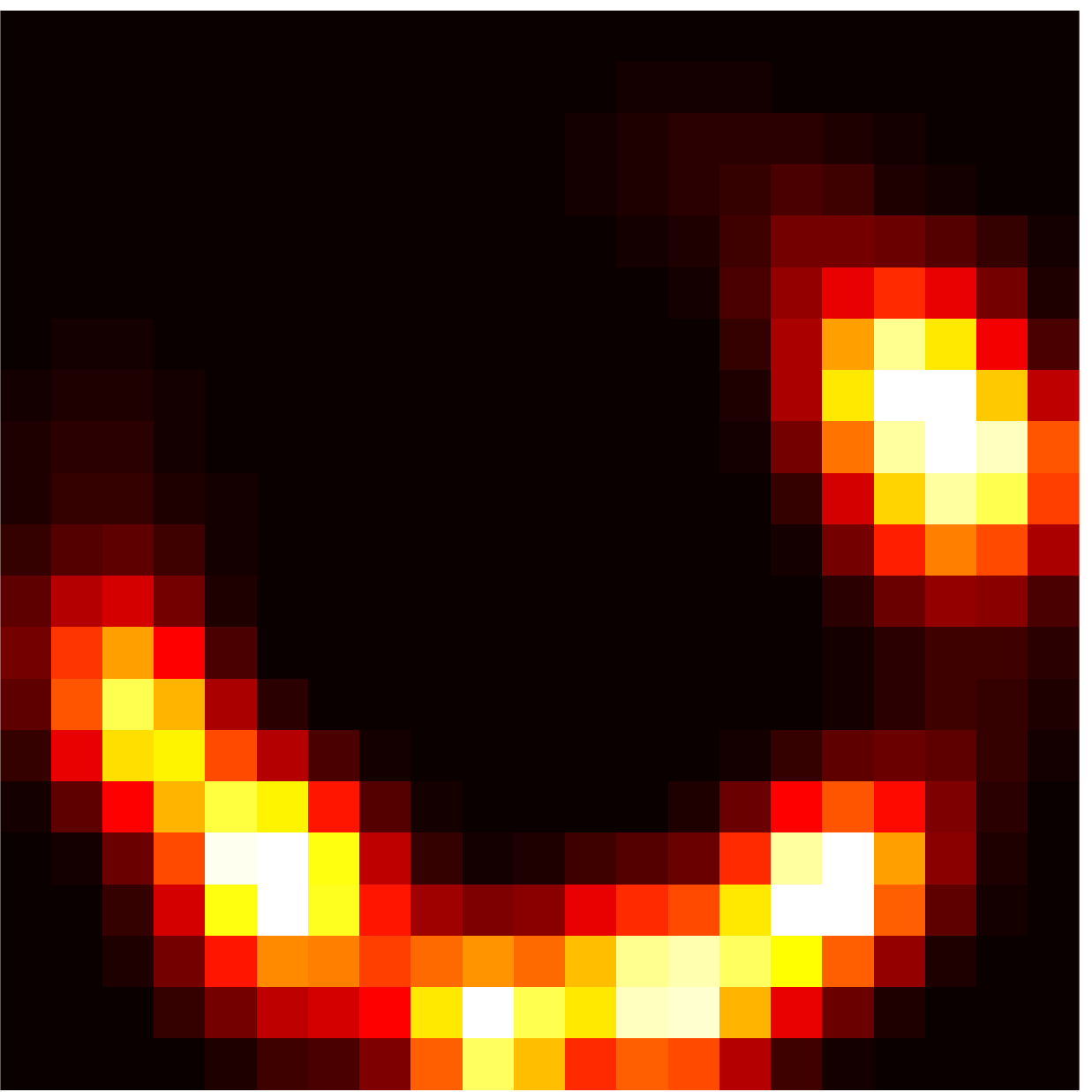} \\ 
\includegraphics[width=1.3cm]{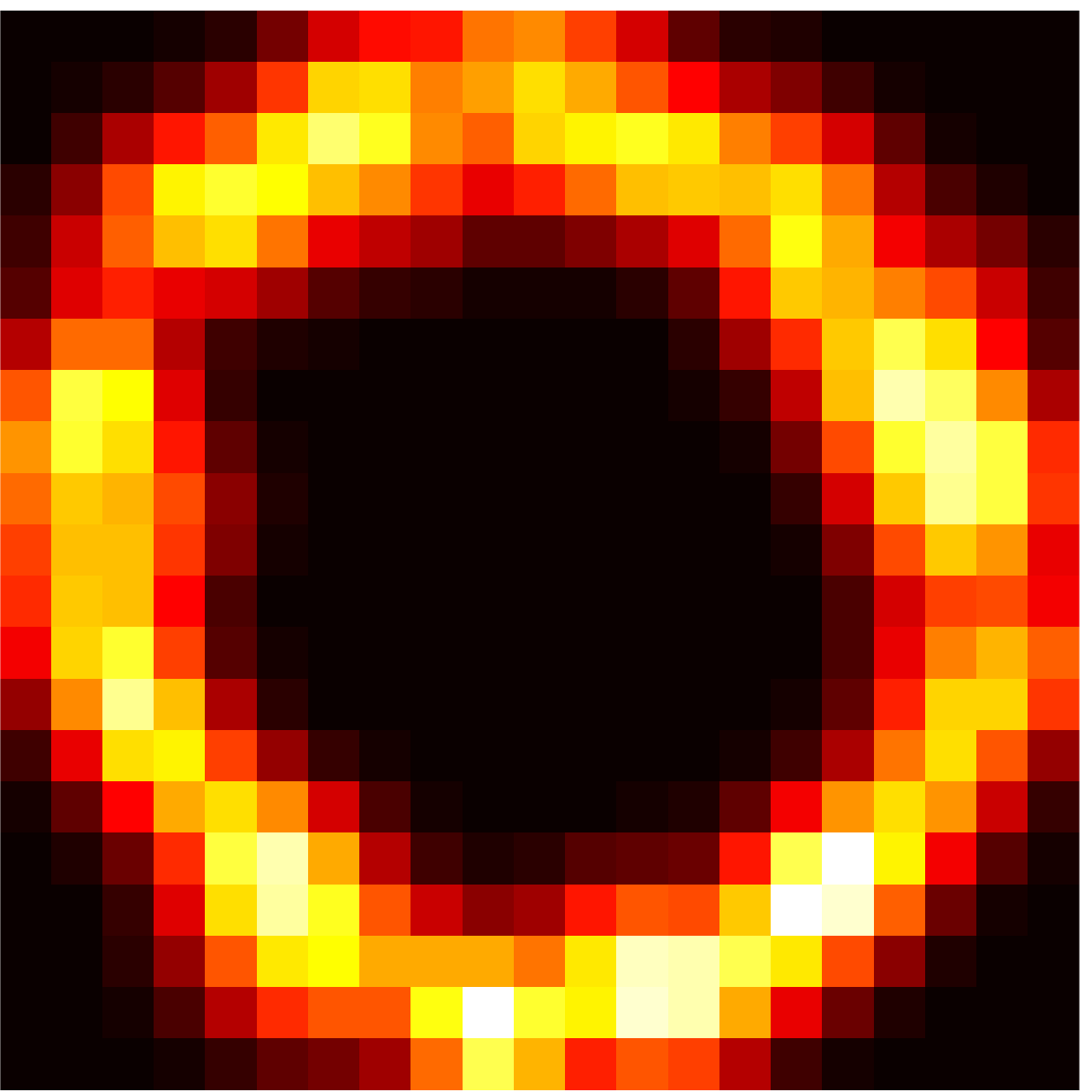} & \rota{OKECA}   & \includegraphics[width=1.3cm]{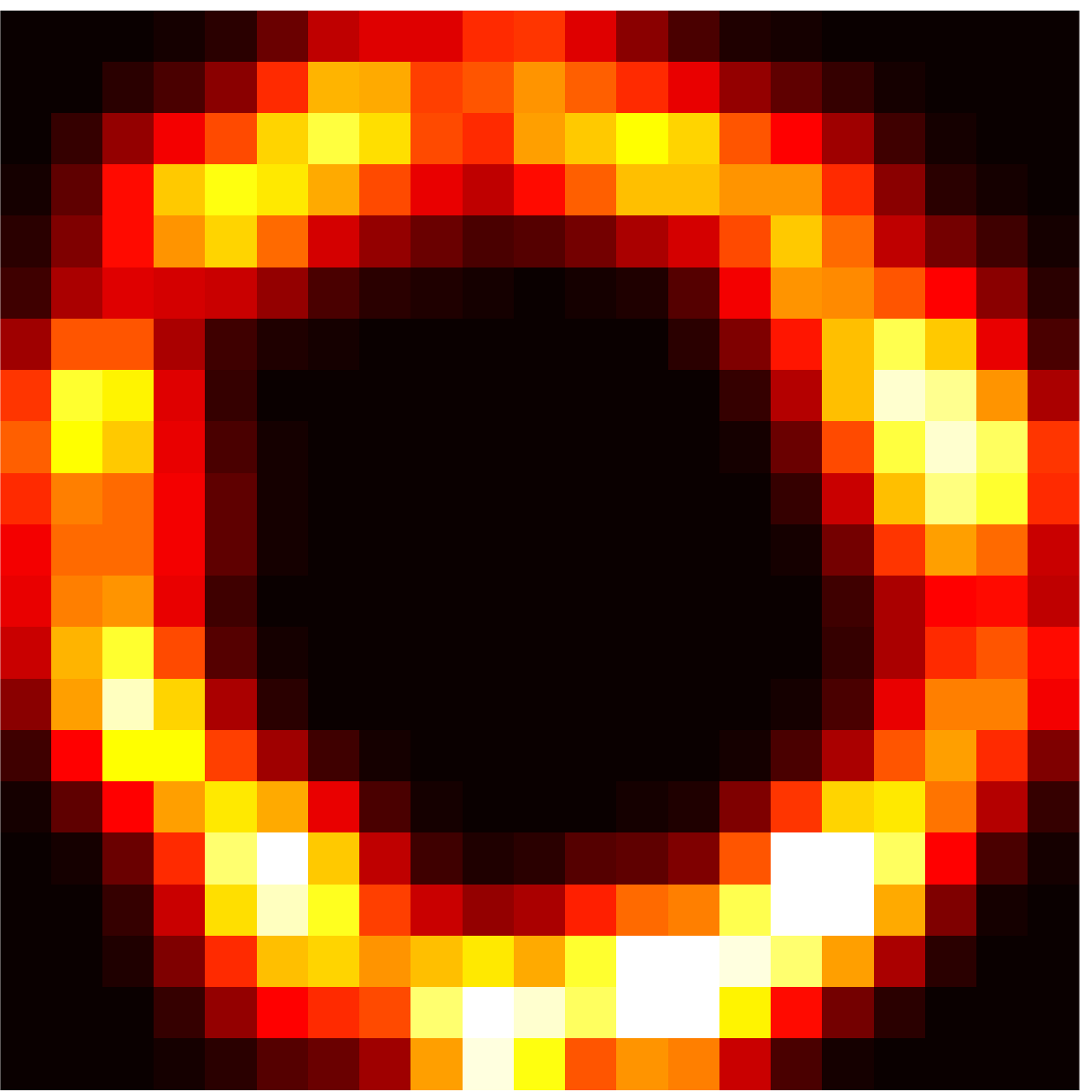} & \includegraphics[width=1.3cm]{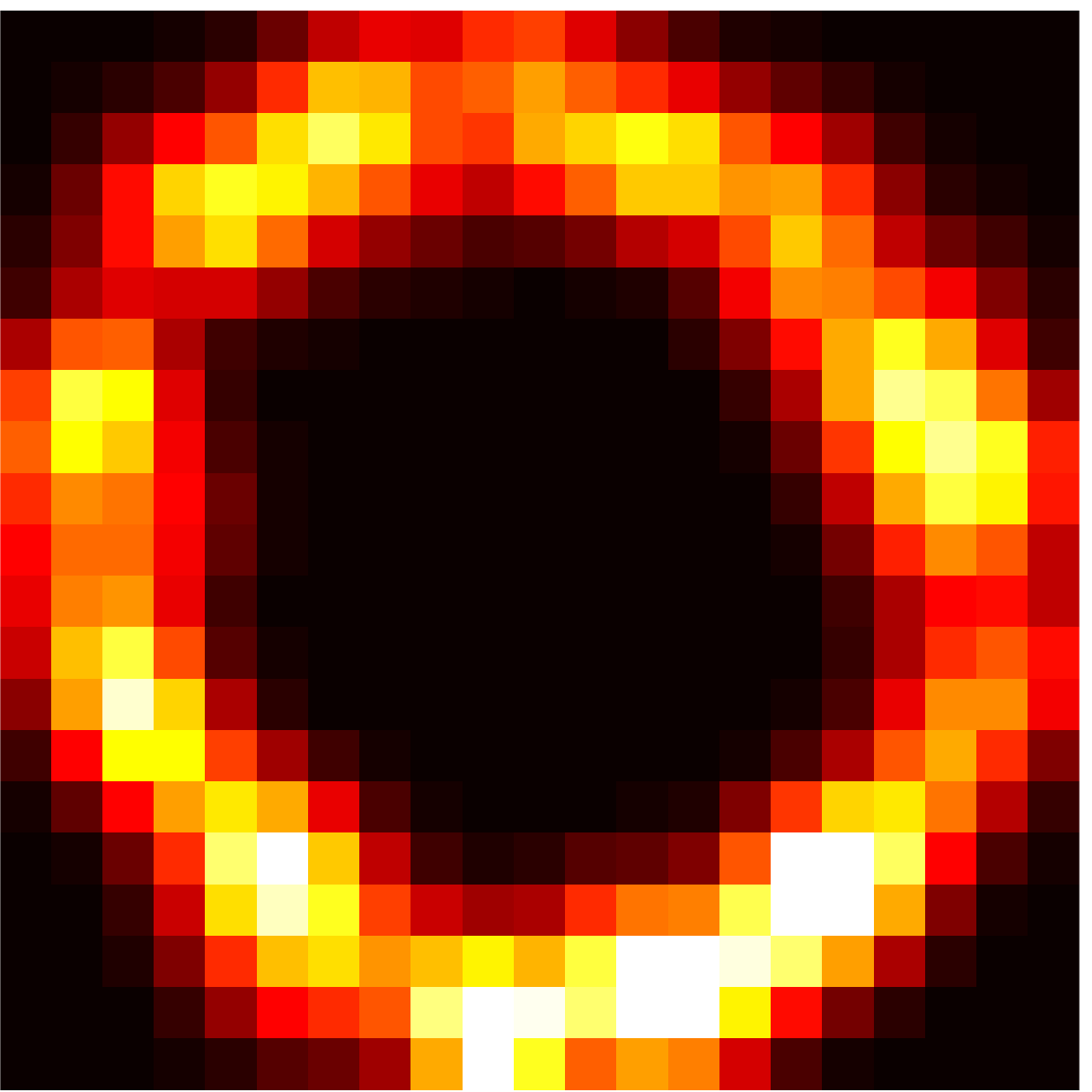} & 
\includegraphics[width=1.3cm]{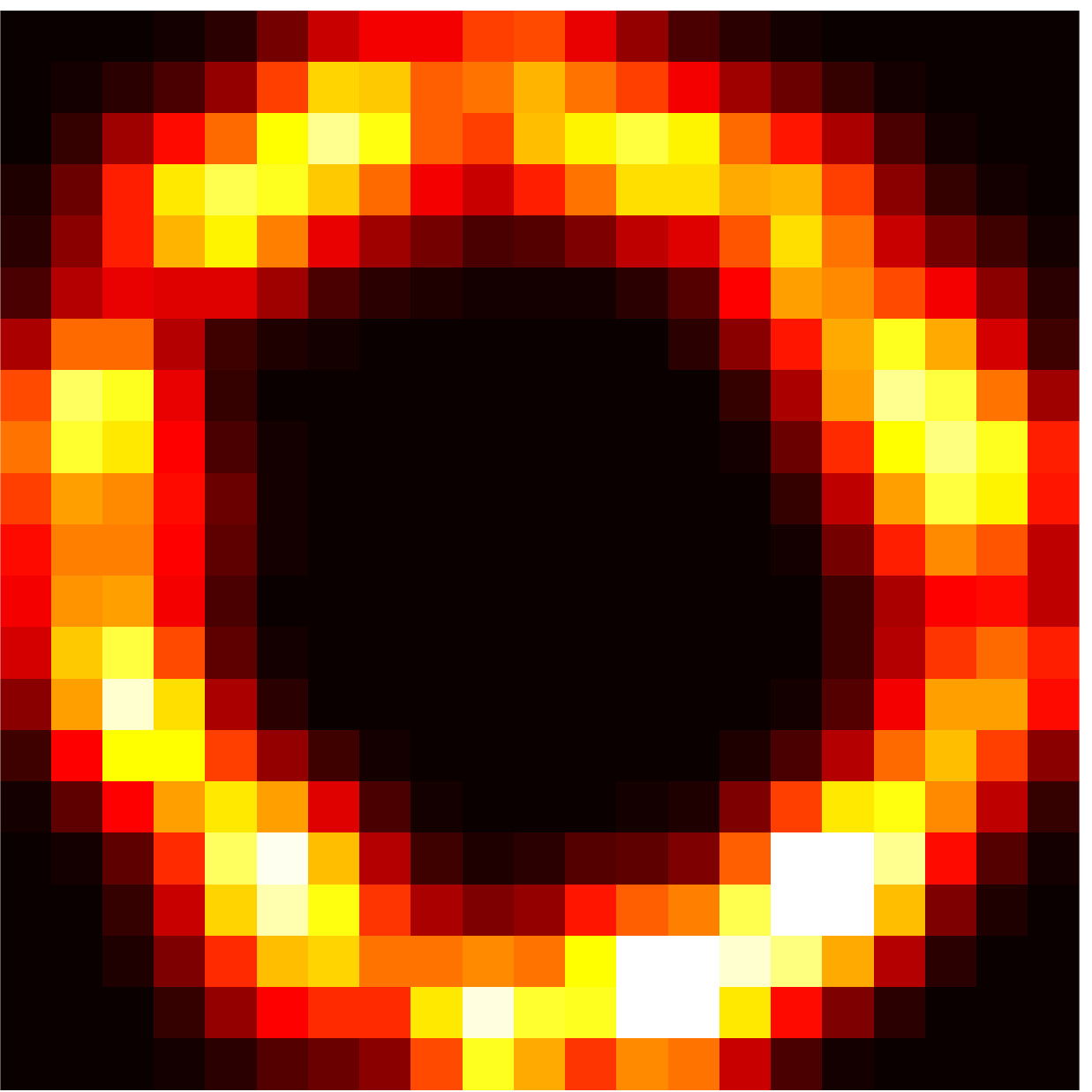} & \includegraphics[width=1.3cm]{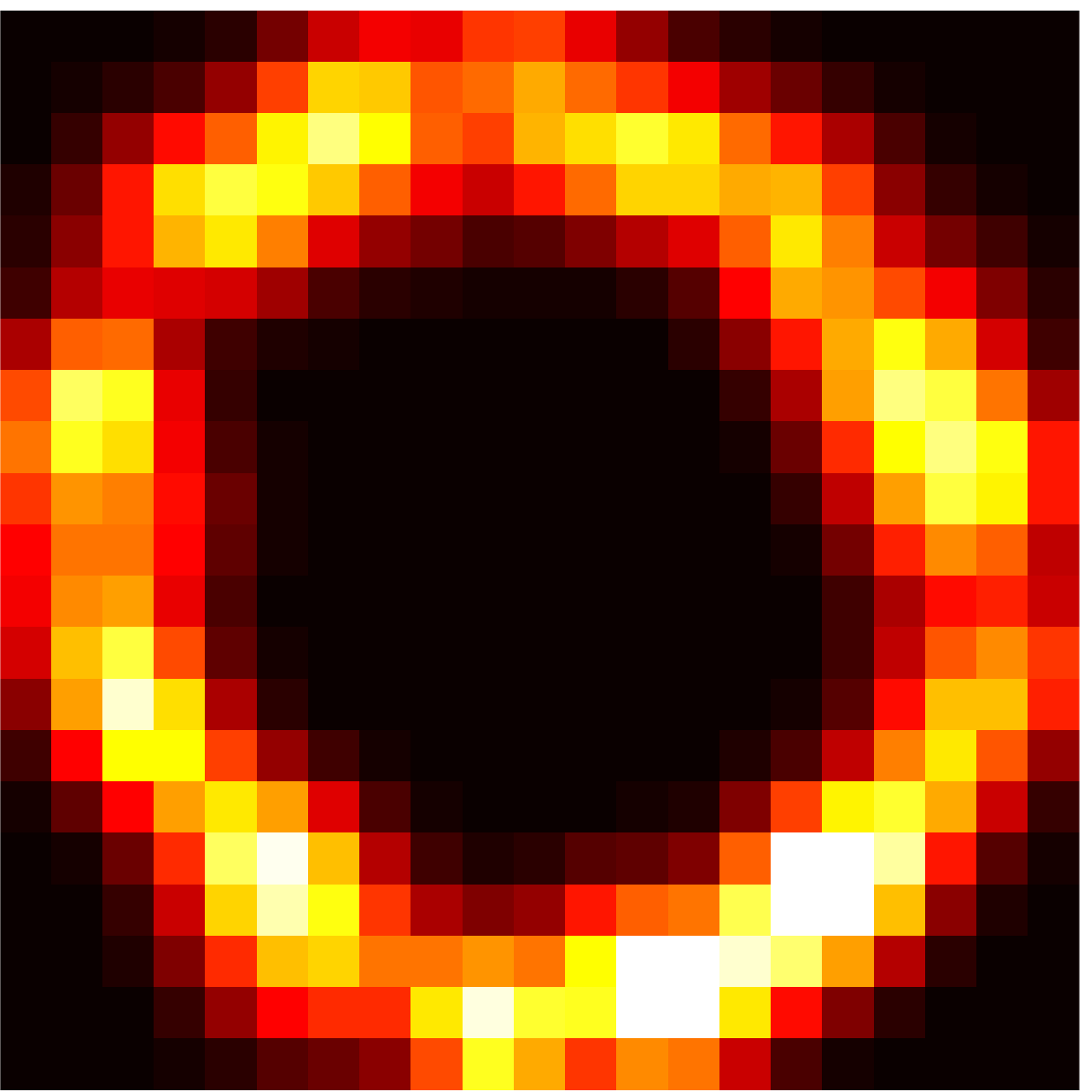} & \includegraphics[width=1.3cm]{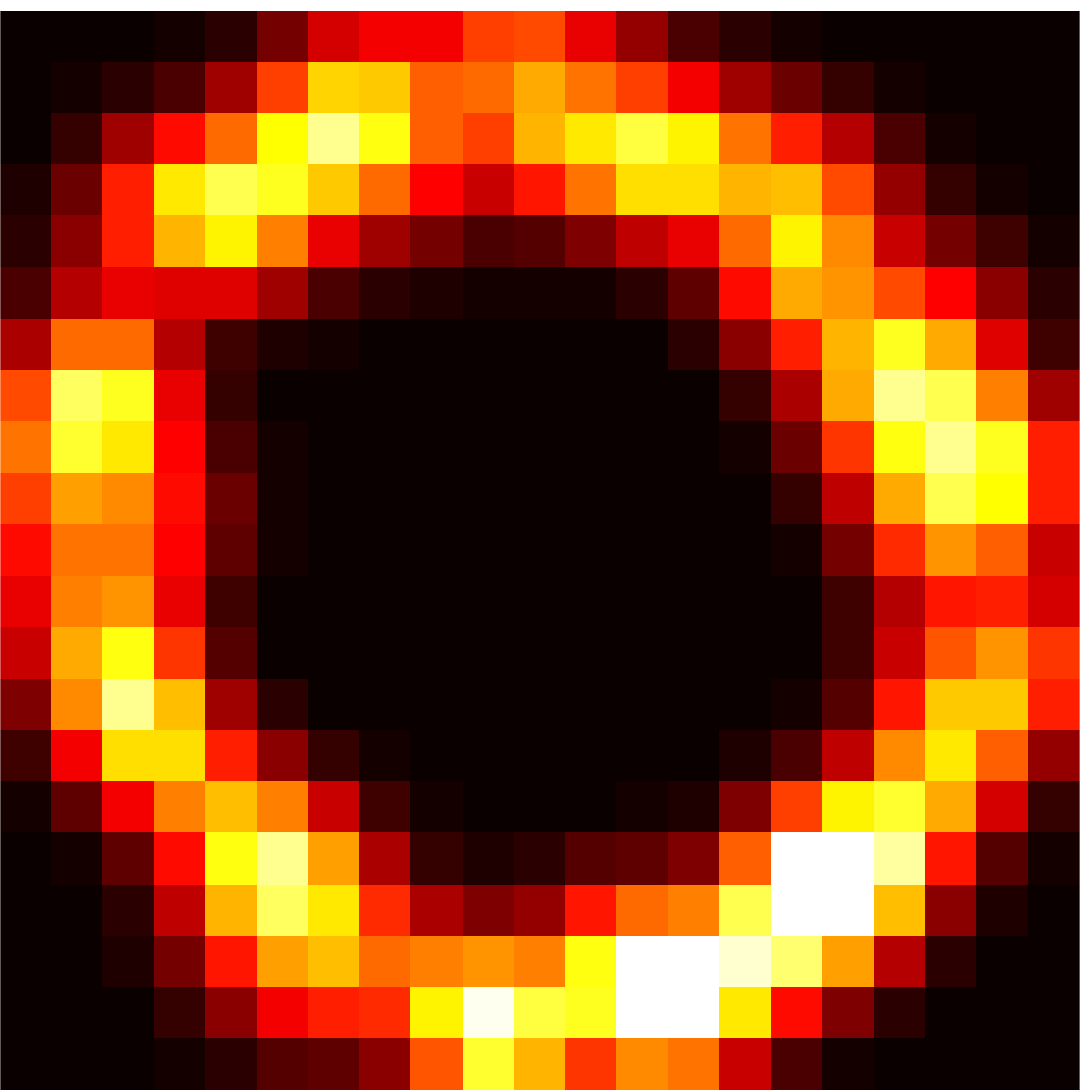}\\ 
\hline
$\sigma_{Silv}$ & \rota{KECA}  &\includegraphics[width=1.3cm]{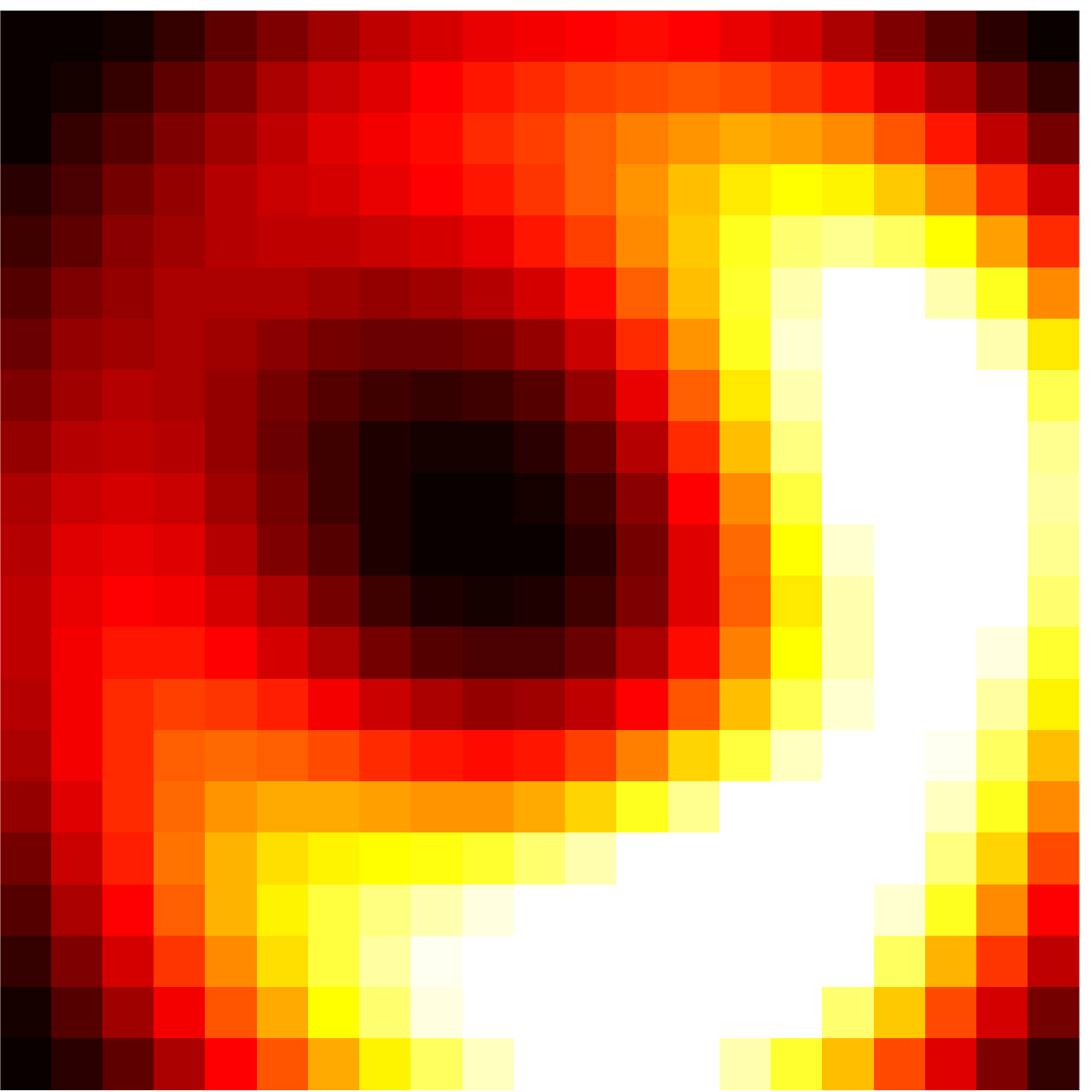} & \includegraphics[width=1.3cm]{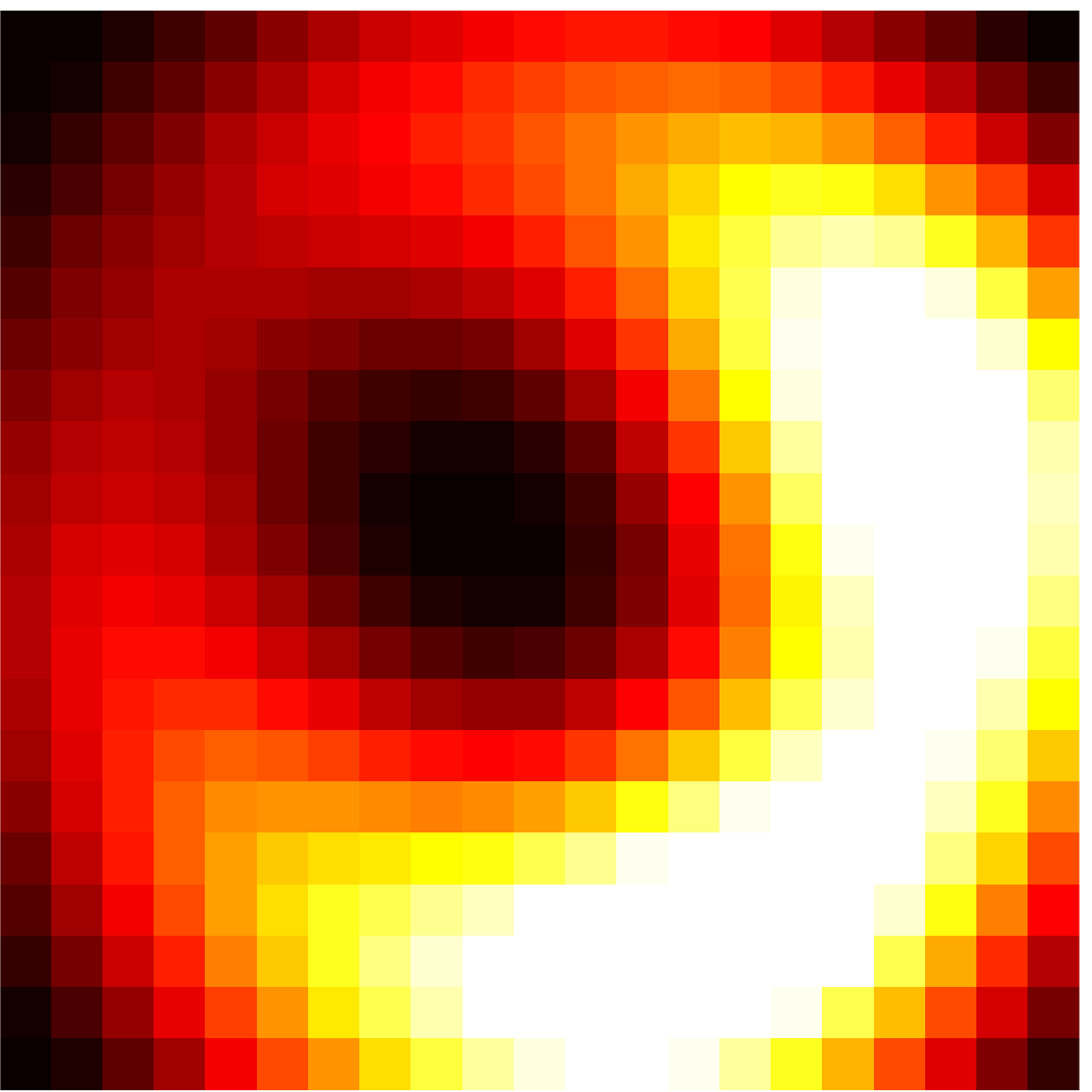} & \includegraphics[width=1.3cm]{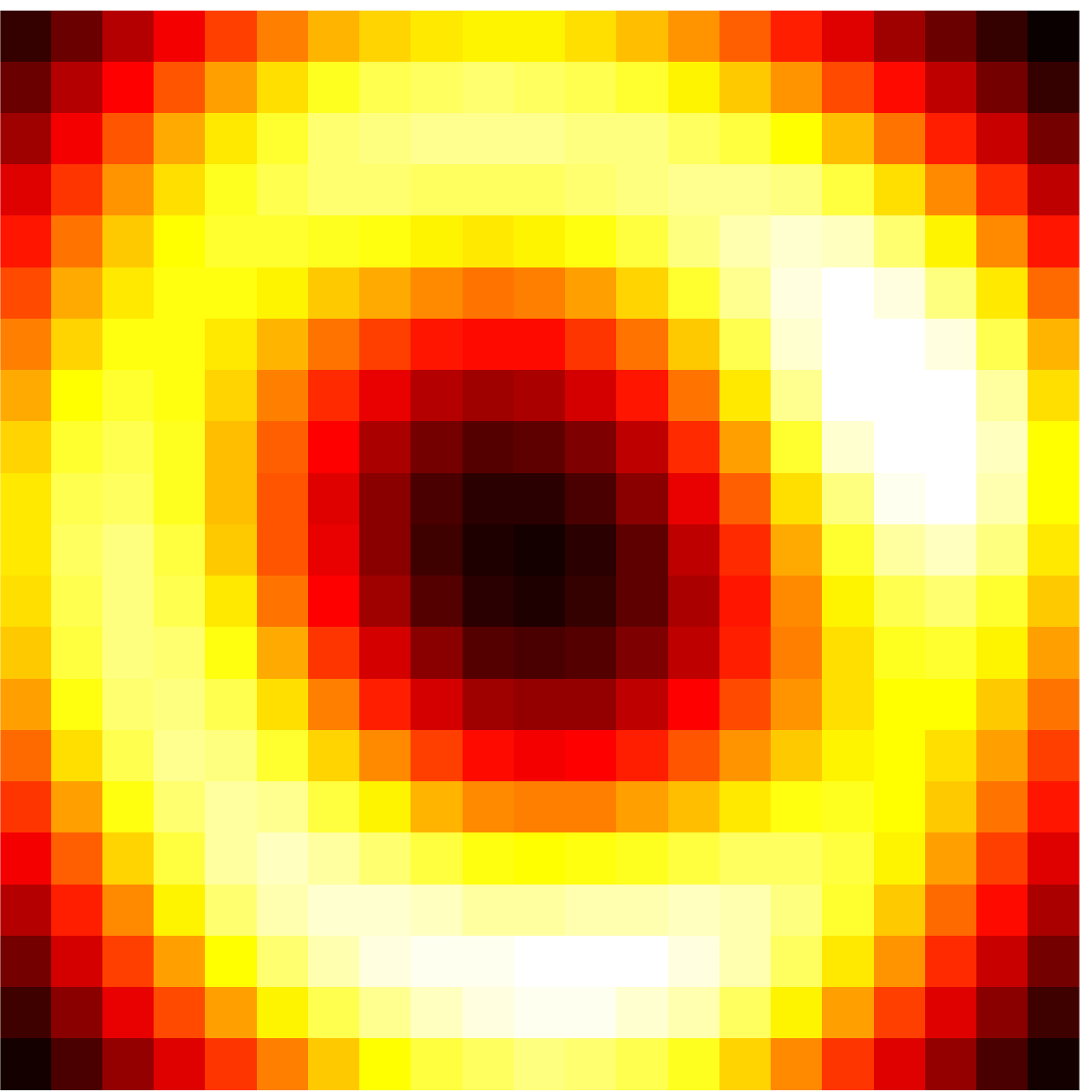} & \includegraphics[width=1.3cm]{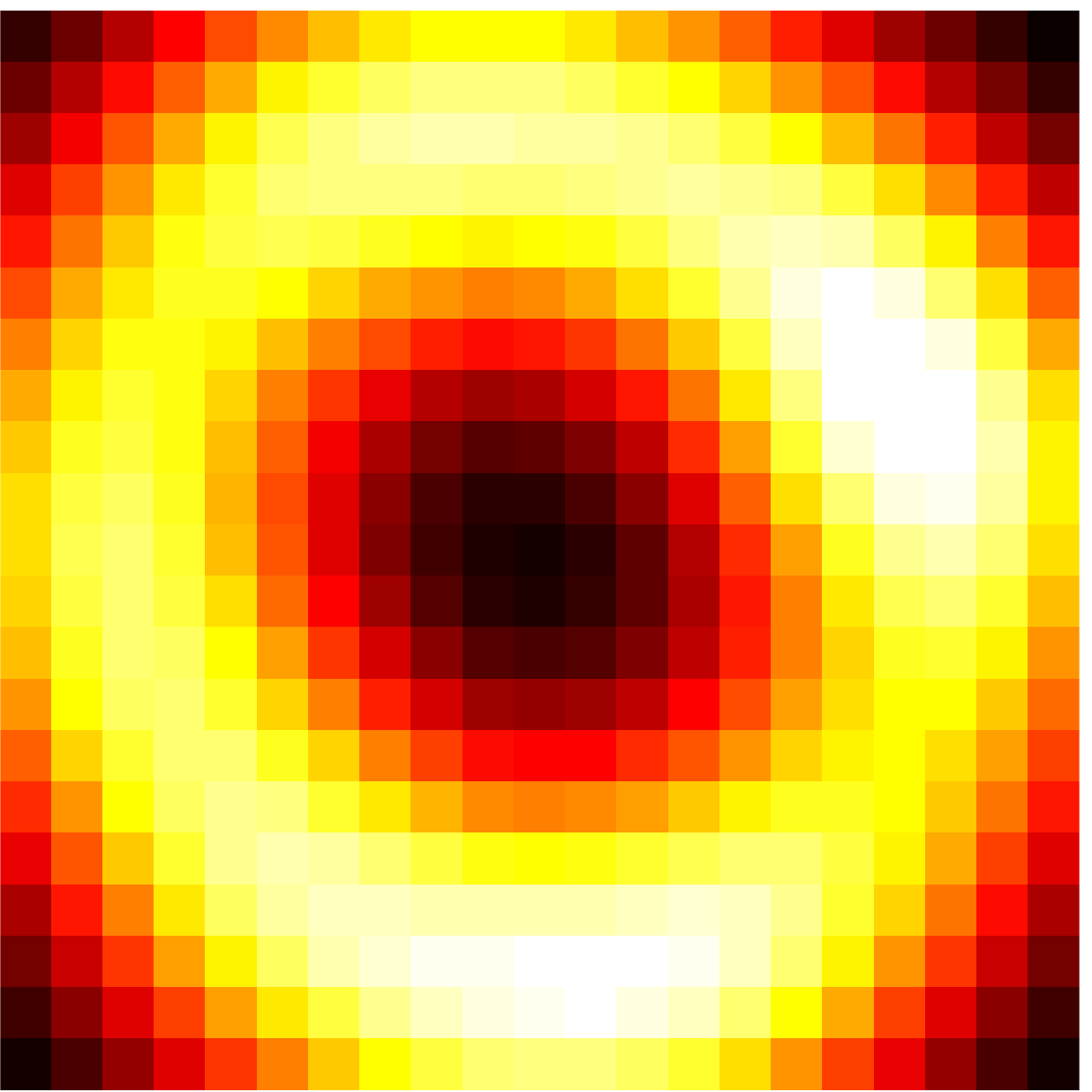} & \includegraphics[width=1.3cm]{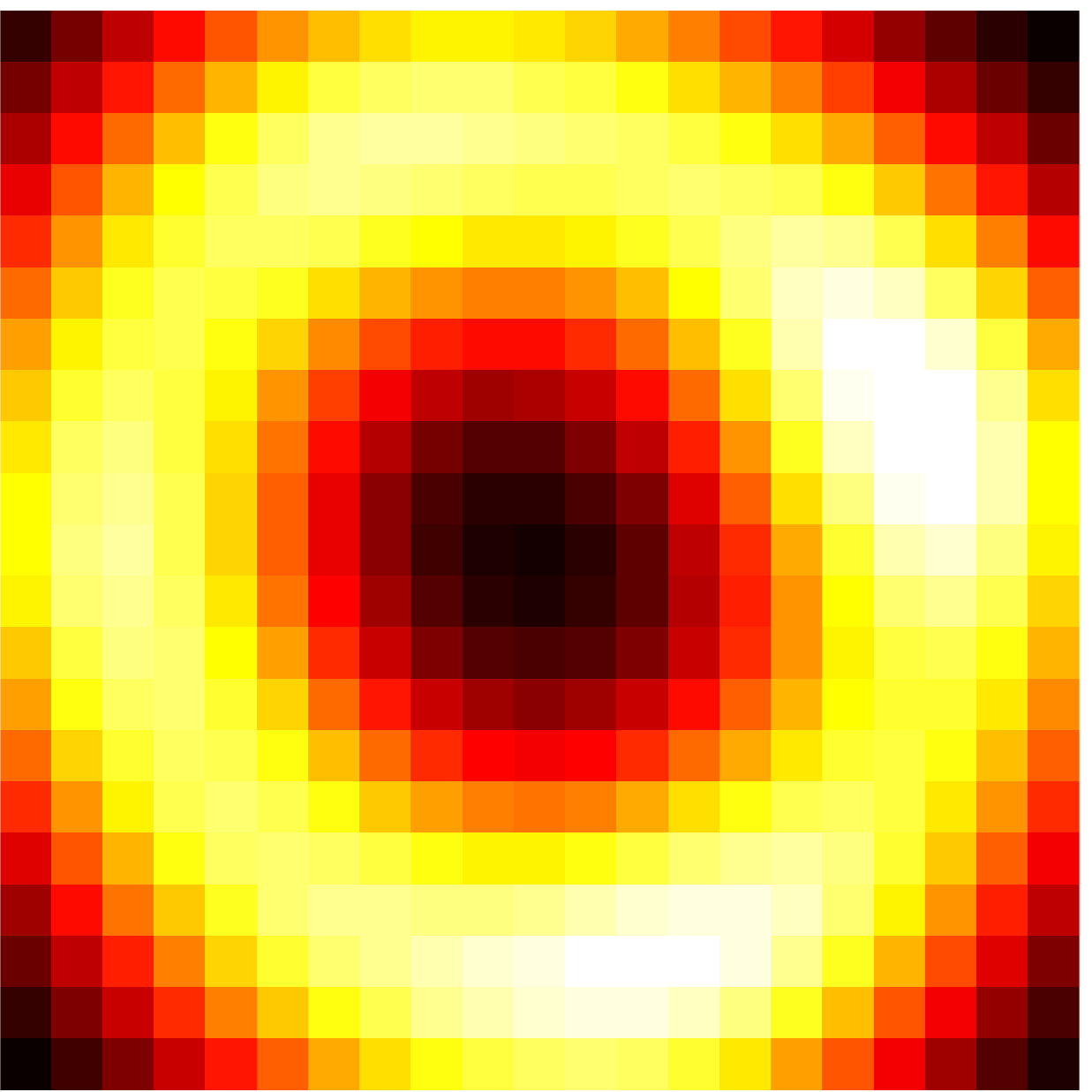} \\ 
\includegraphics[width=1.3cm]{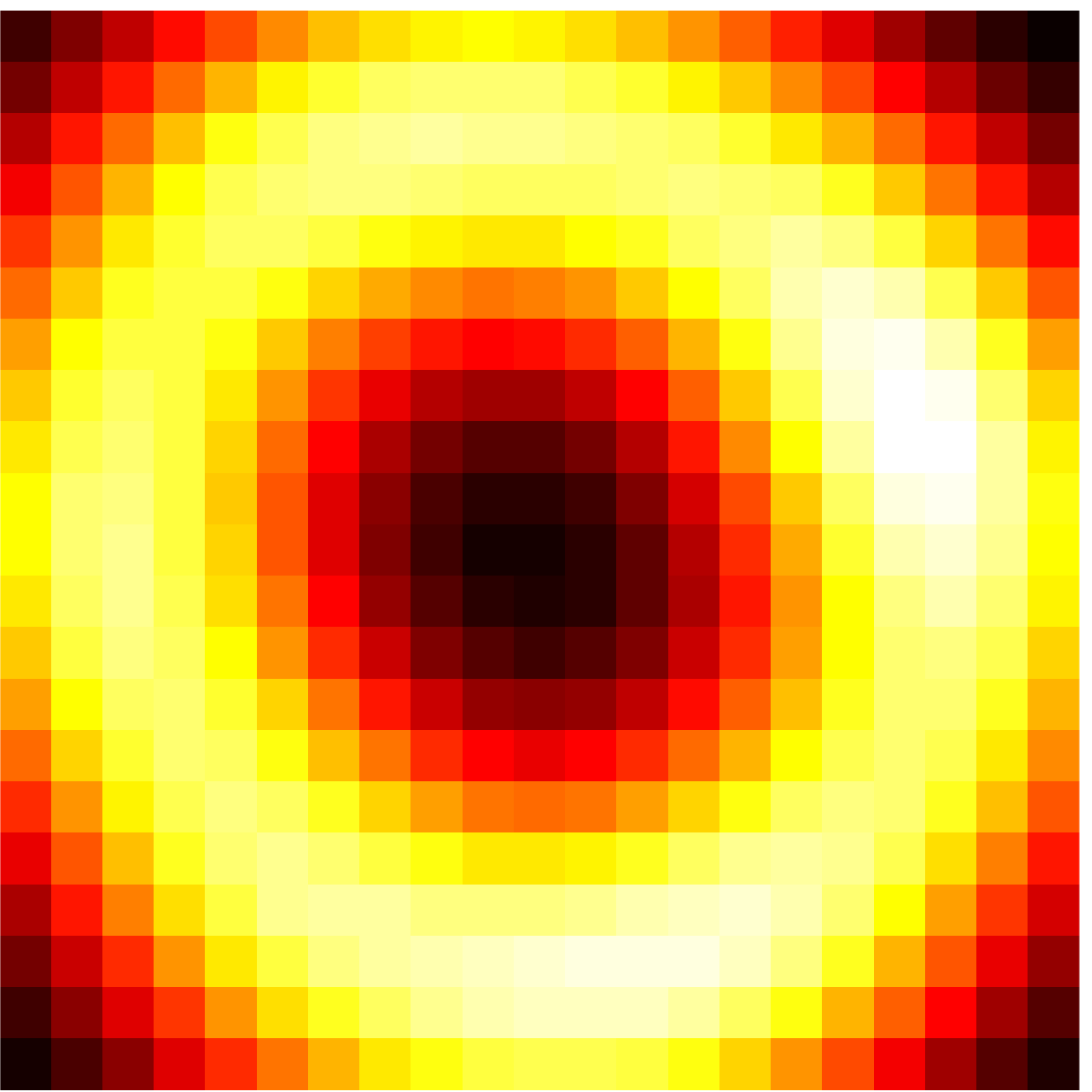} & \rota{OKECA}   & \includegraphics[width=1.3cm]{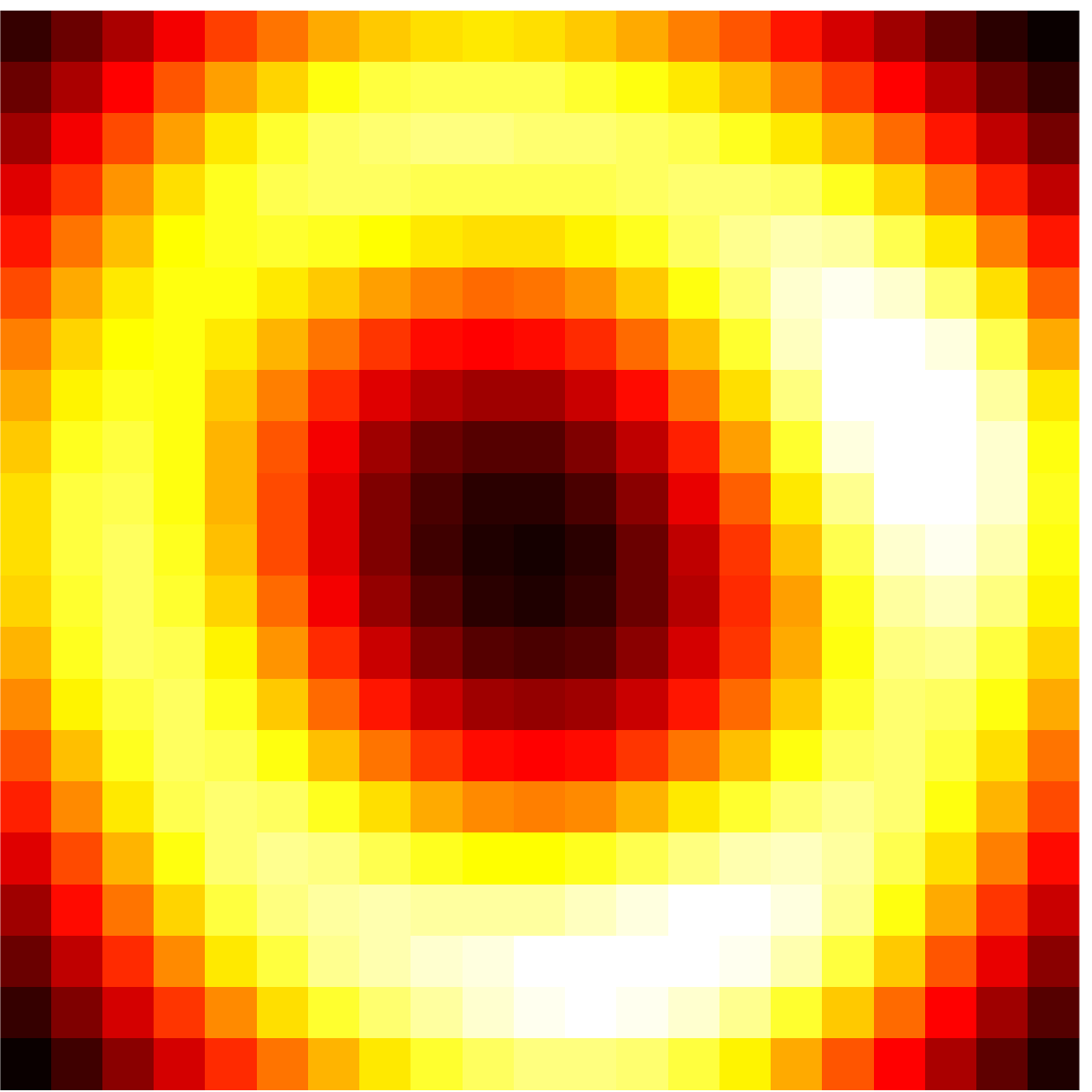} & \includegraphics[width=1.3cm]{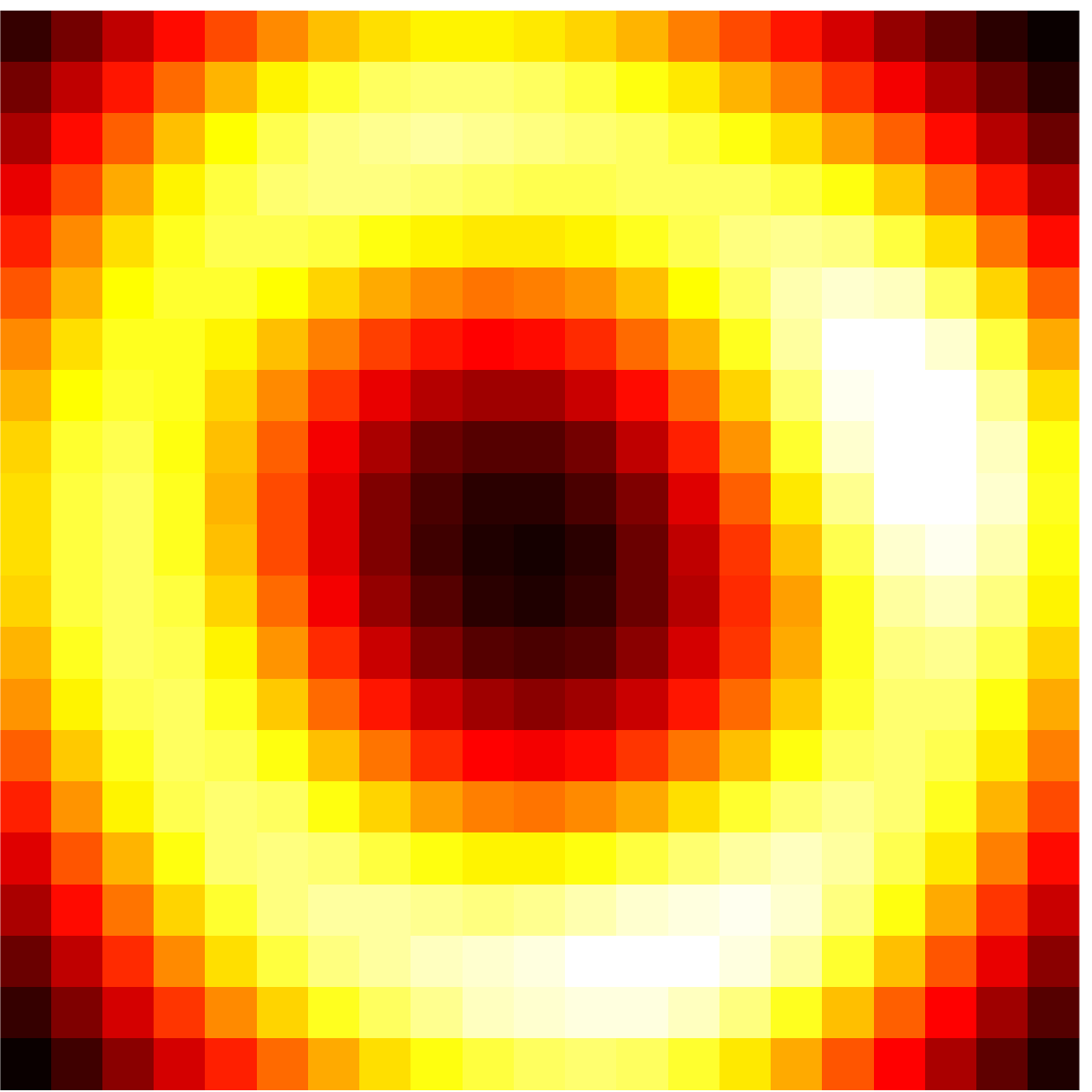} & \includegraphics[width=1.3cm]{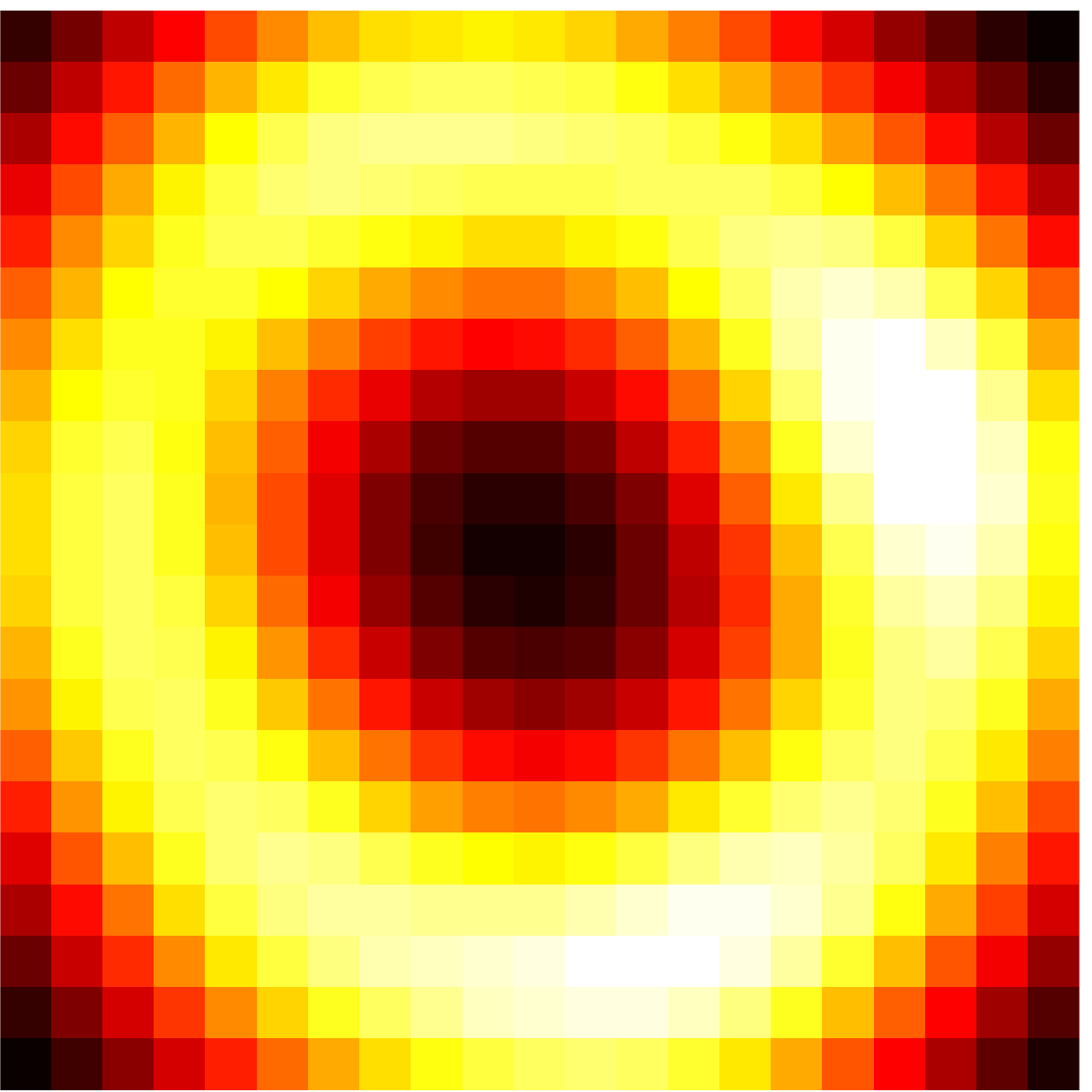} & \includegraphics[width=1.3cm]{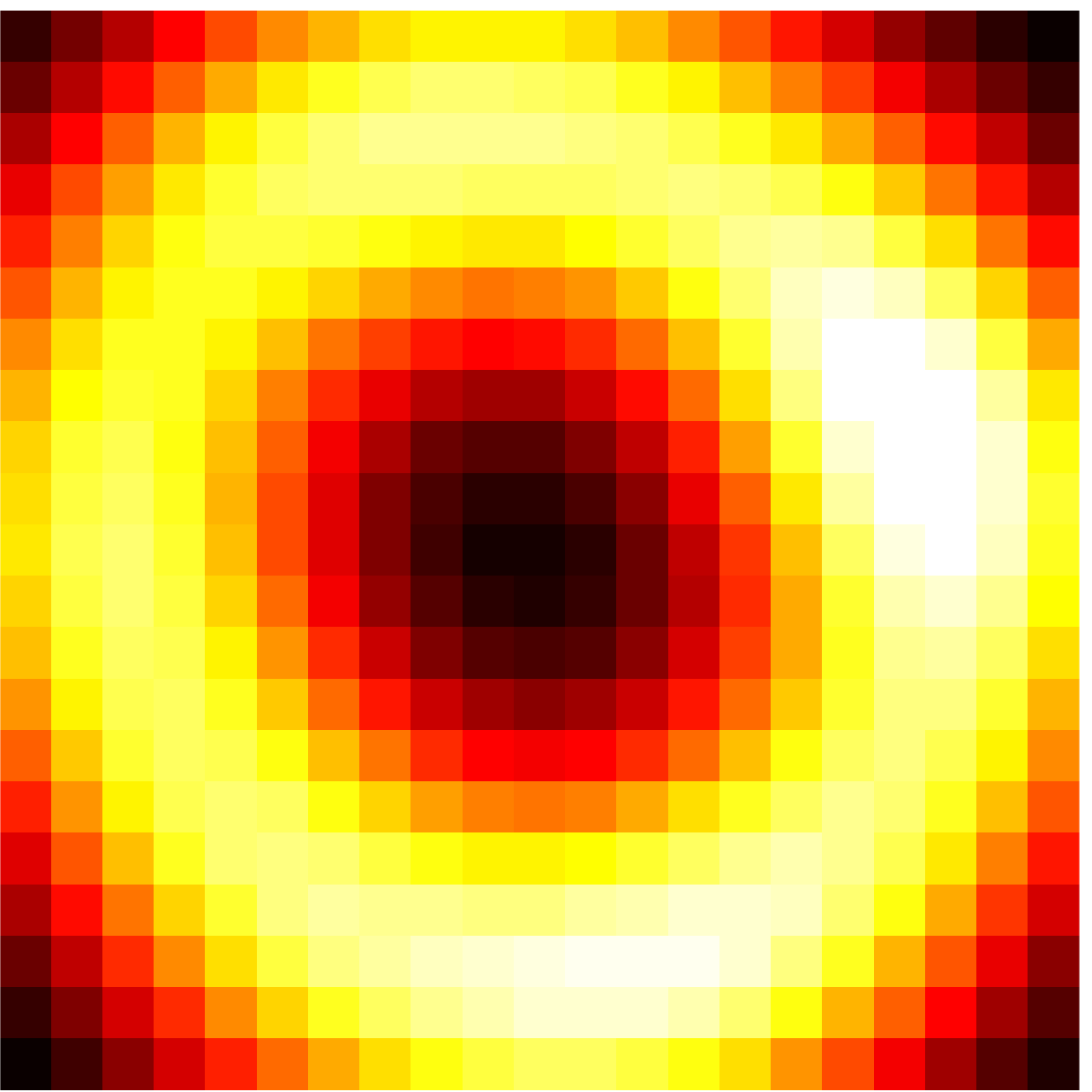} & \includegraphics[width=1.3cm]{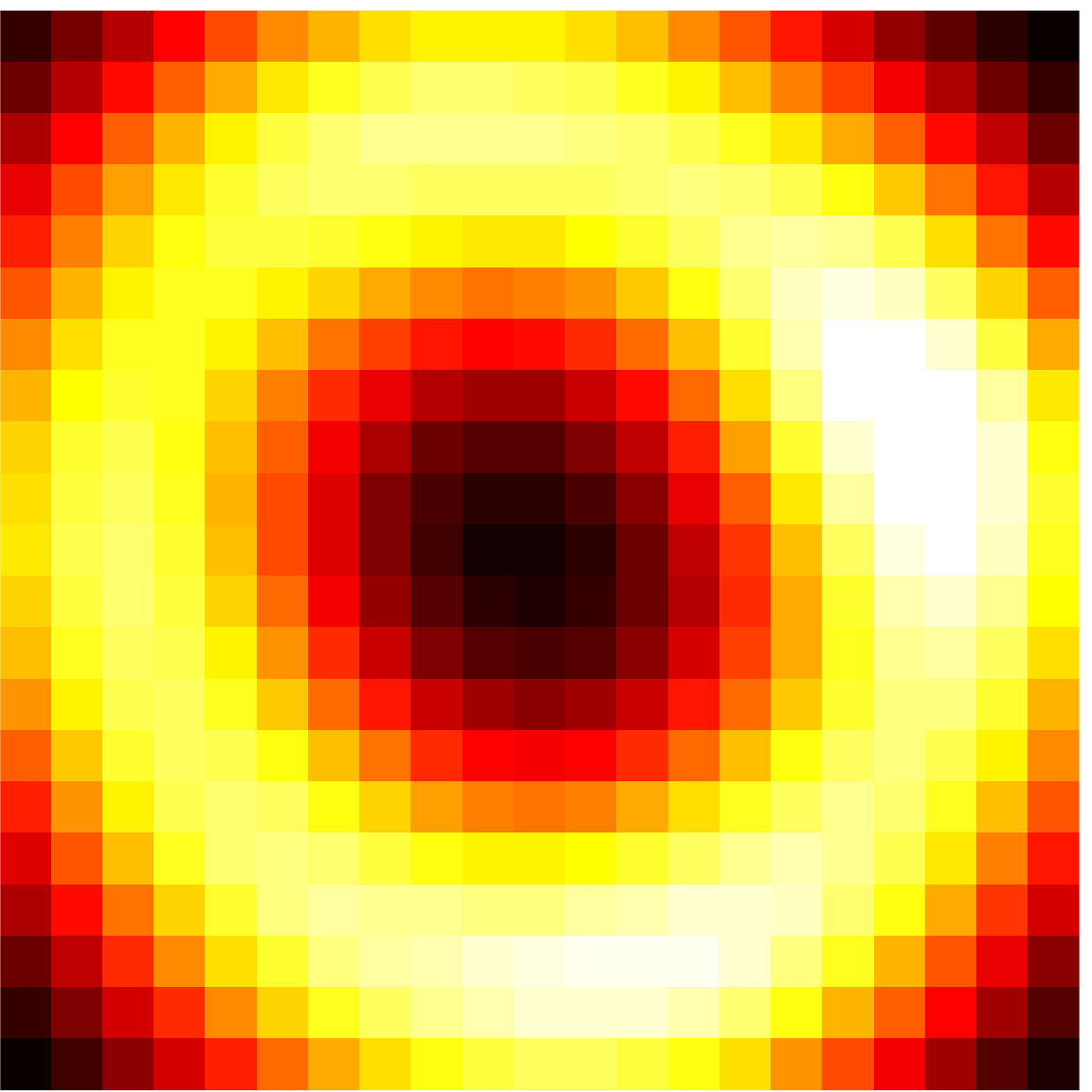}\\ 
\hline
$\sigma_{d1}$ & \rota{KECA}  &\includegraphics[width=1.3cm]{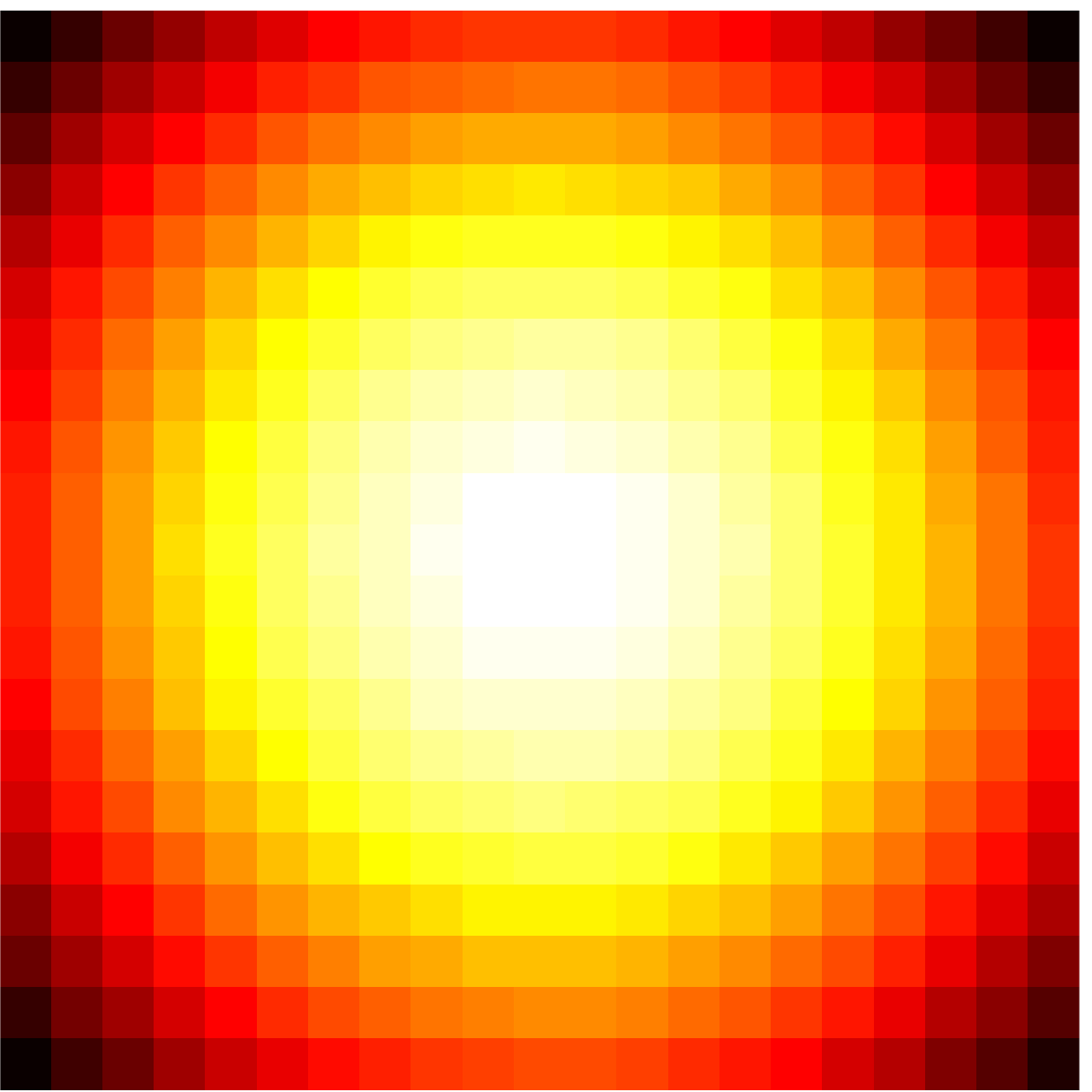} & \includegraphics[width=1.3cm]{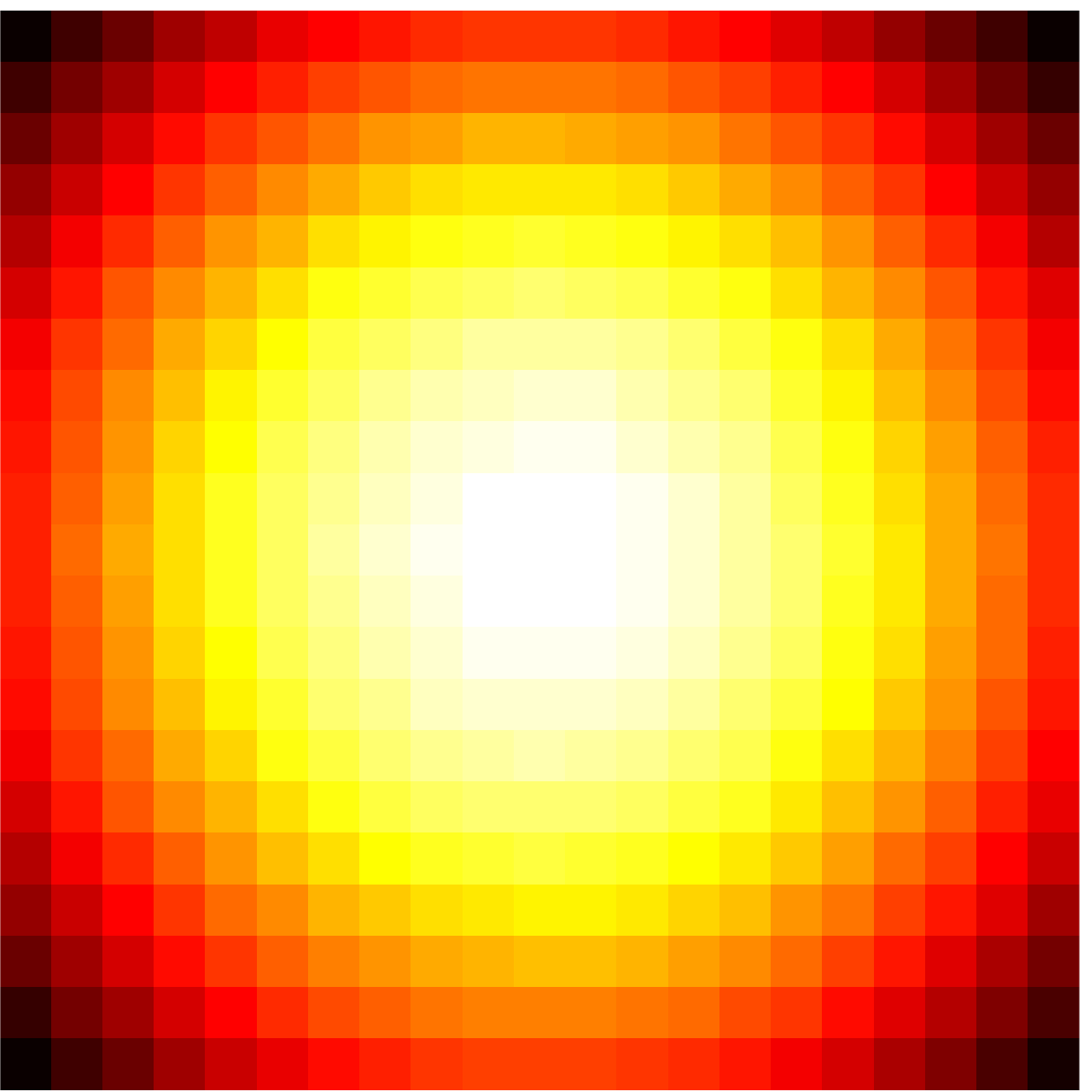} & \includegraphics[width=1.3cm]{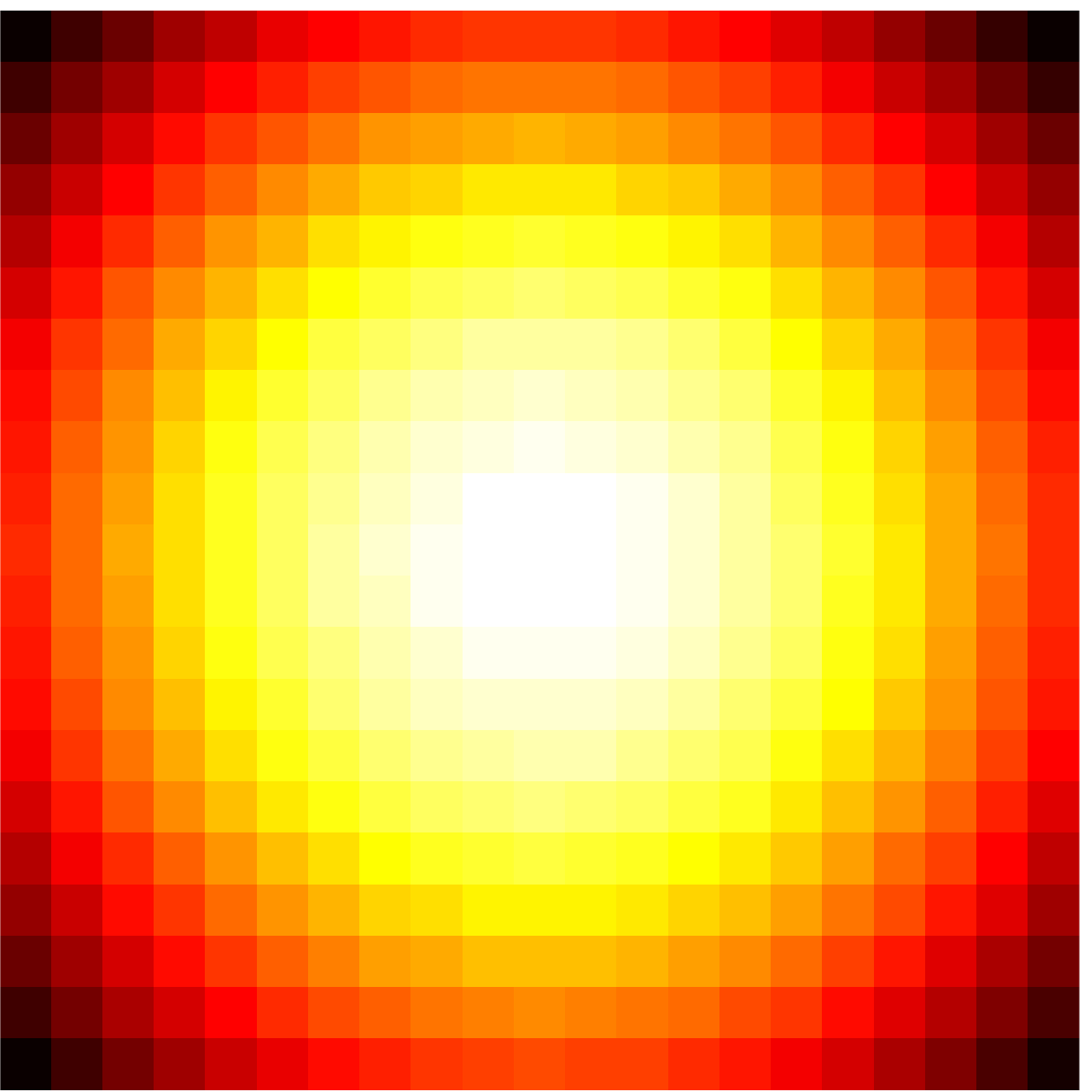} & \includegraphics[width=1.3cm]{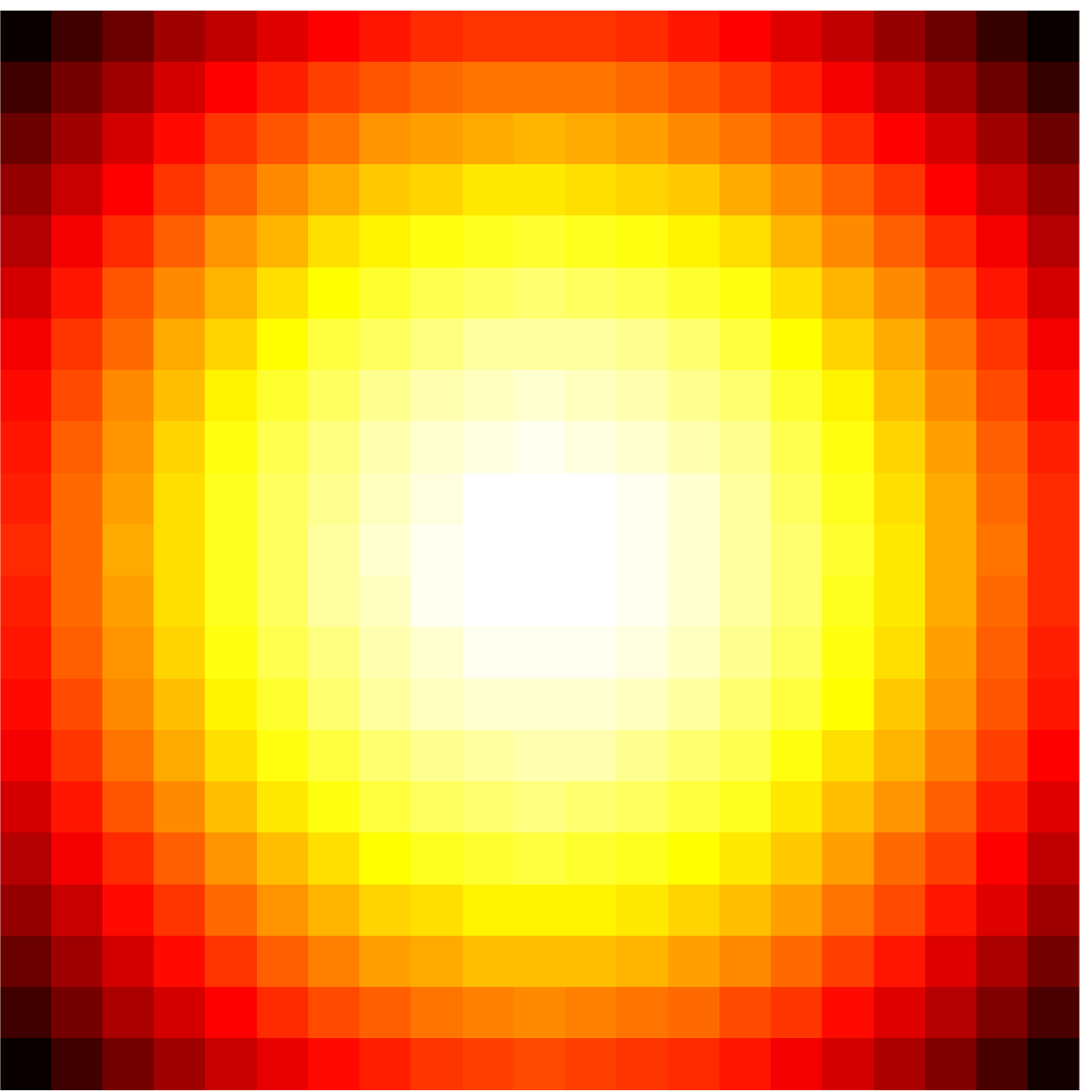} & \includegraphics[width=1.3cm]{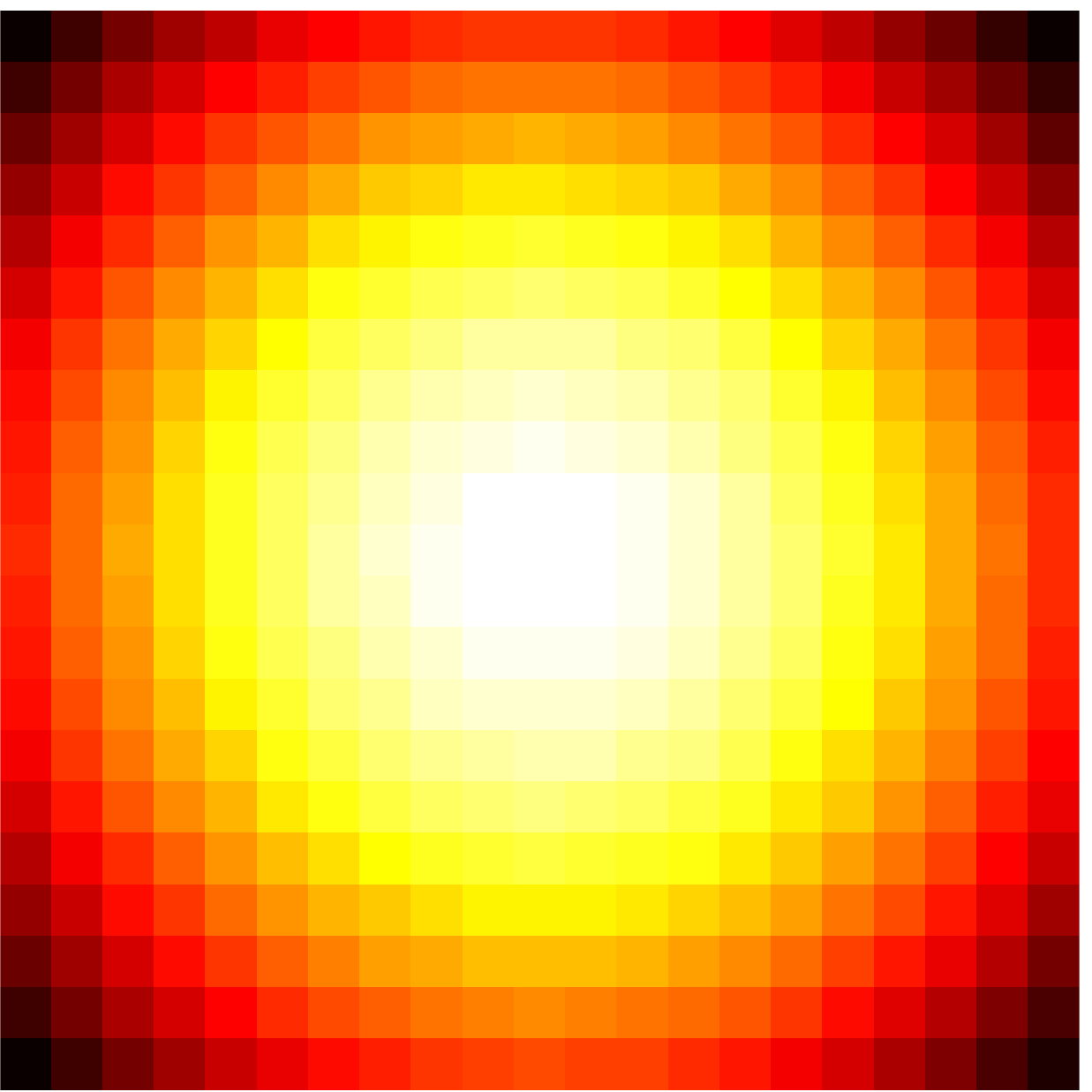} \\ 
\includegraphics[width=1.3cm]{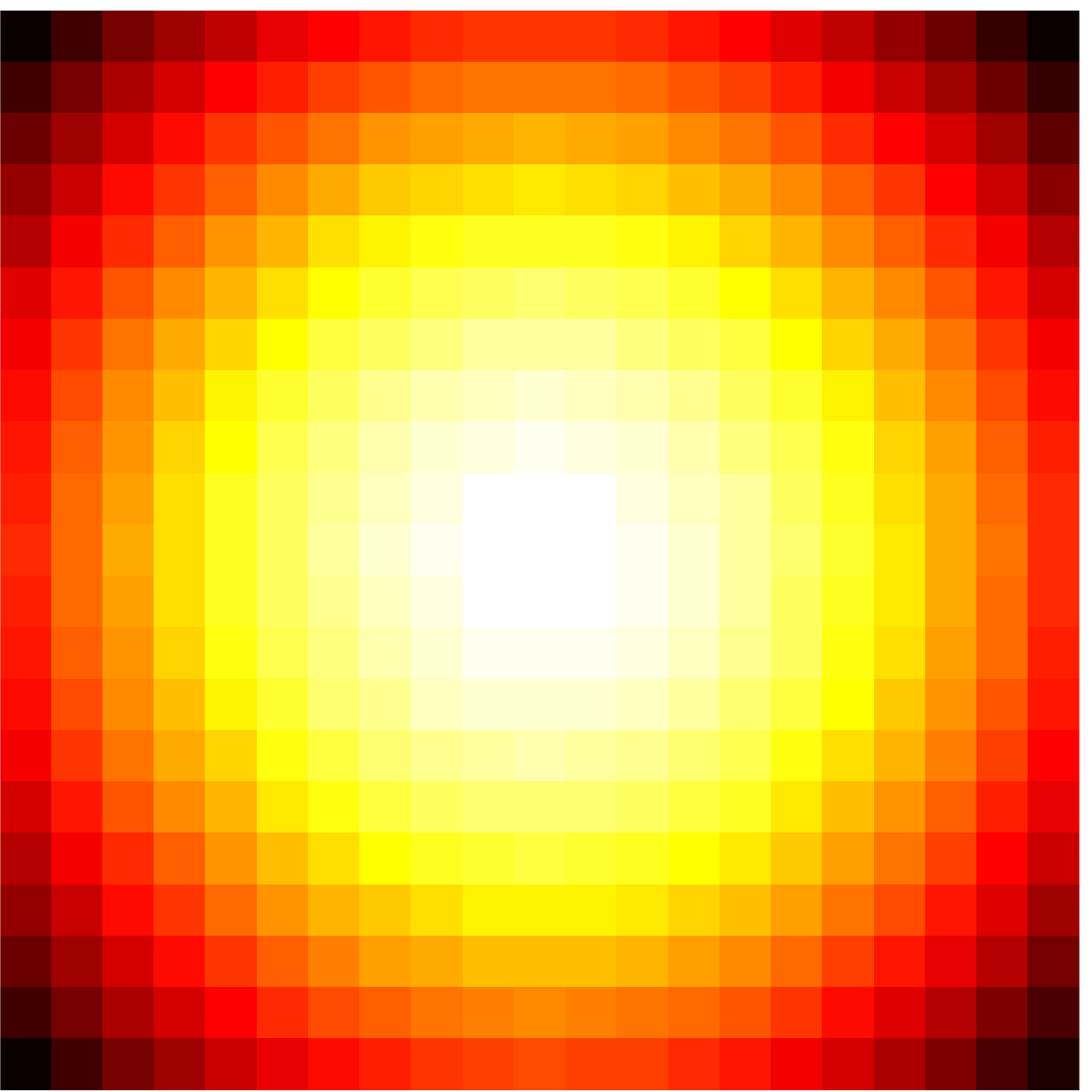} & \rota{OKECA}   & \includegraphics[width=1.3cm]{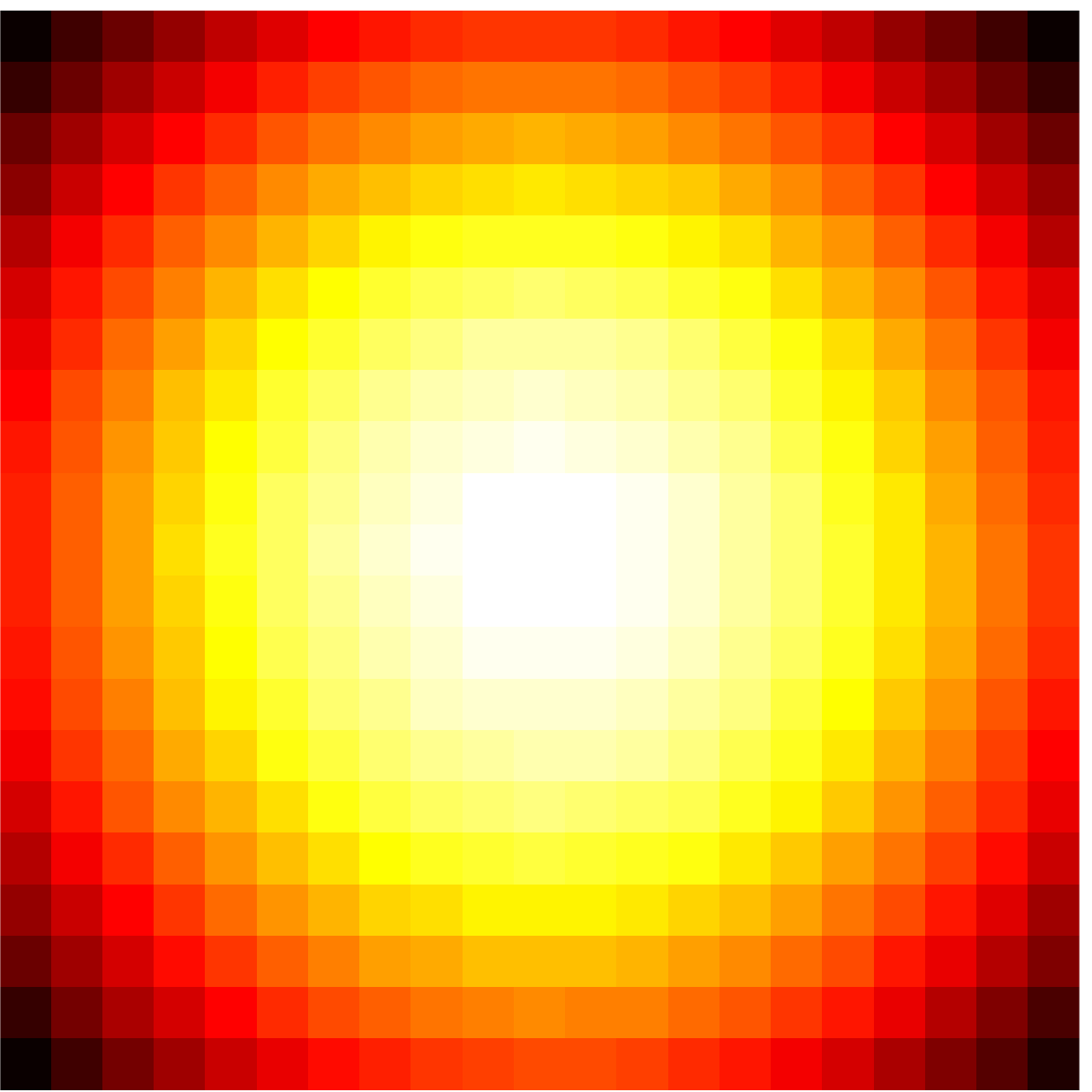} & \includegraphics[width=1.3cm]{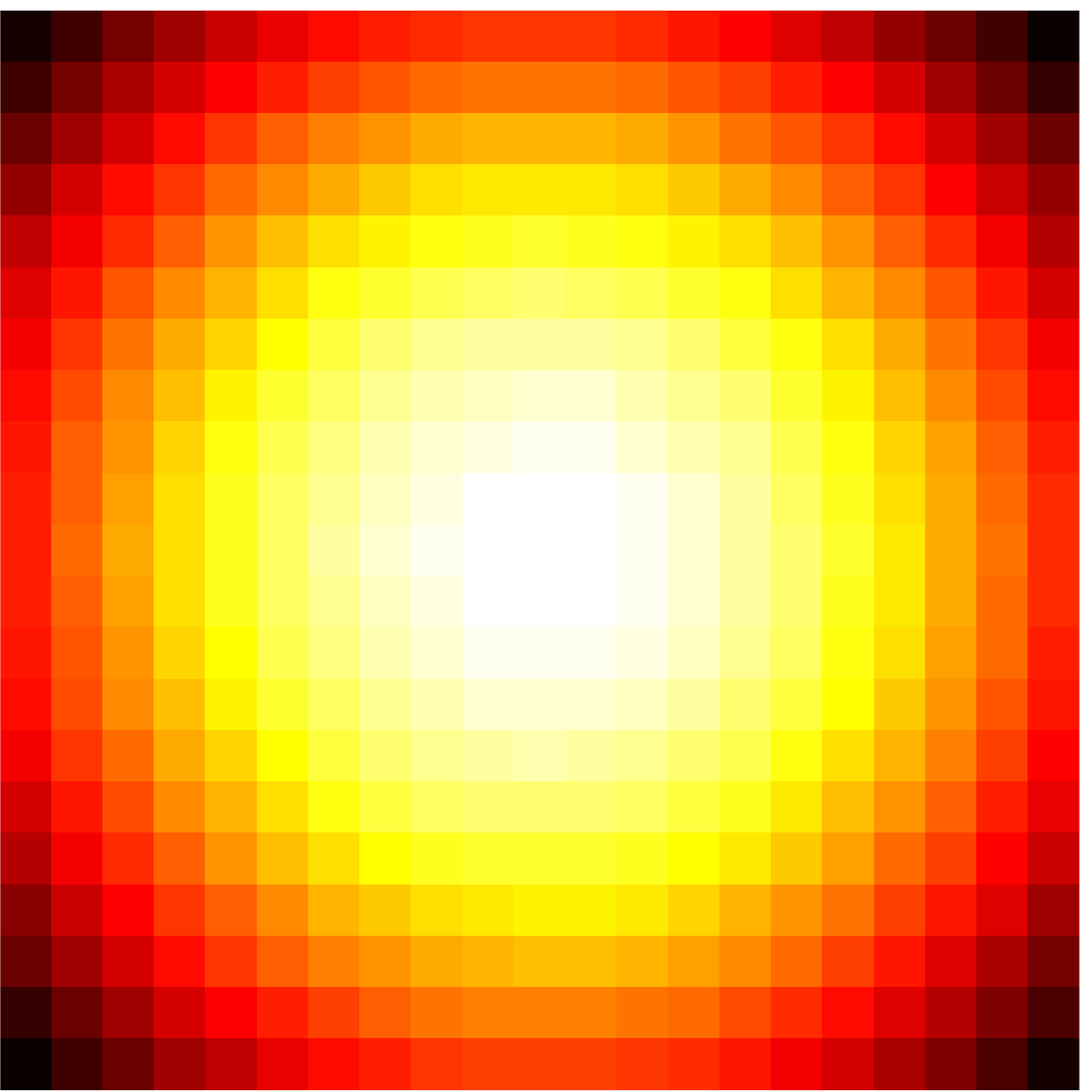} & \includegraphics[width=1.3cm]{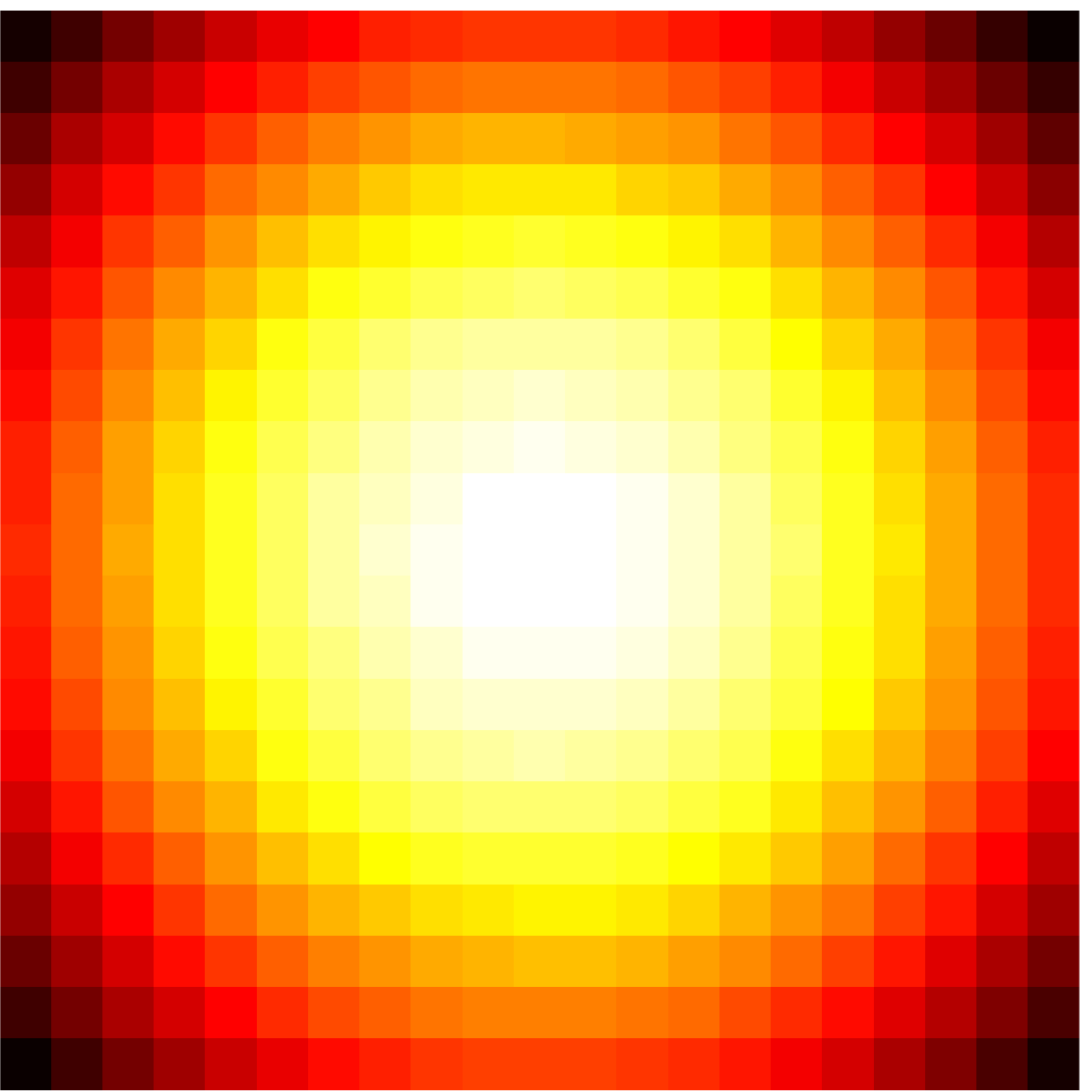} & \includegraphics[width=1.3cm]{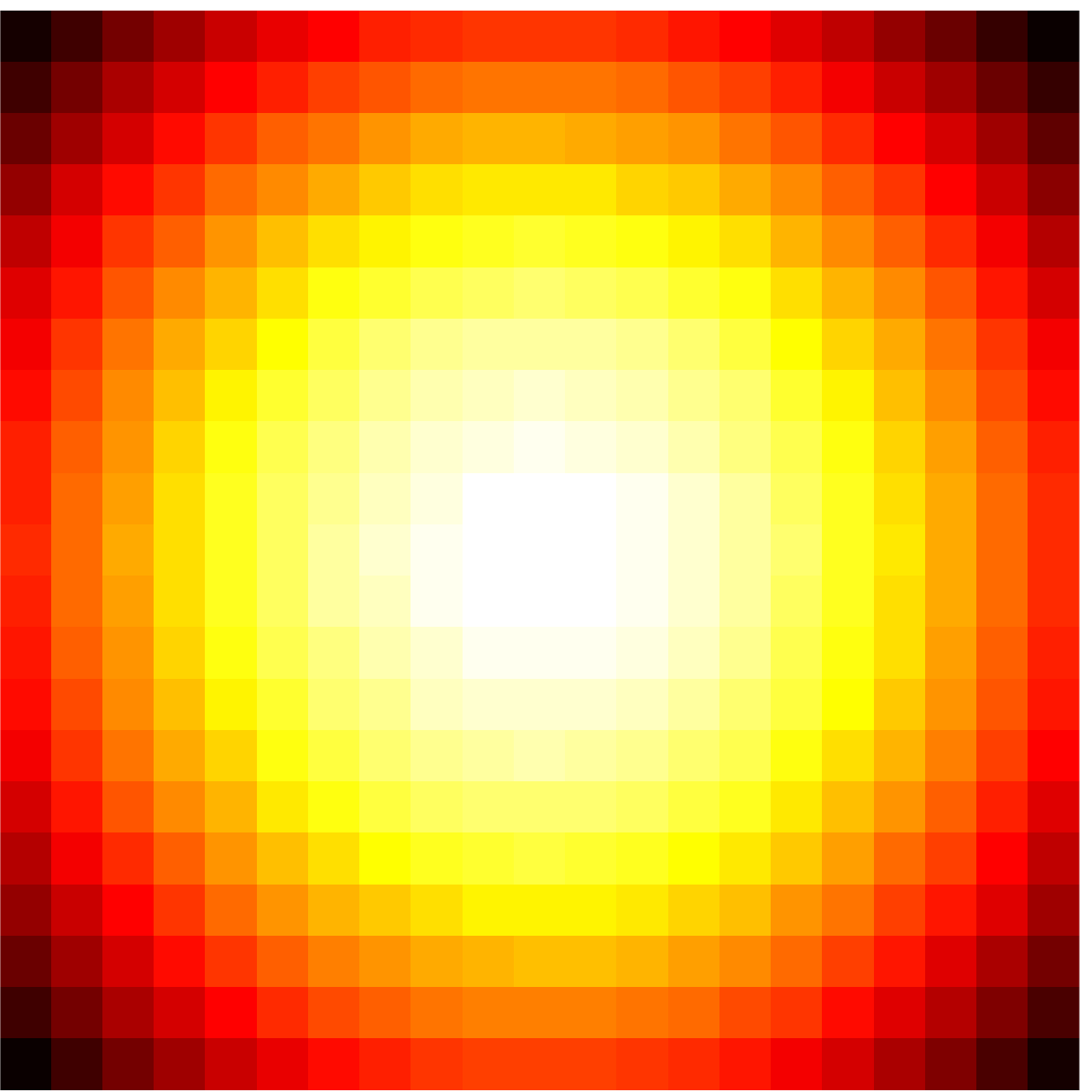} & \includegraphics[width=1.3cm]{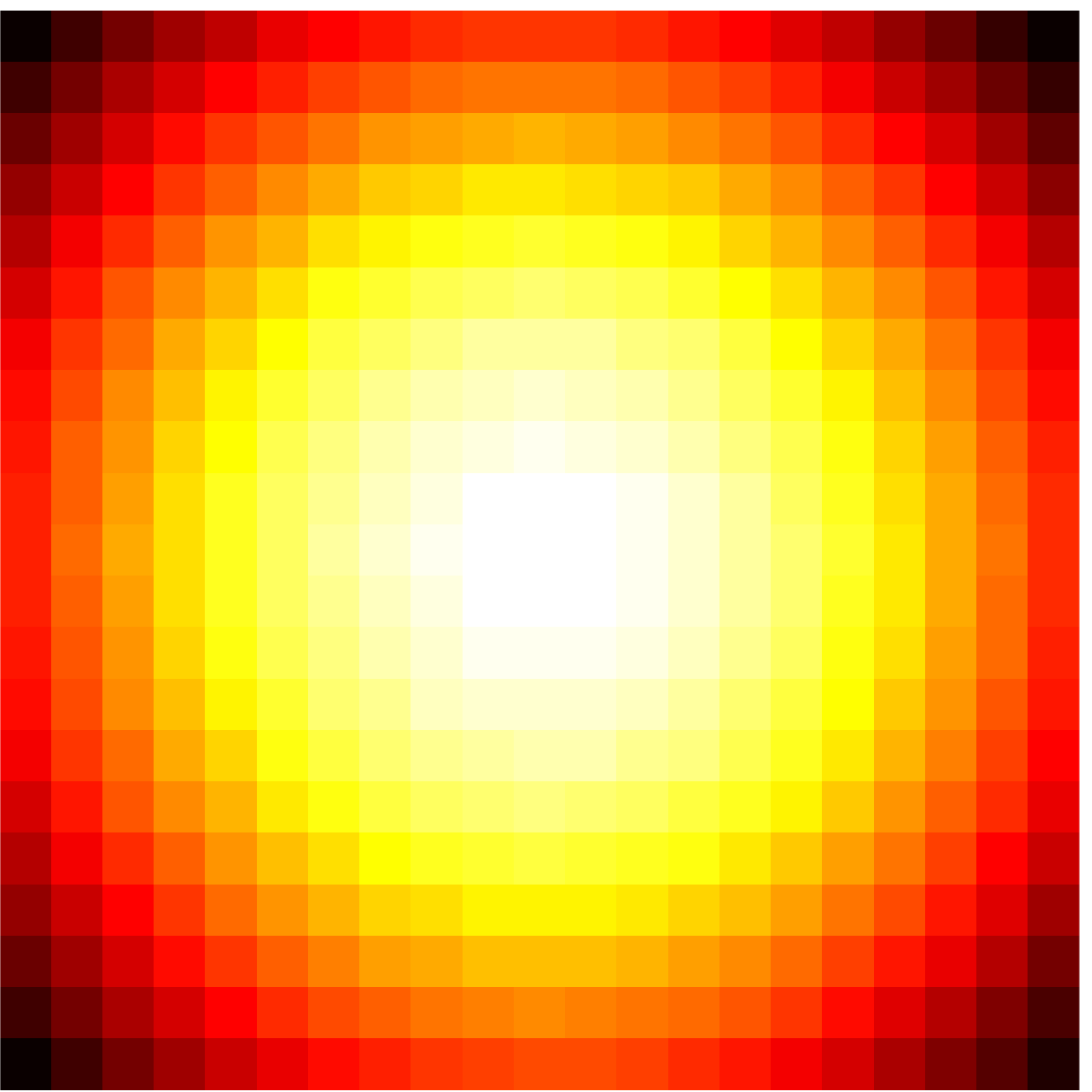}\\ 
\hline
$\sigma_{d2}$ & \rota{KECA}  &\includegraphics[width=1.3cm]{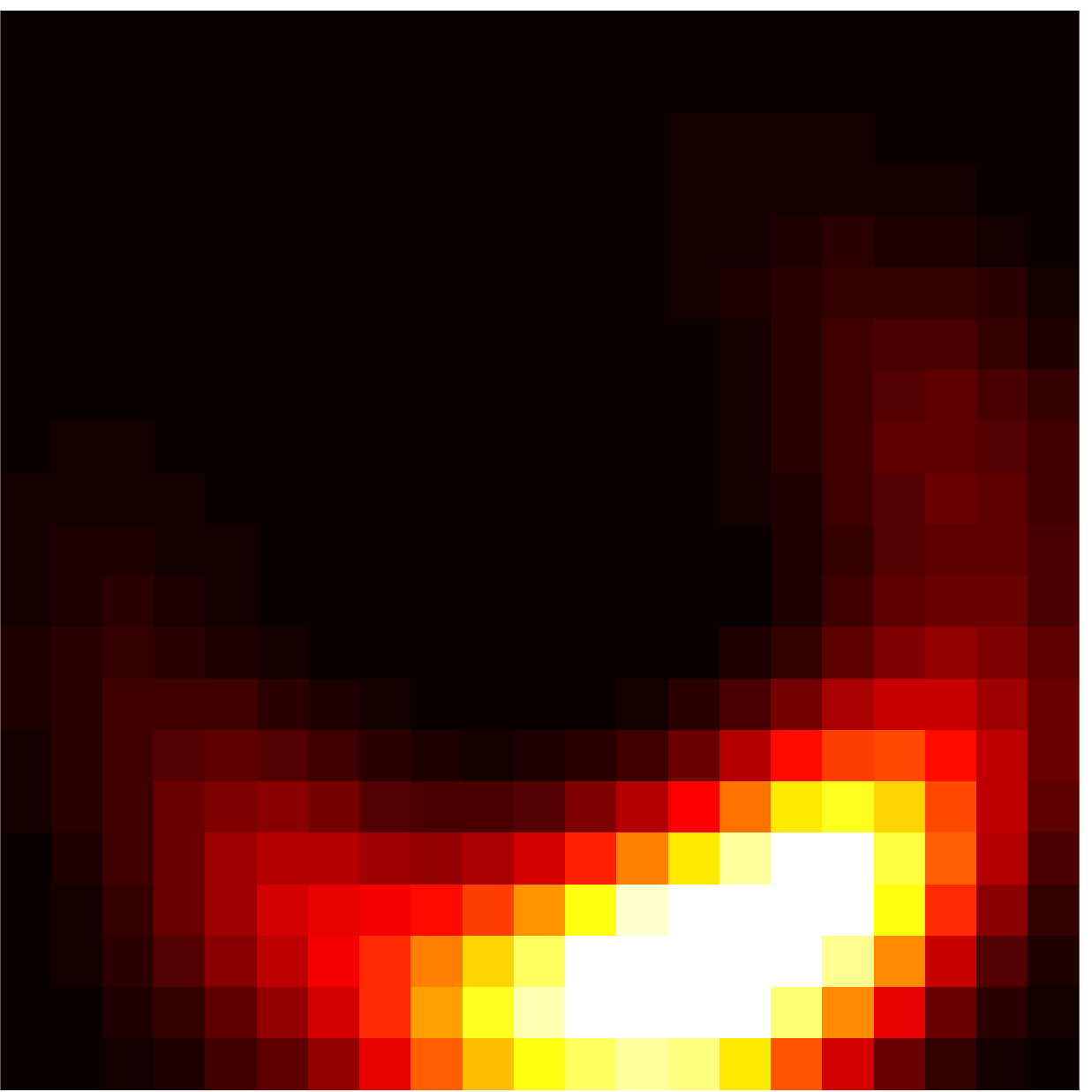} & \includegraphics[width=1.3cm]{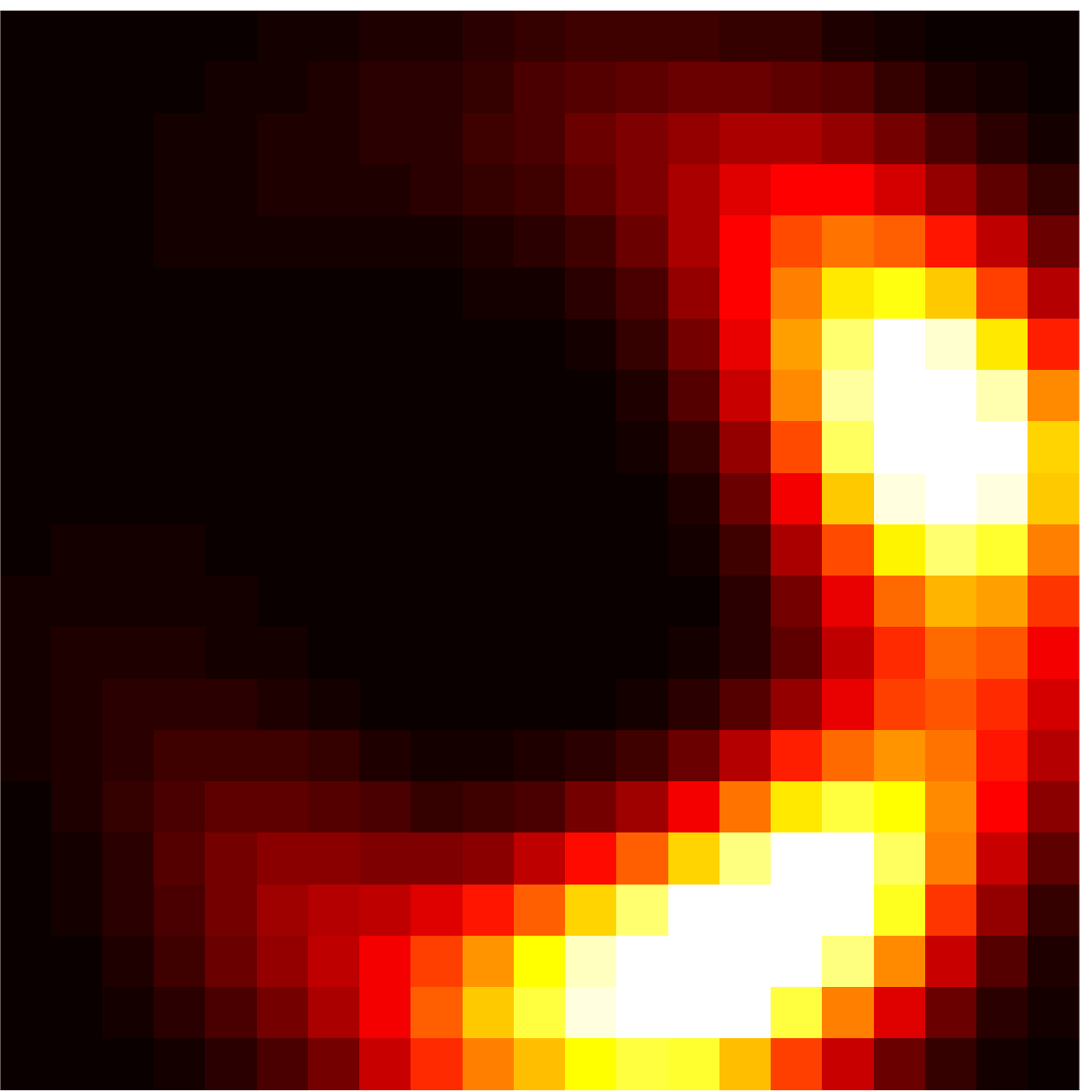} & \includegraphics[width=1.3cm]{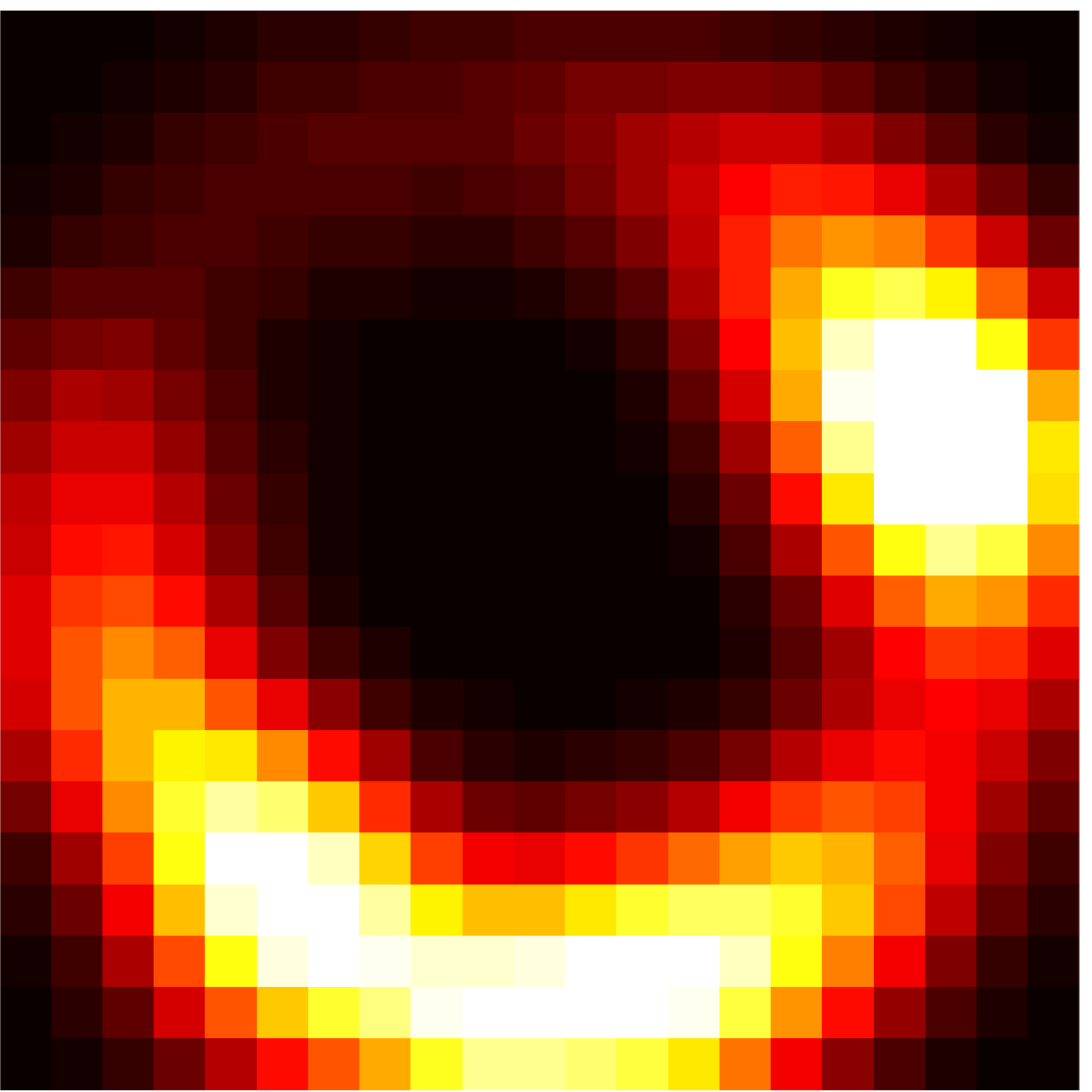} & \includegraphics[width=1.3cm]{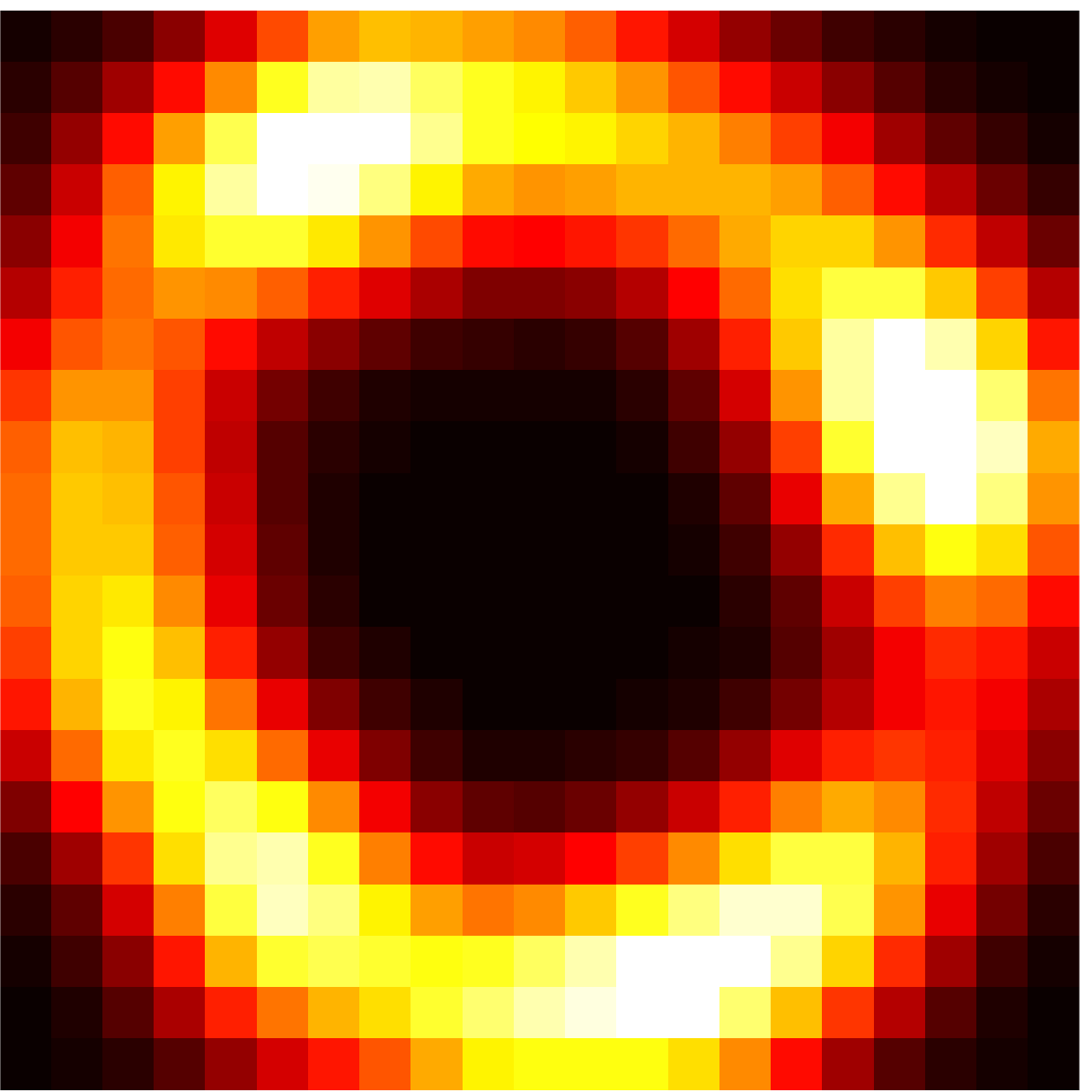} & \includegraphics[width=1.3cm]{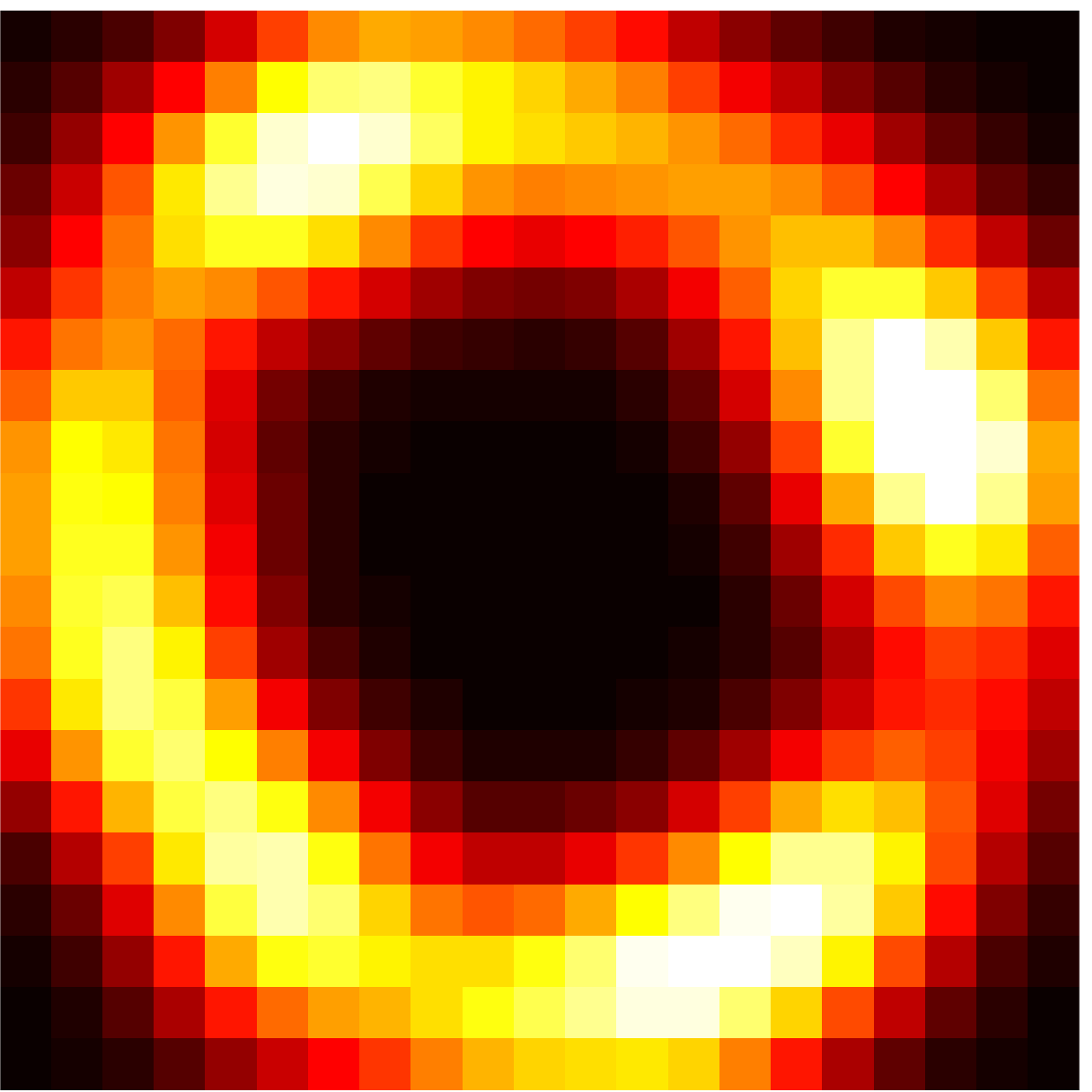} \\ 
\includegraphics[width=1.3cm]{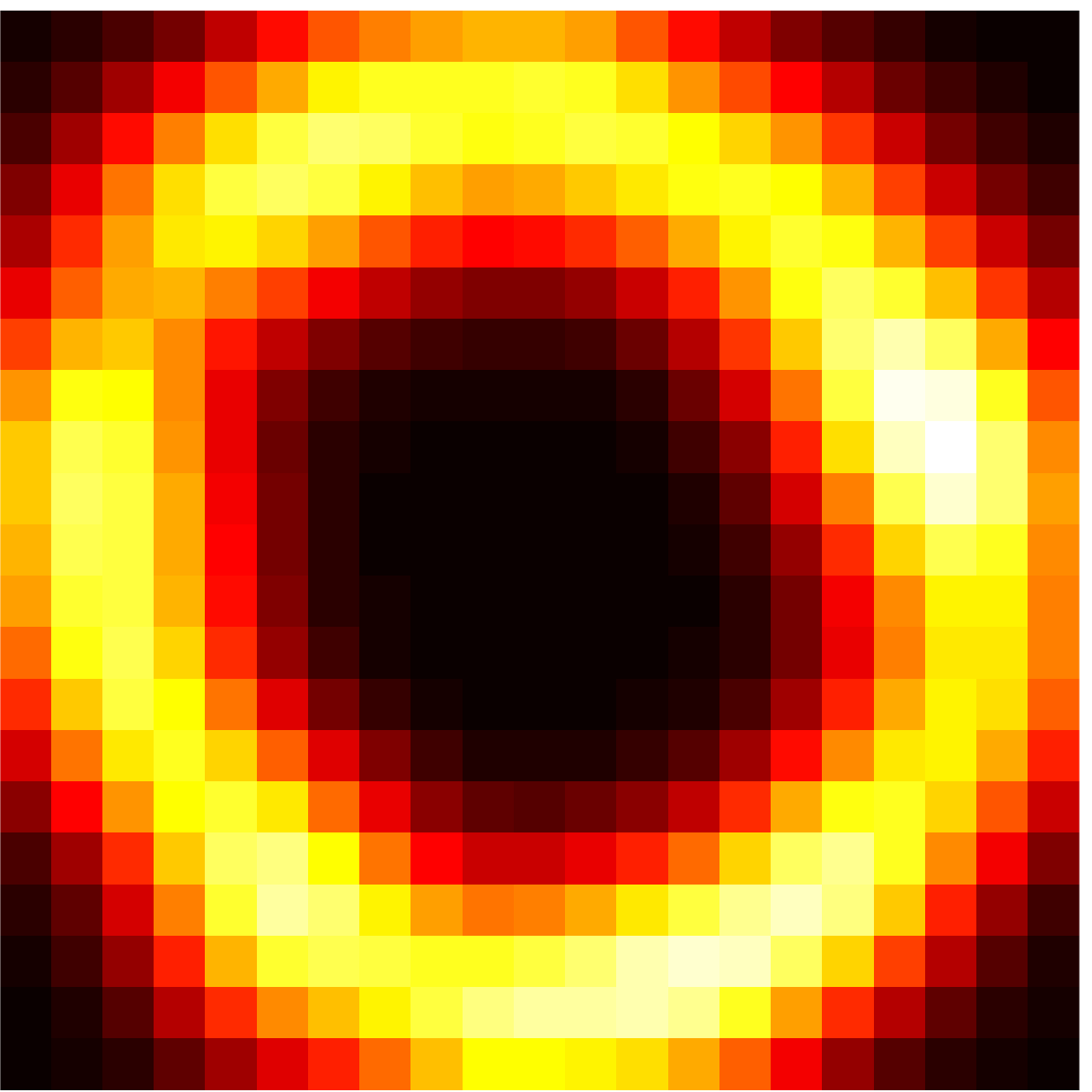} & \rota{OKECA}   & \includegraphics[width=1.3cm]{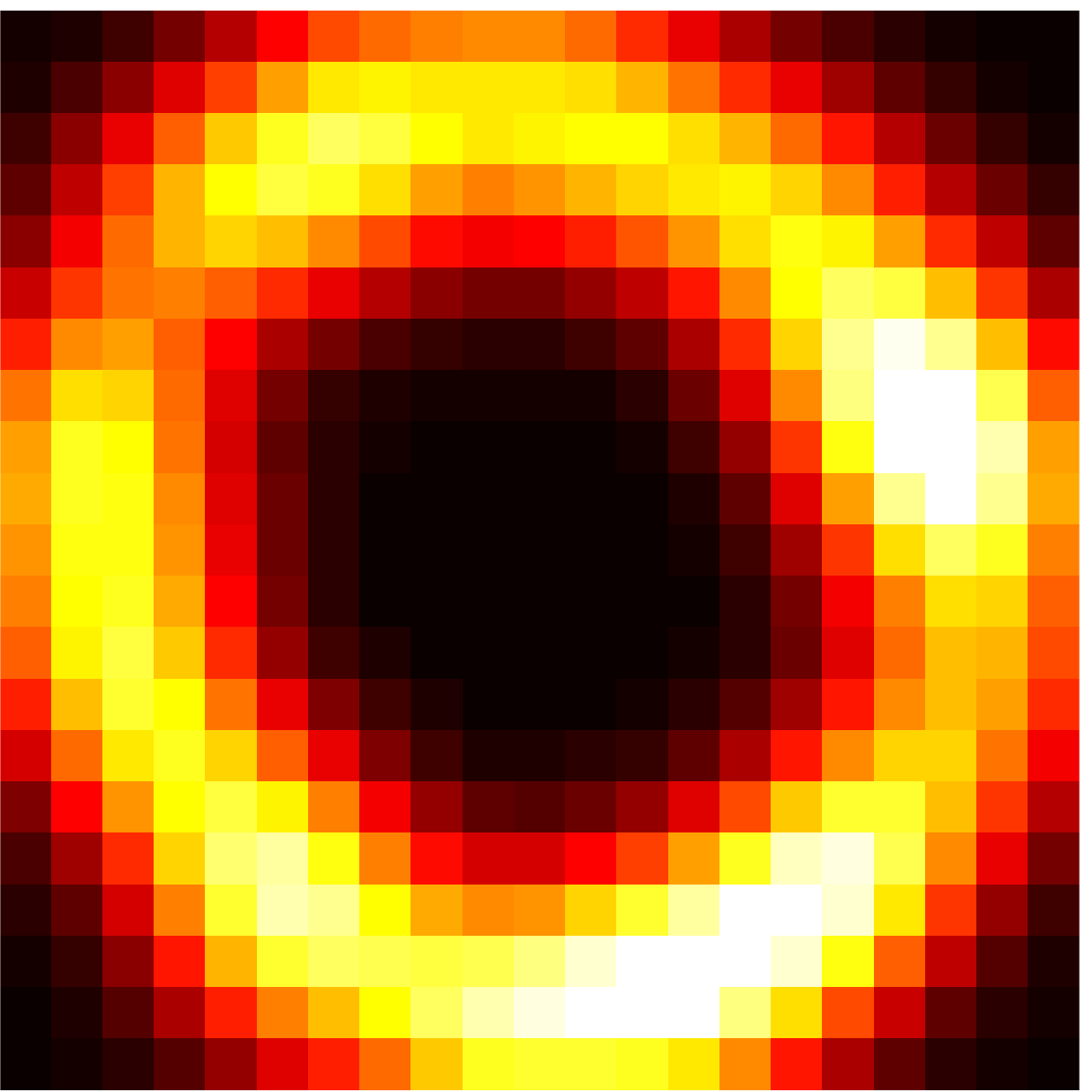} & \includegraphics[width=1.3cm]{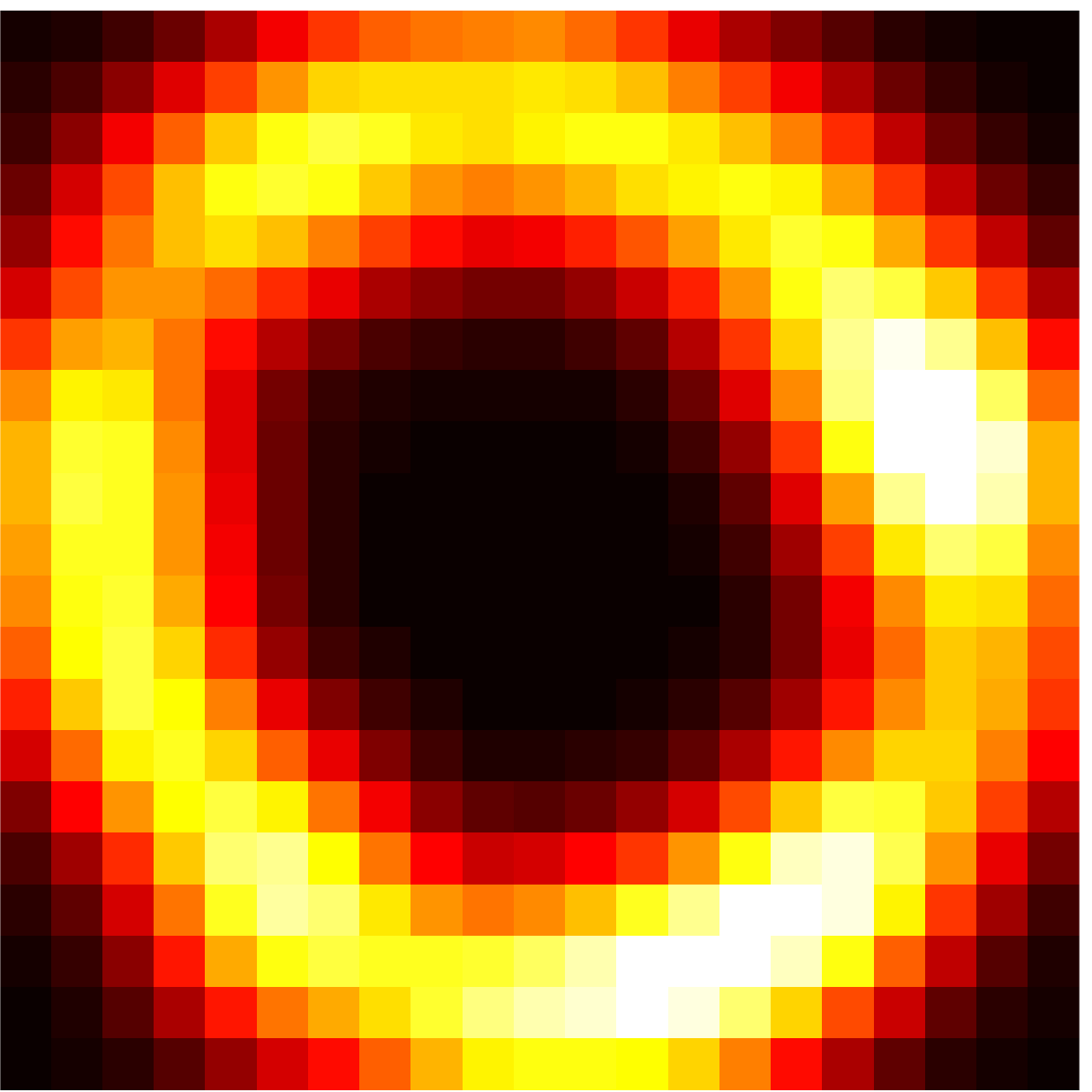} & \includegraphics[width=1.3cm]{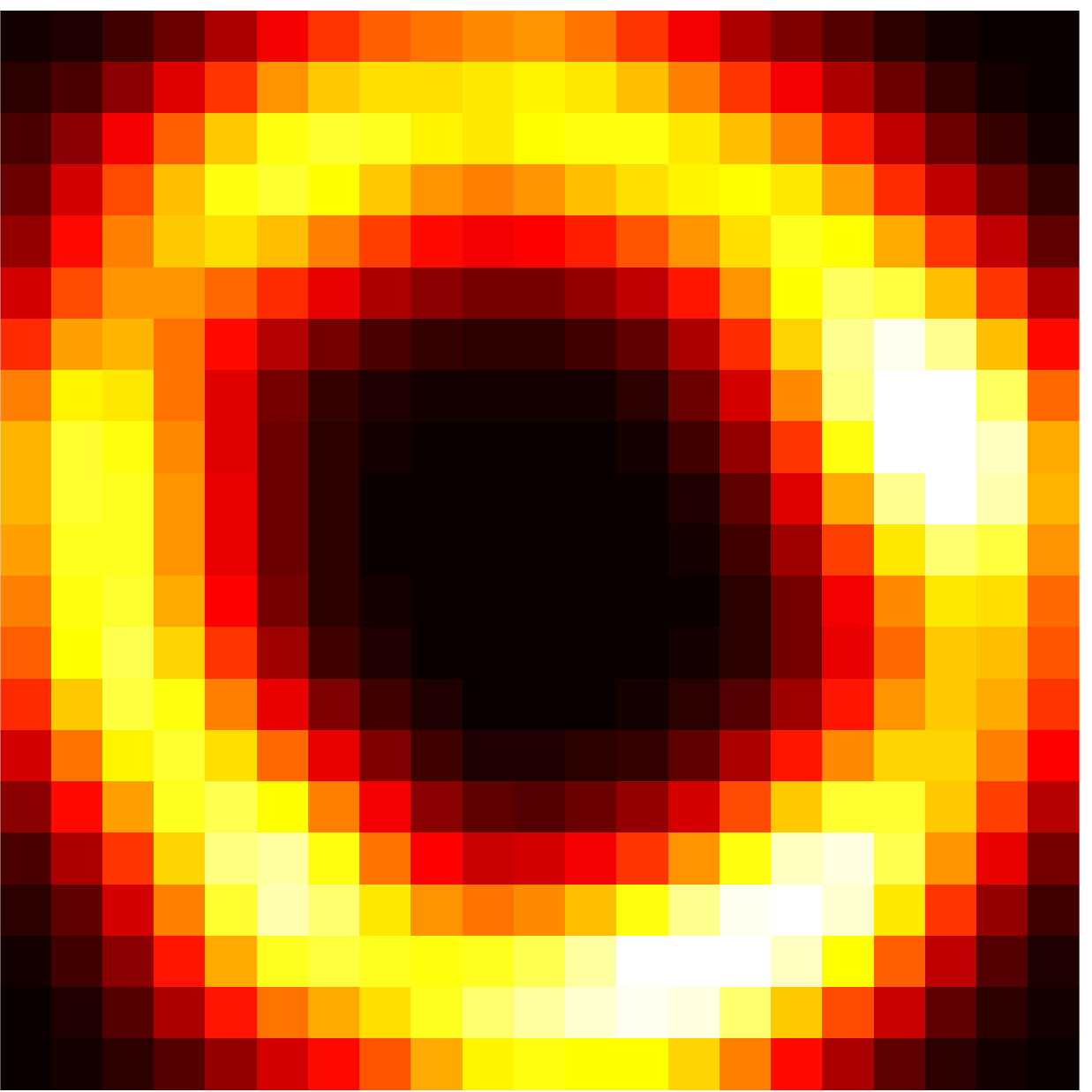} & \includegraphics[width=1.3cm]{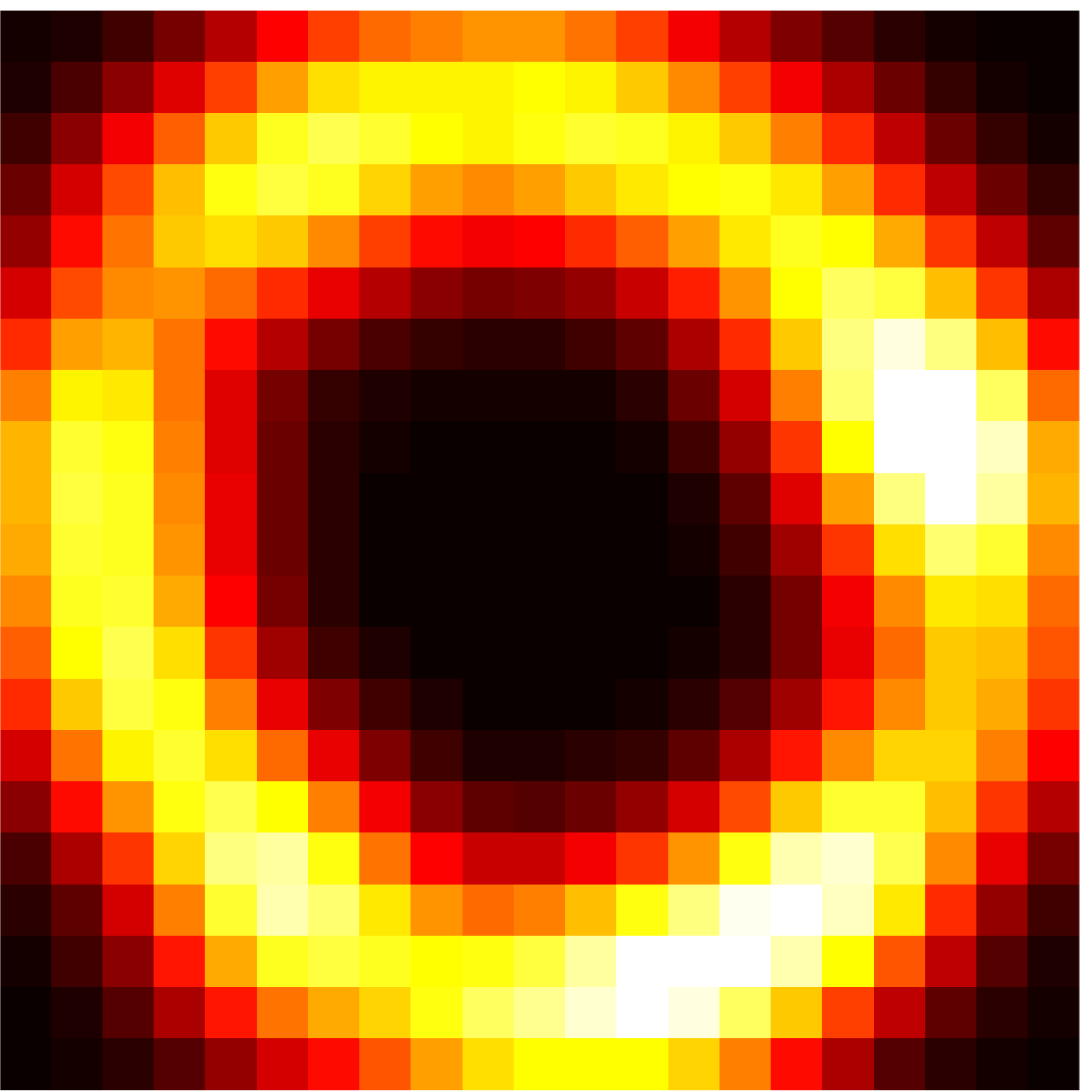} & \includegraphics[width=1.3cm]{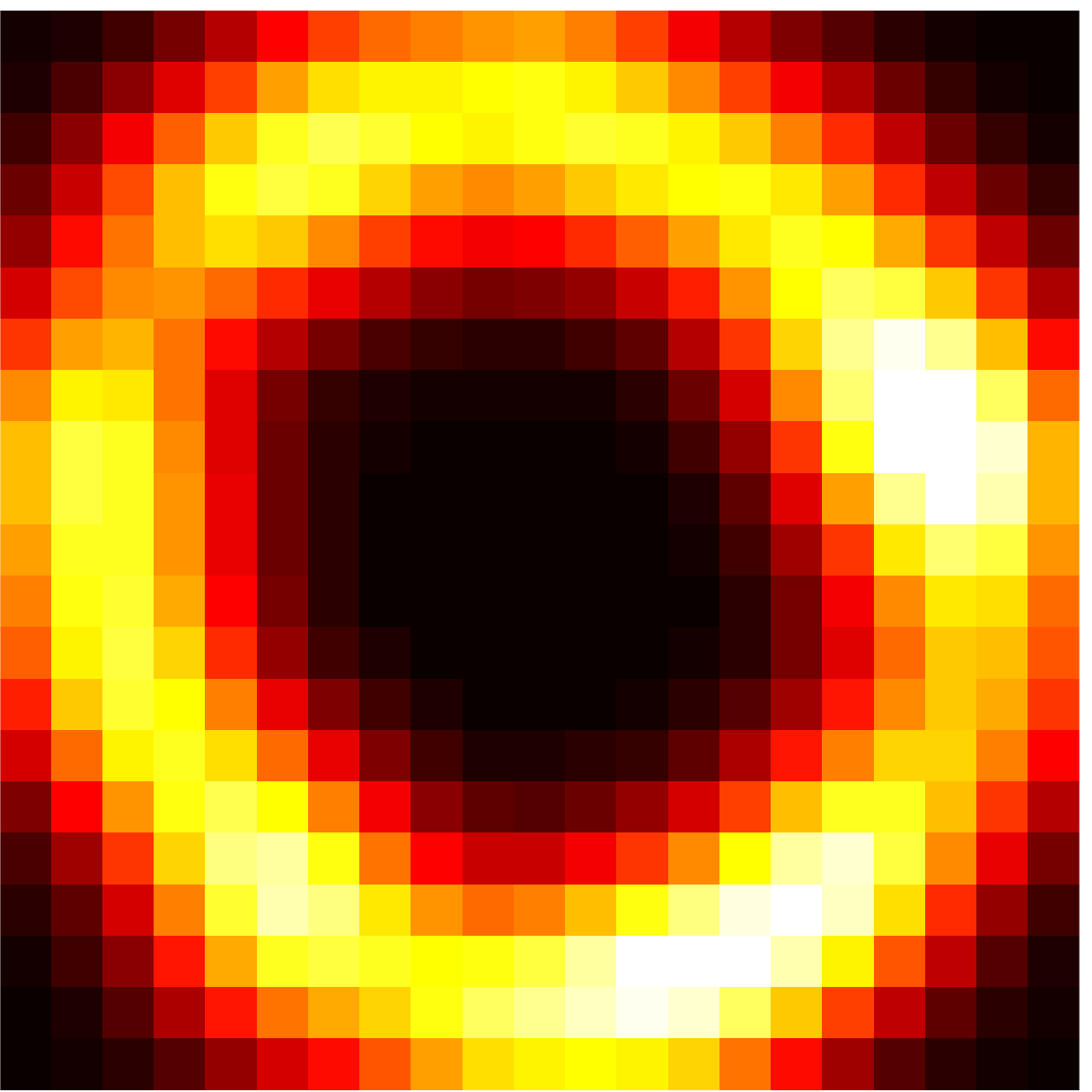}\\ 
\hline
  \end{tabular}
  \caption{\label{fig:PDF_est} Density estimation for the ring dataset by KECA and OKECA using different number of extracted features $N_c$ and approaches to estimate the kernel length-scale parameter $\sigma$. Black color represents low PDF values and yellow color high PDF values.}
\end{figure}

Figure 2 illustrates the ability of KECA and OKECA for density estimation in the ring dataset (see first example in Fig. 1 to analyze the ring dataset distribution). We merely applied Eq.~(\ref{pdf3}) for different number of components $r$ in $\A_r$. 
Note that, for the proposed OKECA, the first projection concentrates most of the entropy information. This agrees with the fact that just one dimension is needed to obtain a good PDF estimation. On the contrary, KECA cannot estimate correctly the PDF using only the first component and actually needs at least five components. This issue is even more dramatic when using $\sigma_{ML}$ (see Fig.\ \ref{fig:PDF_est}). 
It is worth noting that $\sigma_{ML}$ and $\sigma_{d2}$ give rise to the best PDF estimates.  

\subsection{OKECA components for data classification}
\label{PDF_est_examples}

We here illustrate the capabilities of OKECA for data classification. {A great many feature extraction methods exist, both linear like PCA \cite{bk:Jollife86}, ICA \cite{Hyvarinen01} or AWICA \cite{Mammone12} and nonlinear such as the family of kernel multivariate analysis~\cite{Arenas13spm}. In this paper, however, for the sake of a fare comparison, we restrict to compare our OKECA proposal to the original KECA counterpart, which are the only existing unsupervised feature extraction kernel methods based on the same principle of entropy maximization.} The experiments are conducted on a wide range of synthetic and real problems: 1) The \emph{two moons} and the \emph{pinwheel} datasets considered previously in Sec. \ref{Entro}; 2) Six real datasets from the University California Irvine (UCI) Machine Learning Repository\footnote{\url{http://archive.ics.uci.edu/ml/datasets.html}}; and 3) A real satellite multispectral image classification problem. In order to evaluate the data classification, we have used the overall classification accuracy (OA) which is obtained as the average of samples correctly predicted in percentage terms. While one could classify on top of the extracted features, we here rely intentionally on the class-dependent estimated densities and perform maximum a posteriori (MAP) classification. 

\subsubsection{Synthetic datasets}

\begin{figure}[t!]
  \centering
\setlength{\tabcolsep}{0.01pt}
  \begin{tabular}{cc}
\includegraphics[width=4.6cm]{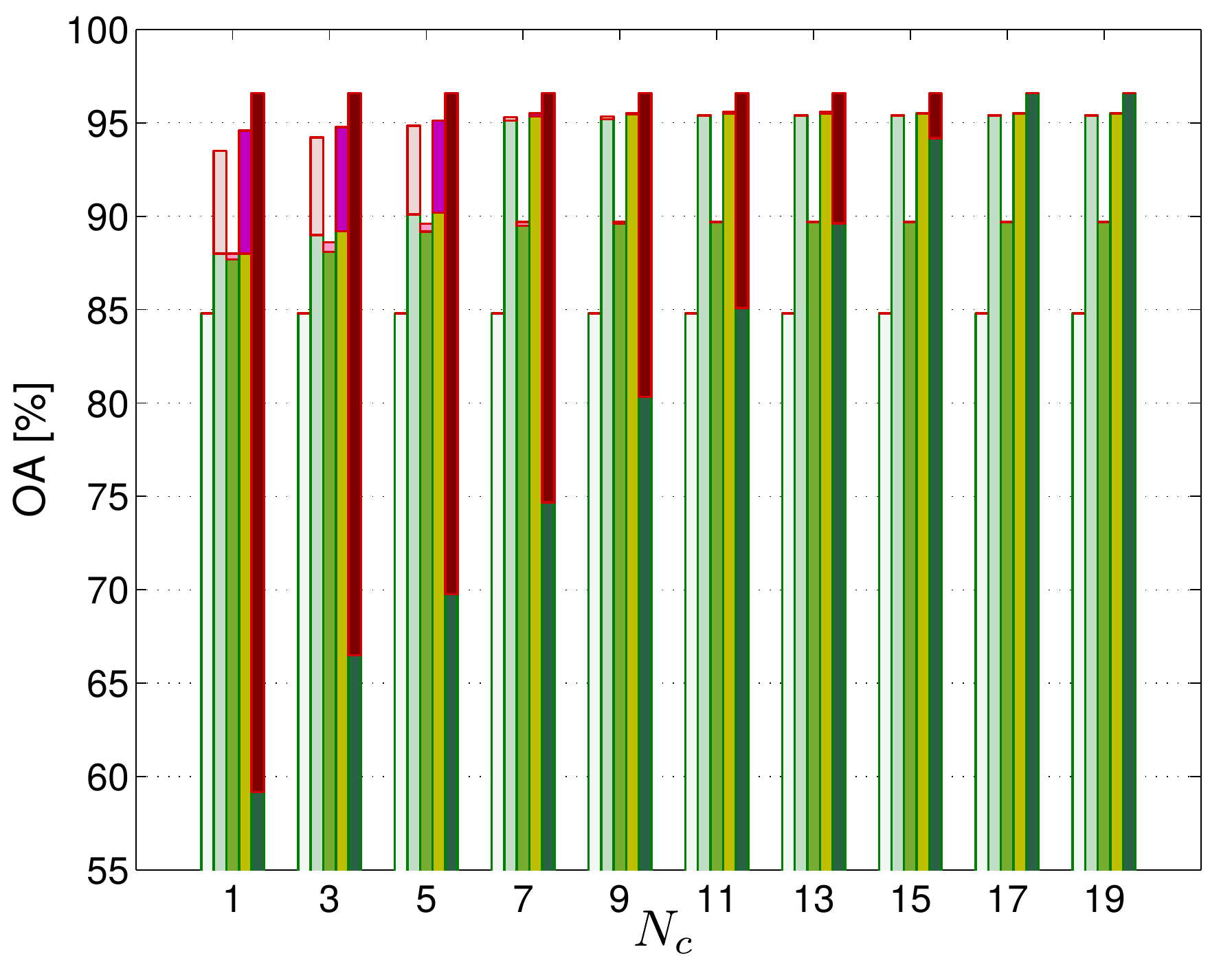} & 
\includegraphics[width=4.6cm]{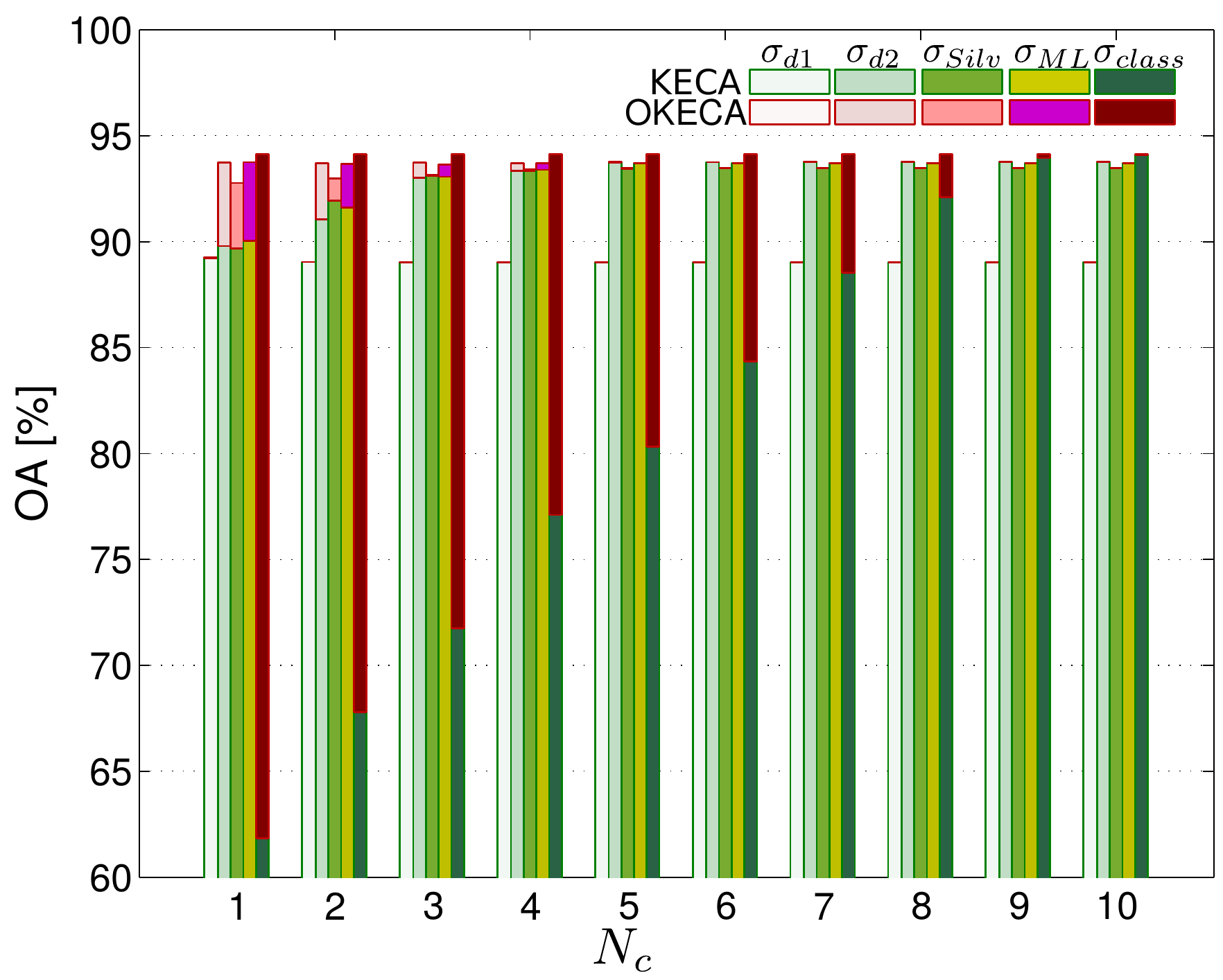} 
  \end{tabular}
  \caption{\label{classIDA} Overall Accuracy obtained with {\em two-moons} (left) and {\em pinwheel} (right) datasets. The five bars for every number of retained features are, from left to right: $\sigma_{d1}$, $\sigma_{d2}$, $\sigma_{Silv}$, $\sigma_{ML}$ and $\sigma_{class}$.}
\label{ClassIDA}
\end{figure}

Figure \ref{ClassIDA} shows the test OA obtained with different $\sigma$ values and different number of retained dimensions with KECA and the proposed OKECA on the {\em two-moons} and {\em pinwheel} datasets. The five bars for every number of retained features are, from left to right: $\sigma_{d1}$, $\sigma_{d2}$, $\sigma_{Silv}$, $\sigma_{ML}$ and $\sigma_{class}$. The value of $\sigma_{class}$ has been optimized for classification using all features in a $5$-fold cross-validation scheme. We used $20$ samples and $45$ samples {\em per} class for training {\em two-moons} and {\em pinwheel} respectively, and $500$ {\em per} class for testing the models and computing the test OA in both datasets. Note that the OKECA method achieves better classification results than KECA for all $\sigma$ values, confirming that to seek for optimally entropic data descriptors may benefit classification. Smaller differences between methods are observed as the number of components increases. When all $n$ features are used, OKECA 
and KECA are trivially equivalent. 

\begin{figure}[t!]
{
 \centering
\setlength{\tabcolsep}{0.01pt}
  \begin{tabular}{cc}
\includegraphics[width=4.6cm]{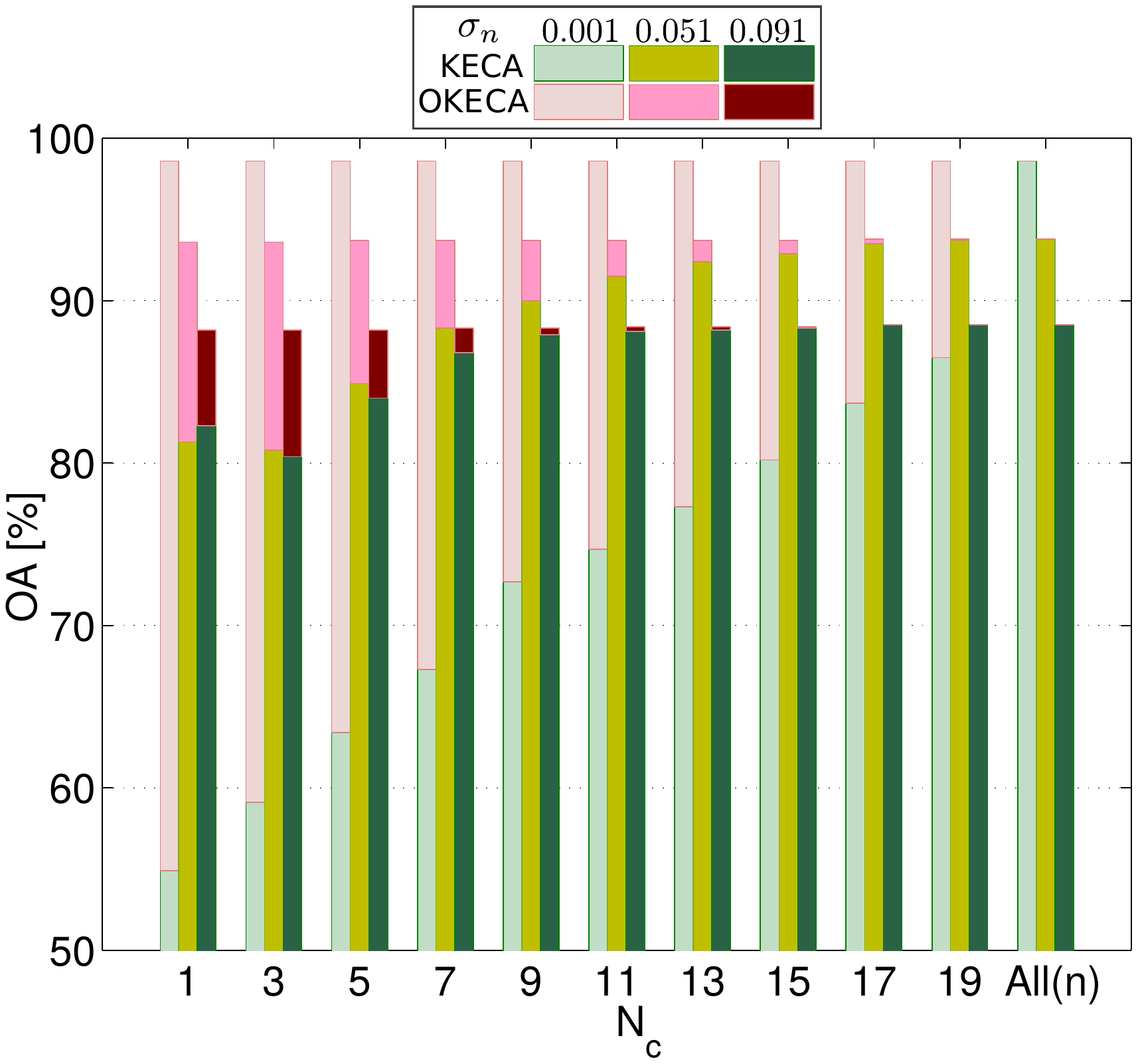} & 
\includegraphics[width=4.6cm]{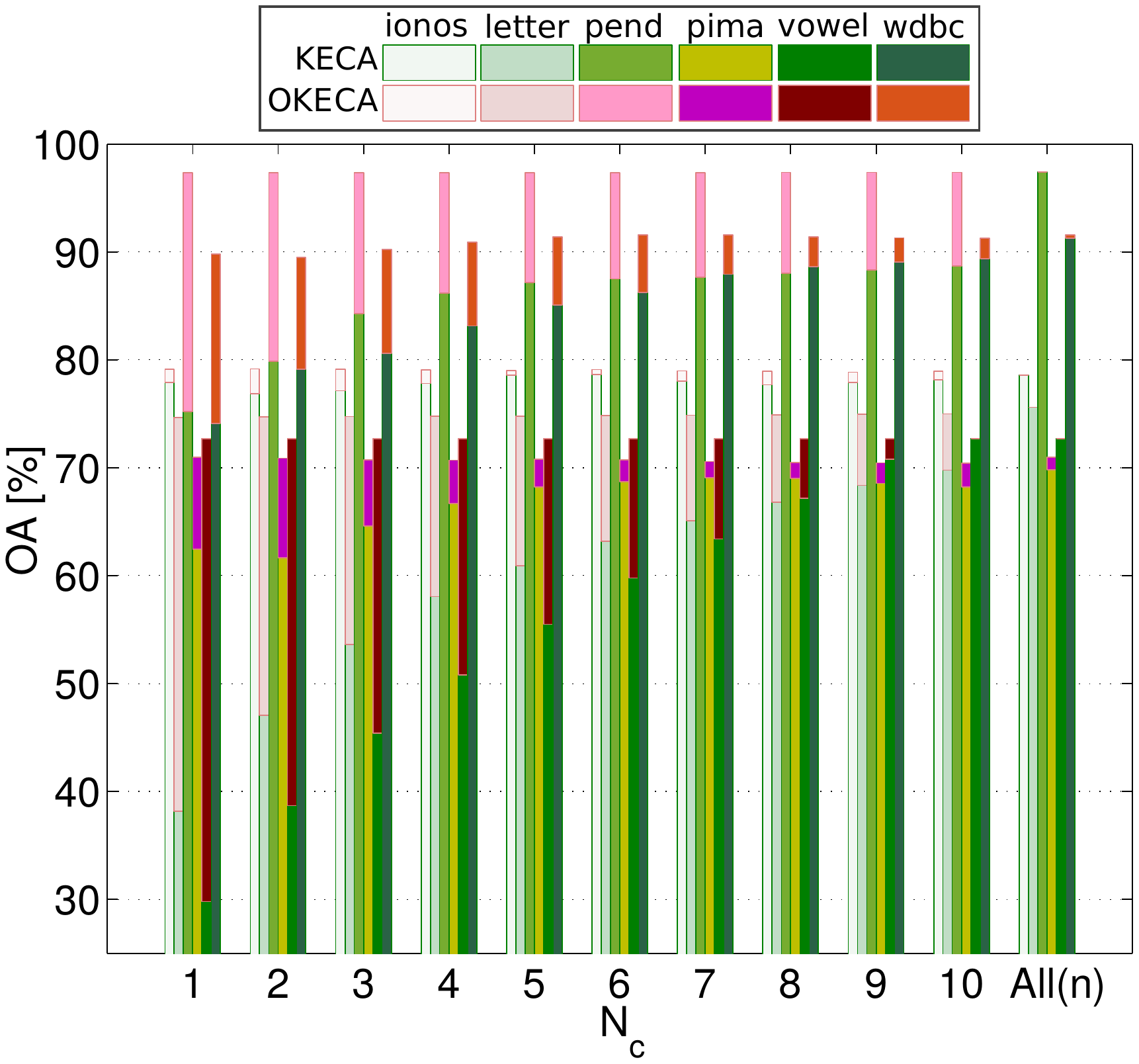} 
  \end{tabular}
  \caption{{ Overall Accuracy obtained by KECA and OKECA methods using different values of noise (left) and UCI database (right) with different number of dimensions. The bars for every number of retained features are different noise values (left) and different databases (right), respectively.}}
\label{Noise}}
\end{figure}

In the following, we discuss the capabilities of OKECA in the presence of distorted distributions. The question raised is how sensitive is the optimization algorithm to the presence of noise.  
To this end, a toy example of the KECA and OKECA projections in presence of noise is analyzed. 
We used $50$ samples of {\em two-moons} dataset for training and $500$ samples to test the classifier. 
Gaussian noise was added to the original data distributions by varying the dimension-wise standard deviation of the Gaussian noise $\sigma_n$ from $0.001$ to $0.091$. 
{Numerical results are shown in Figure~\ref{Noise} [left]}. Note that one main aim of feature extraction and dimensional reduction methods is achieved using the proposed method: KECA needs at least 12 components to obtain similar results to the ones obtained by OKECA with just one component, even in high-noise regime.

\subsubsection{UCI benchmark datasets}

We used six datasets from the UCI machine learning repository of different sizes and dimensionality: the {\tt Ionosphere} dataset is a  binary classification problem of the quality of the radar signal returned from the ionosphere; the goal for the {\tt Letter} dataset is to detect each of a large number of black-and-white rectangular pixel displays as one of the 26 capital letters in the English alphabet; the {\tt Pendigits} problem deals with the recognition of pen-based handwritten digits; the {\tt Pima-Indians} dataset constitutes a classical problem of diabetes diagnosis in patients from clinical variables; the {\tt Vowel} dataset deals with the vowels detection problem in japanese and contains data from a large number of time series of cepstrum coefficients taken from speakers; and finally {\tt wdbc} is another clinical problem for diagnosis of breast cancer in malignant/benign classes. The datasets were intentionally selected either because of the observed high collinearity between input features or 
the diversity in number of classes. Table~\ref{DataBase} gives details on the dimensionality, number of classes and training and test samples used in the experiments that follow. 

\begin{table}[t!]
\begin{center}
\footnotesize
\caption{UCI database description ($d$: number of dimensions, $n_c$: number of class, $N_{train}$: number of training samples, and $N_{test}$: number of test samples.\label{DataBase}}
\begin{tabular}{|l| c| c| c| c| c|}
 \hline
\hline
  {\bf Database} & $m$ & $d$ & $n_c$  & $N_{train}$ & $N_{test}$\\
\hline
 \hline
 Ionosphere & $351$ & 33 & 2  & 60 & 172\\
 \hline
 Letter & $20000$ & 16 & 26  & 780 & 3874 \\
 \hline
  Pendigits & $10992$ & 16 & 9 & 450 & 3498 \\
 \hline
  Pima-Indians & $768$ & 8 & 2 & 180 & 330\\
 \hline
  Vowel & $990$ & 12 & 10 & 100 & 330\\
 \hline
  wdbc & $569$ & 30 & 2 & 60 & 344\\
\hline
\hline
\end{tabular}
\end{center}
\end{table}

We run KECA and OKECA for all datasets for different numbers of extracted components. {The average of the OA for the ten first dimensions is shown in Figure~\ref{Noise} [right]}. In this case we restrict ourselves to $\sigma_{ML}$ because of the good performance in the previous experiments and for the sake of simplicity. In general, the OKECA method outperforms the KECA method and, as observed before, OKECA saturates its performance with just the first extracted dimension.

\subsubsection{Satellite image classification}

In this experiment, we apply KECA and OKECA to the segmentation of remotely-sensed multispectral images. Nowadays, sensors mounted on satellite or airborne platforms may acquire the reflected energy by the Earth with high spatial detail and in several wavelengths or spectral channels~\cite{Camps11}. This allows an improved detection and classification of the pixels in the scene. We consider a real multispectral image acquired over a residential neighbourhood of the city of Z\"urich by the QuickBird satellite in 2002. The analyzed image has $329 \times 347$ pixels. Additional spatial information was added by means of morphological operators, so the dataset has 22 input features. The images contain nine classes of interest: water, meadows, trees, asphalt, brick roofs, bitumen, parking lots, bare soil and shadows. The classes of training samples have been labeled by photointerpretation. The considered data is not only high-dimensional but also shows high collinearity since spectral and spatial features 
are 
stacked together at a pixel level. The problem may be 
quite challenging for classification and feature extraction.

\begin{figure}[t!]
  \begin{center}
  \setlength{\tabcolsep}{0pt}
  \begin{tabular}{cc}
$\sigma_{ML}$ &  $\sigma_{Silv}$ \\
\includegraphics[width=4.3cm]{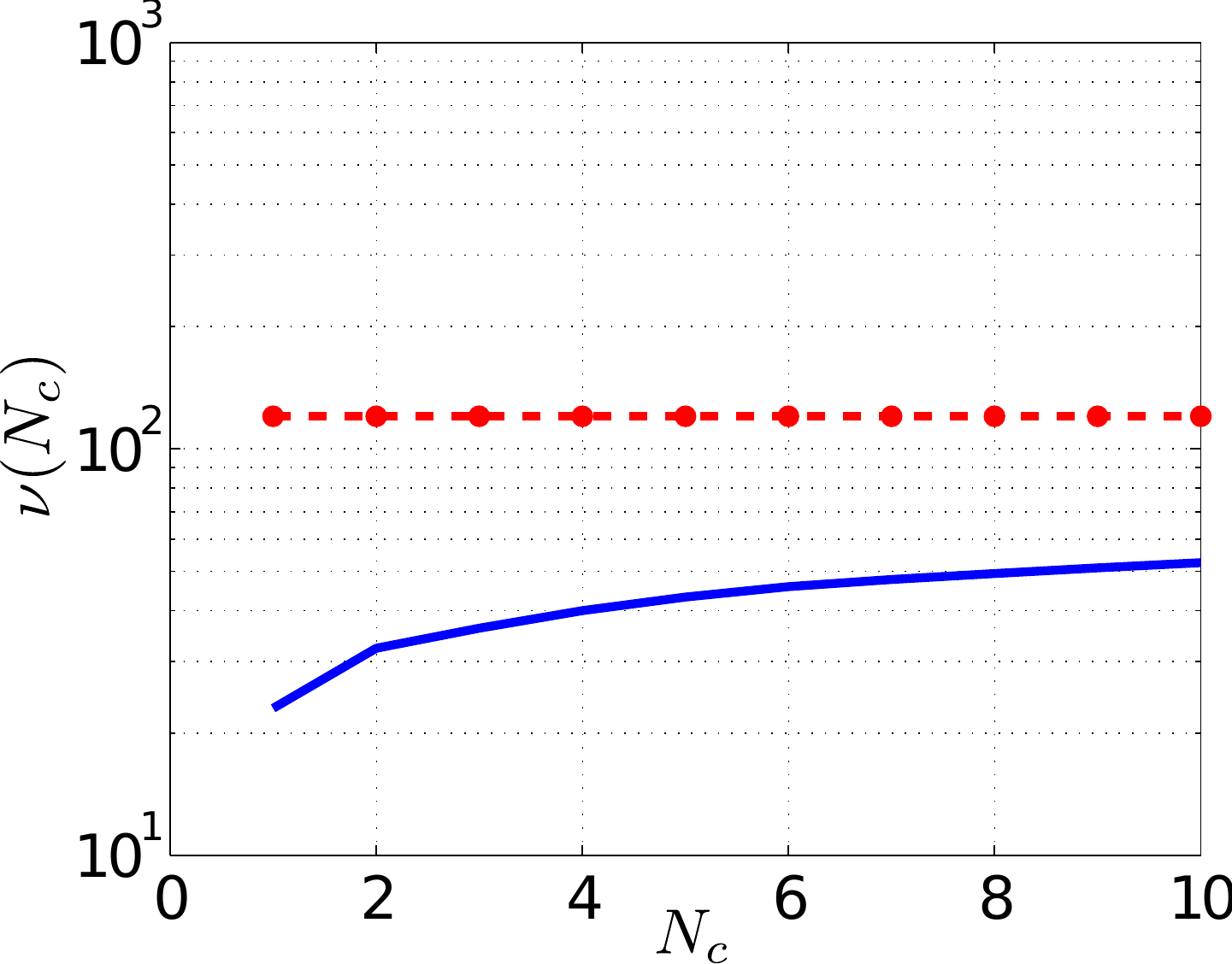} &
\includegraphics[width=4.3cm]{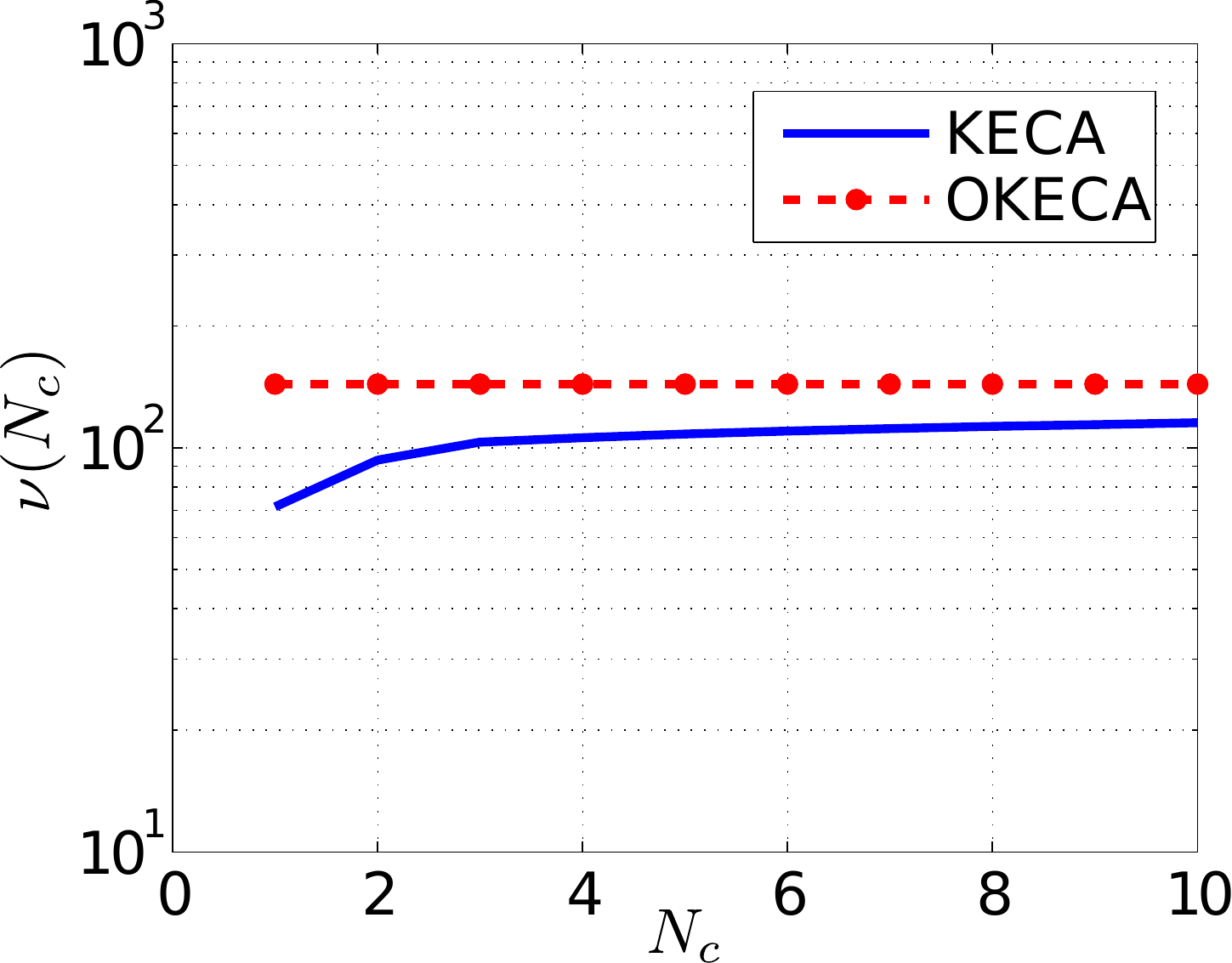} \\ 
$\sigma_{d1}$ &  $\sigma_{d2}$ \\
\includegraphics[width=4.3cm]{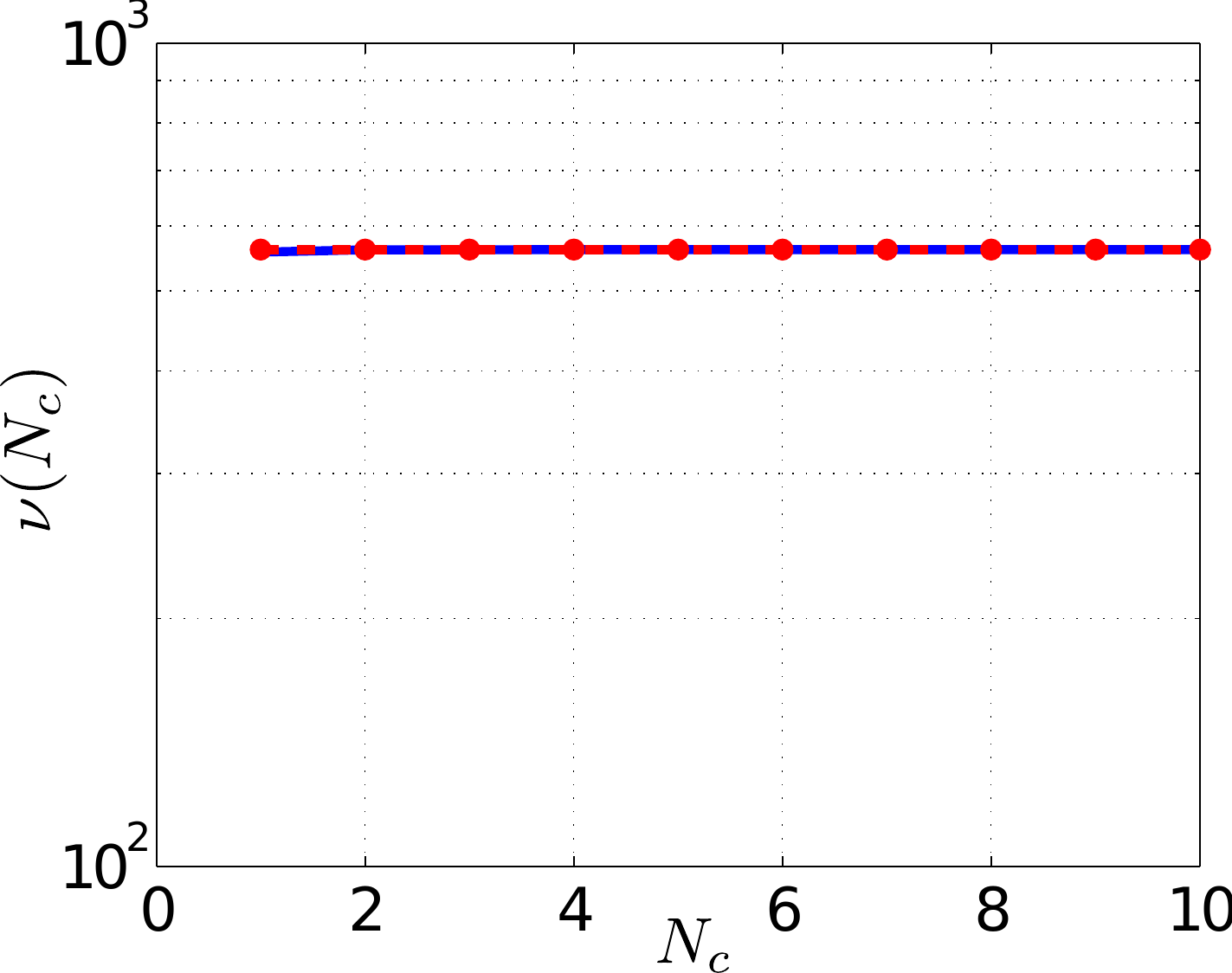} & 
\includegraphics[width=4.3cm]{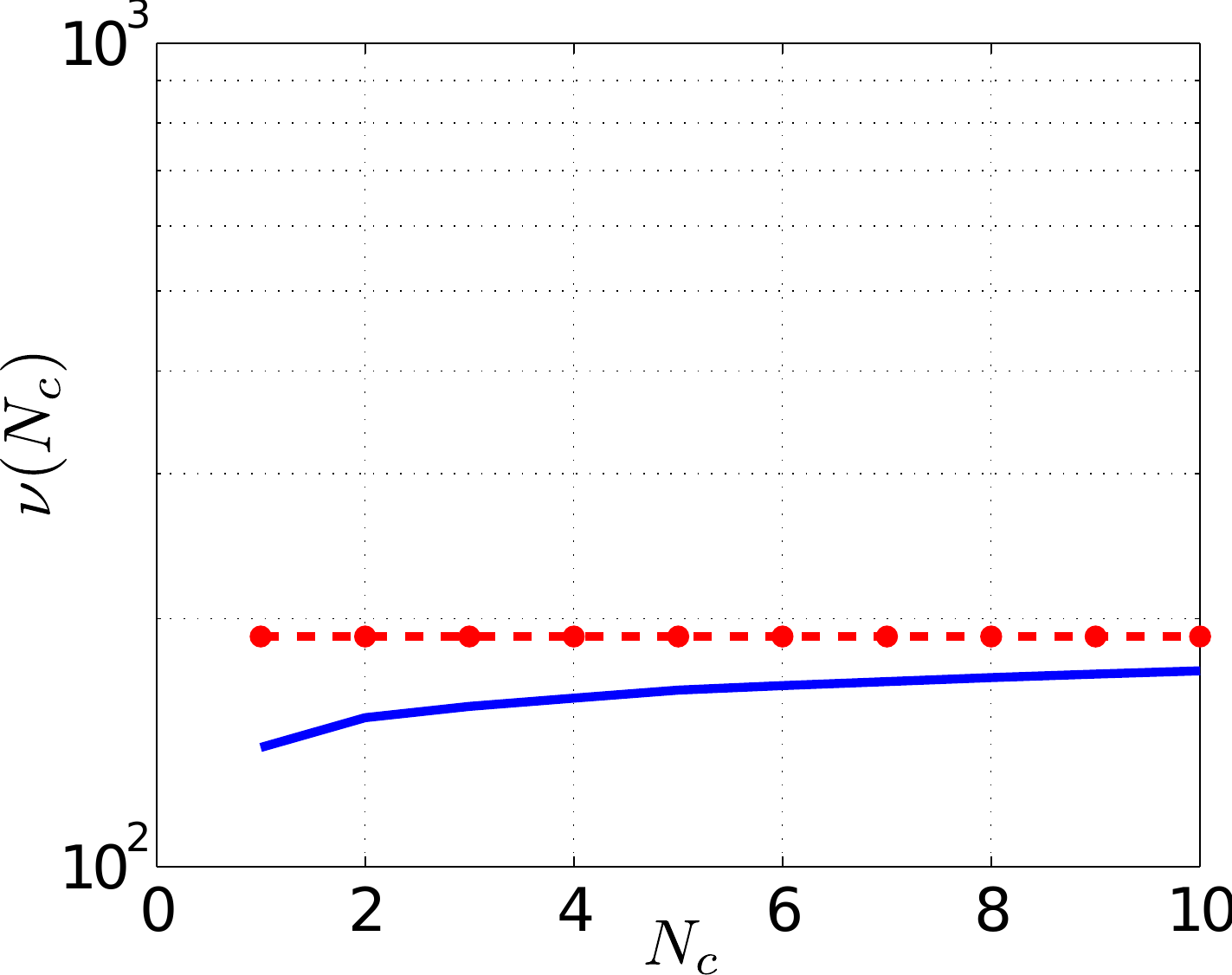} 
  \end{tabular}
  \end{center}
  \caption{\label{fig:entropyquick} The cumulative information potential for the multispectral image dataset using KECA and OKECA and different $\sigma$ estimation approaches. Results for KECA and OKECA in $\sigma_{d1}$ are equal (appear overlapped).}
\end{figure}

The KECA and OKECA cumulative information potential values follow similar trends to the toy examples, see Fig.~\ref{fig:entropyquick}. OKECA reaches the maximum with just one feature, while KECA needs much more components to achieve similar informative content, especially noticeable for the $\sigma_{ML}$ and $\sigma_{d2}$ criteria. Such dependence with the criterion is not shared by OKECA. These results suggest that the sharpness in the component selection made by OKECA is relevant in cases of high feature redundancy as well.

\begin{figure}[t!]
  \centering
  \begin{tabular}{c}
\includegraphics[width=8.5cm]{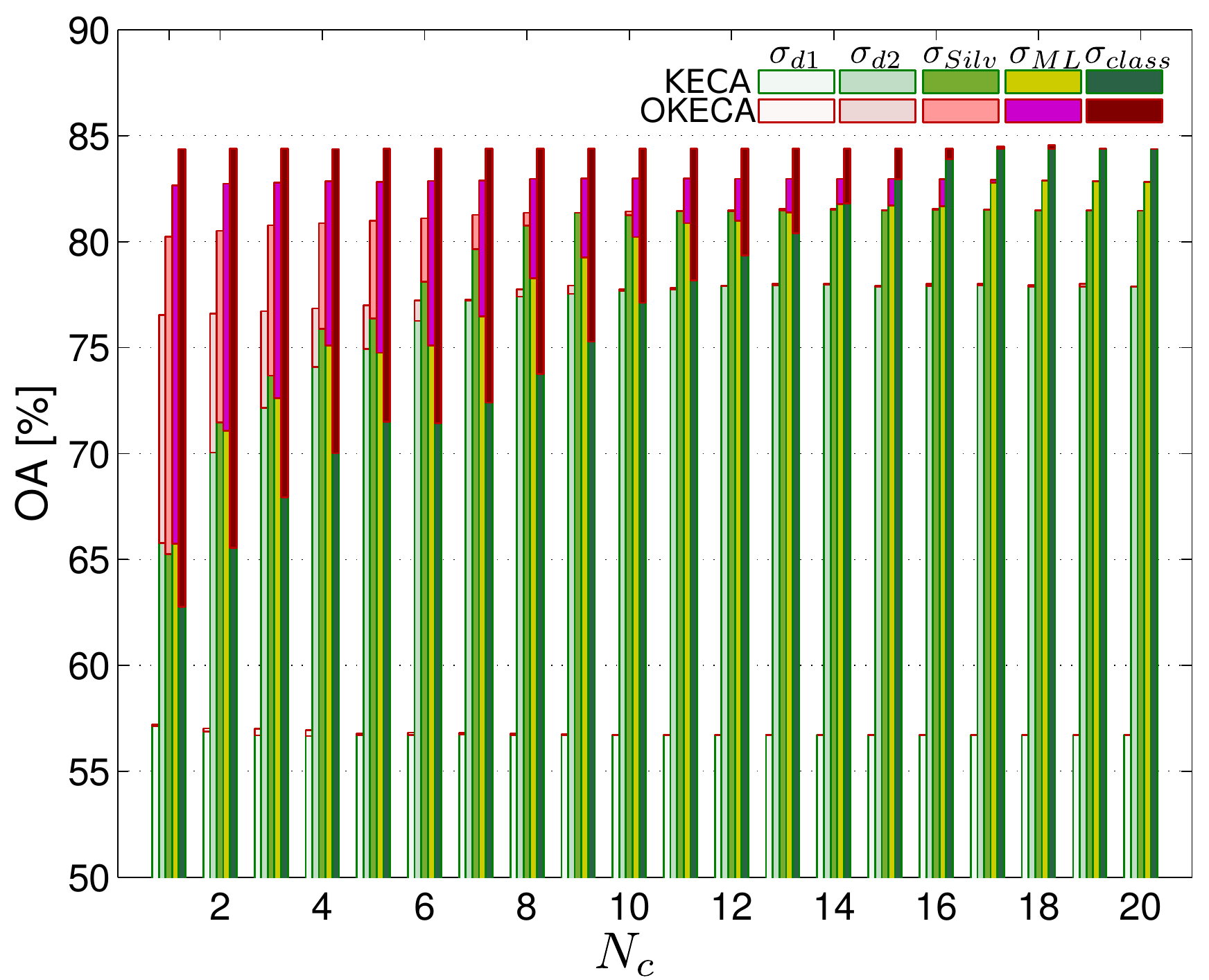} 
  \end{tabular}
  \caption{\label{fig:PDF_class_3} Classification results for the Z\"urich QuickBird satellite image for different $\sigma$ values and numbers of retained dimensions by KECA and OKECA. The five bars for every number of retained features are, from left to right: $\sigma_{d1}$, $\sigma_{d2}$, $\sigma_{Silv}$, $\sigma_{ML}$ and $\sigma_{class}$.} 
\end{figure}

\begin{figure}[h!]
  \centering
\setlength{\tabcolsep}{1pt}
  \begin{tabular}{cc}
RGB composite & Groundtruth map \\
\includegraphics[width=4cm,height=4cm]{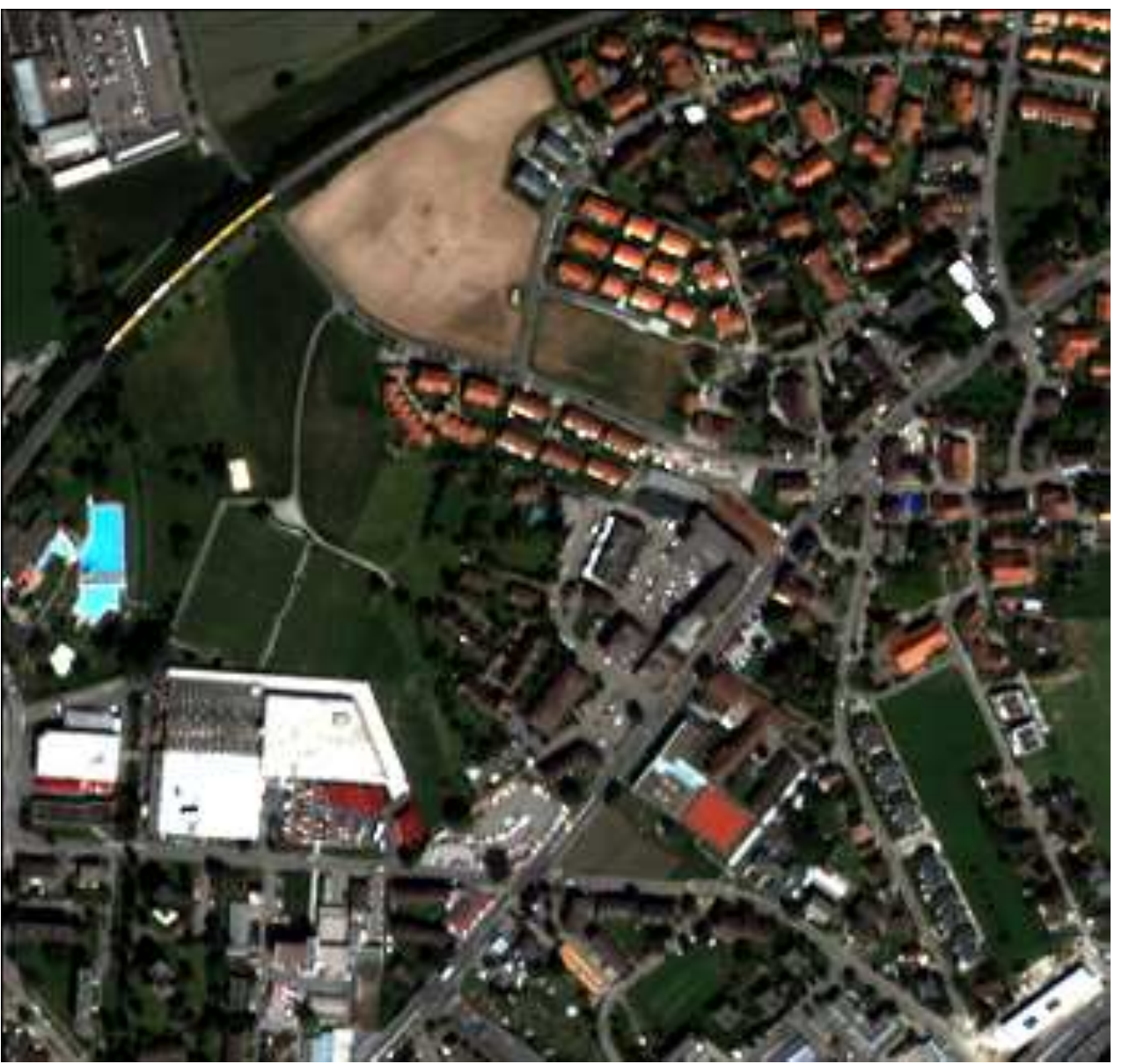}&
\includegraphics[width=4cm,height=4cm]{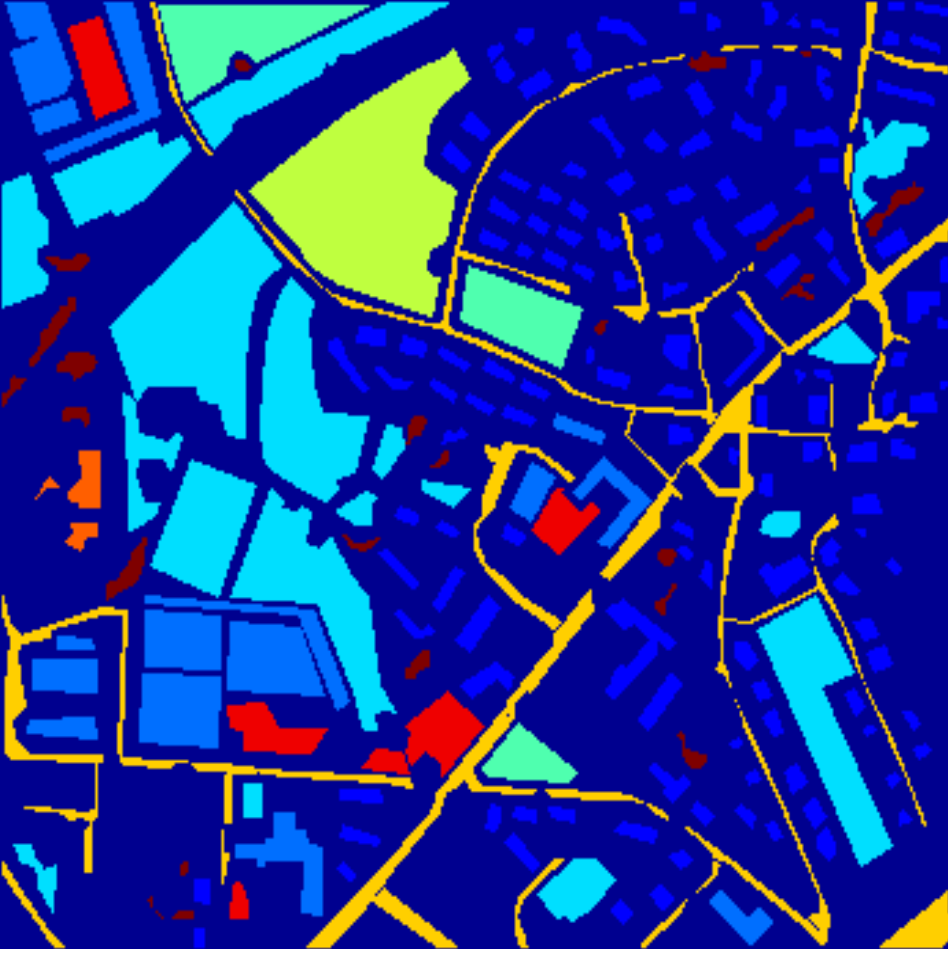} \\ 
KECA $(84.92\%)$ & OKECA $({\bf 90.33}\%)$ \\
\includegraphics[width=4cm,height=4cm]{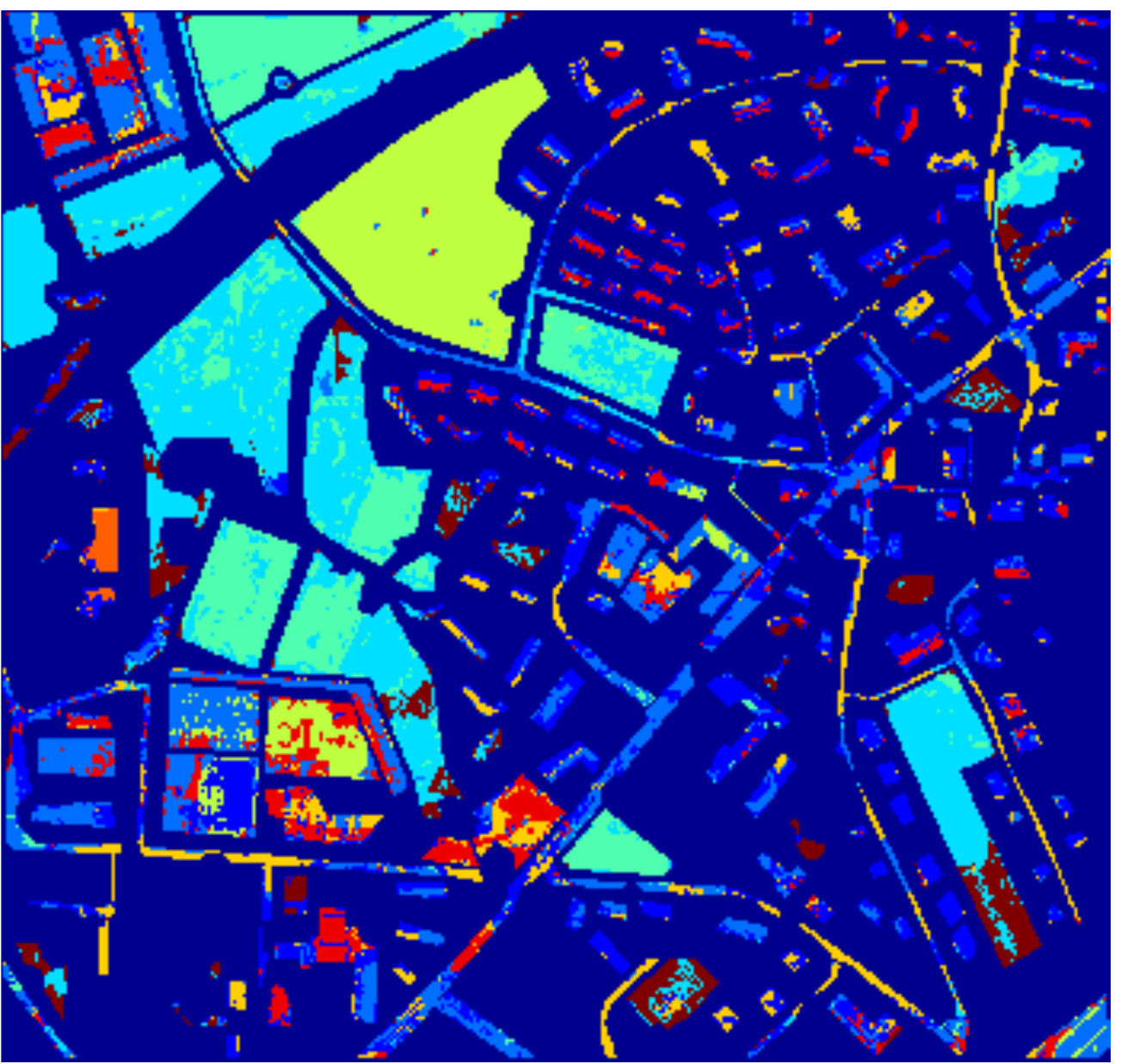}&
\includegraphics[width=4cm,height=4cm]{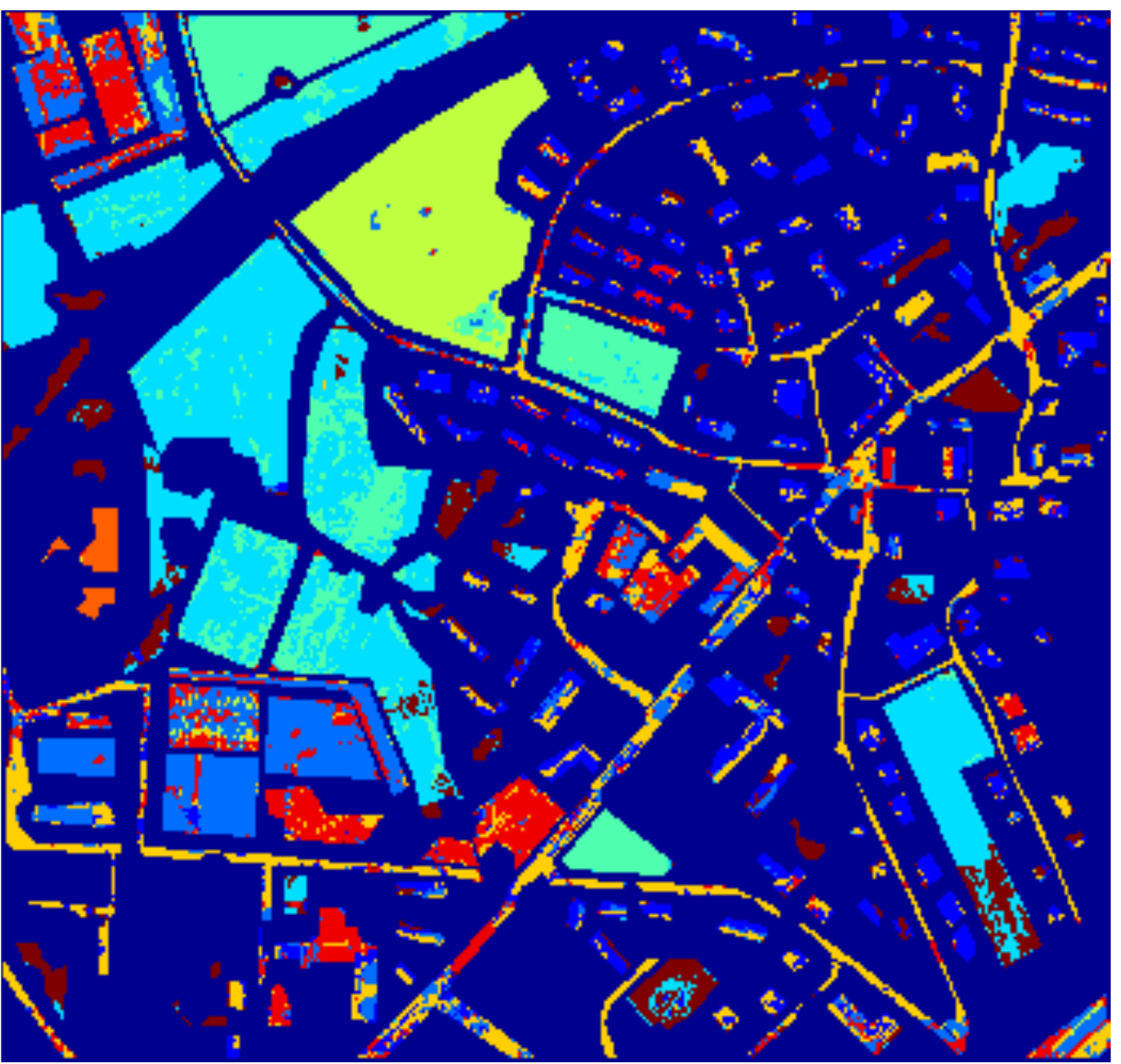} 
  \end{tabular}
  \caption{\label{classBrut} Classification maps for the Z\"urich QuickBird satellite image using three features and $\sigma_{class}$ in KECA and OKECA. Top-left: RBG version of the original image; top-right: ground truth classification map (each color represents a different land-cover class); bottom-left: classification map obtained with KECA; bottom-right: classification map obtained with OKECA.}
\end{figure}

Figure \ref{fig:PDF_class_3} shows the classification results obtained using different $\sigma$ values and different number of retained dimensions. In this case, we use $22$ and $200$ samples {\em per} class for training and testing the models, respectively. Both methods achieve the best results using the $\sigma_{ML}$ and $\sigma_{class}$ criteria. Finally, note that $\sigma_{d1}$, which is a common choice in unsupervised kernel methods, provides very poor results for both methods.
Figure \ref{classBrut} shows the classification maps obtained using three retained features and $\sigma_{class}$ for both methods. Note how OKECA outperforms KECA in general for all the classes.

\subsection{Discussion}

The OKECA potential has been shown in the different experiments. In all of them, the proposed method presents an extraordinary advantage: the information is compacted in very few features (often in just one or two) with higher expressive power optimizing the information potential. That is demonstrated from experimental viewpoint, not only in PDF estimation but also in classification tasks. As we have shown, the proposed method reduces the number of features required clearly improving the results. OKECA estimates correctly the PDF using one dimension concentrating most of the entropy information better than the KECA method, and it is more robust to the selection of the kernel parameter. Furthermore, the experiments show an improvement of the classification results even in the presence of noise. In some situations, using just one OKECA feature is enough to achieve the best overall performance, extracting more components does not add new complementary information and classification results do not change significantly.

\section{Conclusions}
\label{conclusions}

We proposed a highly efficient modification of the KECA algorithm for optimal extraction of entropic kernel components. While KECA reduces to sort the kernel eigenvectors by entropy, OKECA explicitly searches for the features that retain most informative content. We have illustrated the ability of OKECA to retain more information in PDF estimation and classification on both synthetic and real examples. Results consistently showed that OKECA outperforms KECA in terms of information content and robustness. 
In fact, in many experiments, just one or two OKECA components retain almost all the relevant information for data description. Accounting for optimal entropic features allows us to improve the description of the density shape, which is in turn the core for PDF estimation and PDF-based classification. 
Furthermore we have analyzed the effect of using different unsupervised rules to fit the RBF kernel length-scale parameter on KECA and OKECA performances. In general, the maximum likelihood approach showed the best performance. 


\end{document}